%% file: survey.tex
\documentclass[a4paper,fleqn]{cas-sc}
\usepackage{amsthm}
\usepackage{thmtools}

\usepackage[numbers]{natbib}

\usepackage{multirow}

\usepackage{booktabs}
\usepackage{longtable}
\usepackage{amsmath}  %
\usepackage{amsfonts} %
\usepackage{graphicx}
\usepackage{hyperref}
\usepackage[most]{tcolorbox}
\usepackage{makecell}
\usepackage{array}
\usepackage{arydshln}
\usepackage{cancel}
\usepackage{placeins}
\usepackage{float}
\usepackage{colortbl}

\usepackage[tikz]{bclogo}  %
\usepackage{tcolorbox}     %

\newtcolorbox{notebox}{
    colback=blue!5,        %
    colframe=white,%
    fonttitle=\bfseries,   %
    sharp corners,         %
    enhanced jigsaw,       %
}

\newtheorem{definition}{Definition}

\definecolor{darkgreen}{rgb}{0.0, 0.5, 0.0}

\DeclareMathOperator*{\minimize}{minimise}

\def\tsc#1{\csdef{#1}{\textsc{\lowercase{#1}}\xspace}}
\tsc{WGM}
\tsc{QE}
\tsc{EP}
\tsc{PMS}
\tsc{BEC}
\tsc{DE}

\begin{document}
\let\WriteBookmarks\relax
\def\floatpagepagefraction{1}
\def\textpagefraction{.001}

\shorttitle{Modular Federated Learning: A Meta-Framework Perspective}

\shortauthors{Frederico Vicente et~al.}

\title [mode = title]{Modular Federated Learning: A Meta-Framework Perspective}                      
\tnotemark[1]

\tnotetext[1]{The authors would like to thank NOVA LINCS's support through the project (UIDP/04516/ 2020/BIM/10), Portuguese funding institution FCT - Fundação para a Ciência e a Tecnologia through the project PIDDAC, TaRDIS and 2024.03913.BD. The work of D. Jakovetic was also supported in part by the Ministry of Education, Science and Technological Development, Republic of Serbia, and by the European Union’s Horizon Europe Research and Innovation program, grant agreement IDs 101135775 and 101093006. The work of D. Jakovetic was further supported by the Science Fund of The Republic of Serbia, the LASCADO project, grant no. 7359. Special thanks are extended to the following colleagues for helping with revisions: Filipa Valdeira, Francisco Caldas, Frederico Metelo and Stevo Racković.}

\author[1]{Frederico Vicente}[
    orcid=0000-0002-6807-1644
]
\ead{fm.vicente@campus.fct.unl.pt}

\author[1]{Cláudia Soares}[
    orcid=0000-0003-3071-6627
]
\ead{claudia.soares@fct.unl.pt}

\author[2]{Dušan Jakovetić}[
    orcid=0000-0003-3497-5589
]
\ead{dusan.jakovetic@dmi.uns.ac.rs}

\affiliation[1]{organization={NOVA School of Science and Technology},
    country={Portugal}
}

\affiliation[2]{organization={University of Novi Sad, Faculty of Sciences},
    country={Serbia}
}
\date{\today}

\begin{abstract}
Federated Learning (FL) enables distributed machine learning training while preserving privacy, representing a paradigm shift for data-sensitive and decentralized environments. Despite its rapid advancements, FL remains a complex and multifaceted field, requiring a structured understanding of its methodologies, challenges, and applications. In this survey, we introduce a meta-framework perspective, conceptualising FL as a composition of modular components that systematically address core aspects such as communication, optimisation, security, and privacy. We provide a historical contextualisation of FL, tracing its evolution from distributed optimisation to modern distributed learning paradigms. Additionally, we propose a novel taxonomy distinguishing Aggregation from Alignment, introducing the concept of Alignment as a fundamental operator alongside aggregation. To bridge theory with practice, we explore available FL frameworks in Python, facilitating real-world implementation. Finally, we systematise key challenges across FL sub-fields, providing insights into open research questions throughout the meta-framework modules. By structuring FL within a meta-framework of modular components and emphasising the dual role of Aggregation and Alignment, this survey provides a holistic and adaptable foundation for understanding and advancing FL research and deployment.
\end{abstract}


\maketitle

\input{survey_sections/introduction}
\input{survey_sections/formulation}
\input{survey_sections/history_context}

\input{survey_sections/toolbox}

\input{survey_sections/frameworks}
\input{survey_sections/alternative_paths}
\input{survey_sections/discussion}
\input{survey_sections/next_steps}
\input{survey_sections/conclusion}

\printcredits

\bibliographystyle{cas-model2-names}

\bibliography{cas-refs,consensusOptGradientMethods}

\end{document}

%% file: survey_sections/introduction.tex
\section{Introduction}
\label{sec:introduction}

At present, \textbf{digital privacy} has become a paramount concern, propelling the development of regulations such as the EU AI Act, the EU General Data Protection Regulation (\href{https://gdpr.eu/tag/gdpr/}{GDPR}), and the California Consumer Privacy Act (\href{https://oag.ca.gov/privacy/ccpa}{CCPA}). These concerns and regulations profoundly influence how data science and Machine Learning are conducted (\href{https://artificialintelligenceact.eu/ai-act-explorer/}{EU AI Act}, ~\citet{sok}).
Inter-continental organisations can no longer freely transfer data from the EU to the USA, compelling them to seek alternative methods to derive insights or train models on the entirety of their data. 
Privacy, however, is merely one reason for exploring solutions beyond centralised data learning schemes. Consider the vast proliferation of Internet of Things (IoT) devices generating continuous streams of raw data. Leveraging these data streams effectively can lead to the development of more sophisticated and resilient Machine Learning models. Without employing distributed learning algorithms, this data would need to traverse digital networks, potentially leading to network saturation with sensitive information. Avoiding such network congestion and the high latency associated with centralised learning solutions is crucial. Additionally, mobile phones, being inherently personal devices, necessitate \textbf{personalisation} features in such systems. However, effective Machine Learning systems typically require substantial datasets to perform optimally. With increasing data protection requirements such as GDPR and CCPA, and in scenarios where data is continuously generated by distributed IoT devices, centralised data aggregation is often impractical or even unfeasible. In these contexts, Federated Learning offers a compelling alternative, enabling collaborative model training without exposing raw data.

\begin{figure}[!htbp]
    \centering
    \includegraphics[clip,width=0.5\linewidth]{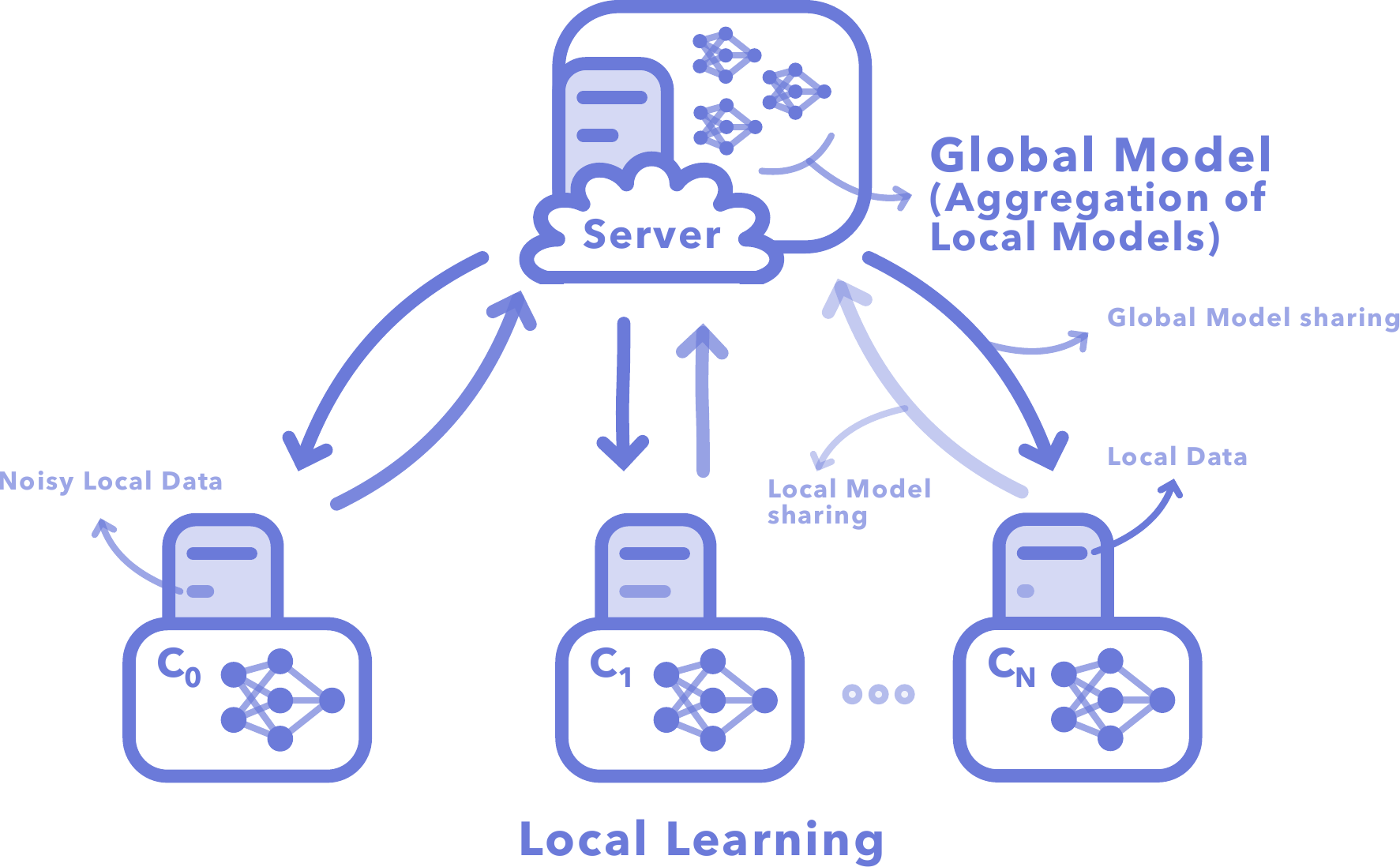}
    \caption{Standard centralised Federated Learning Setting. Each client $c$ learns a local model and shares it with a central server, aggregating the individual model parameters and then sending this new global model back to the clients. This process is repeated for several rounds until a certain criterion is met.}
    \label{fig:star_schema_fl}
\end{figure}

The term Federated Learning was coined by a Google research team~\citep{FedAvg, federated}, introducing the concept of a server-client distributed learning schema for solving non-convex problems. They proposed a distributed learning process over edge devices (i.e. mobile phones), which, without compromising private data, could collaboratively engage in learning tasks (see Fig.~\ref{fig:star_schema_fl} for a visual abstraction of the entities involved in a standard server-client FL network topology).

The paradigm of distributed learning, however, dates back several decades, originating with the concept of \textbf{distributed optimisation}. Over the past forty years, numerous approaches have been developed, ranging from non-cooperative~\citep{game_theory_smart_grids} to cooperative distributed learning environments~\citep{the_ant_system}, and from server-client topologies~\citep{ParameterSF} to decentralised ones~\citep{distributed_subgradient_multi_agent,Tsitsiklis_decentralized_detection}. Various algorithmic families embody the principles of distributed optimisation, including methods such as the Alternating Direction Method of Multipliers (ADMM:~\citet{Boyd2011DistributedOA}), Block Coordinate Descent (BCD), and distributed subgradient methods. These methods share similarities with the optimisation techniques employed in Federated Learning. Moreover, in the same spirit as the Federated Learning genesis paper, many prior works had already addressed the distributed solving of non-convex problems (e.g., neural networks). For example, the 1-bit quantisation of Deep Neural Networks using distributed SGD~\citep{distributed_sgd_20141bit} in 2014 explored many challenges still relevant in today's Federated Learning field, such as communication-efficient methods and quantisation mechanisms. Both distributed optimisation and computer science distributed systems have a rich history~\citep{Lamport1998ThePP}.

In summary, Federated Learning is particularly beneficial in specific scenarios:

\begin{itemize}
    \item It excels in distributed systems where the number of devices far exceeds the number of data centre nodes~\citep{keyboard_prediction}.

    \item It is well-suited for situations where data privacy is paramount~\citep{hybrid_private_fl}.
    \item FL is advantageous when dealing with numerous small datasets distributed across individual devices~\citep{FedAvg}.

    \item Data from edge devices is often unlabeled, but on-device pseudo-labeling techniques can be utilised to personalise user experiences through Federated Learning~\citep{amazon_noisy_feedback}.

    \item Resource-limited devices may find data transmission prohibitive in terms of energy, assuming that Federated Learning communication algorithms are less demanding than sharing raw data~\citep{melon_tee}.

\end{itemize} 

However, alongside its promising applications, Federated Learning faces several critical challenges that must be addressed to ensure optimal performance. These include the need for communication-efficient methods to reduce bandwidth usage, the difficulty of dealing with hardware and data heterogeneity across clients, and the challenge of maintaining strong privacy guarantees while ensuring robustness against adversarial attacks. Moreover, aligning distributed knowledge effectively without significant performance loss remains an open problem, as does the absence of standardized frameworks for FL system design. Finally, the environmental impact, particularly the carbon footprint associated with large-scale decentralized training, also raises concerns about sustainability.

For a comprehensive understanding of the Federated Learning landscape, we encourage readers to consult contemporary surveys addressing various aspects of FL. \citet{liu2023recent_survey} analyses heterogeneous FL, fairness considerations, and aggregation methodologies. 
\citet{li2023interpretable}, take a distinct approach, focusing on the development of interpretable FL pipelines. The integration of Bayesian methods within the FL framework is thoroughly examined in another significant survey~\citep{BayesianFLSurvey}. Furthermore, the survey~\citep{fl_open_problems} offers an extensive discussion on security and privacy issues, fairness in FL solutions, the challenges posed by non-independent and identically distributed (non-iid) data, and strategies for effective and efficient handling of heterogeneous learning environments.
In alignment with the aforementioned surveys, particularly \citet{fl_open_problems}, and \citet{liu2023recent_survey}, our survey aims to provide an accessible introduction to Federated Learning while presenting a comprehensive overview of the research and developments in the field. However, our contribution diverges by proposing a novel perspective on FL as a Meta-Framework composed of modules. Additionally, we find it relevant to discuss FL in the context of distributed optimisation and to address trustworthy Machine Learning within FL contexts, as highlighted by \citet{Mucsanyi2023TrustworthyML}.

We include a representative table that summarizes key directions in current survey literature, outlining their core focus areas and contributions (see Table~\ref{tab:surveys_comparison}).

\input{tables/surveys_comparison}

\paragraph{To provide a comprehensive overview, we now outline our core contributions:}

\begin{itemize}

    \item \textbf{Systematisation of Federated Learning as a Meta-Framework.} We propose a Meta-Framework perspective of Federated Learning, structuring it as a modular and composable system rather than a monolithic methodology. This approach provides a unifying blueprint, organizing key FL dimensions, such as communication, aggregation, security, into interoperable modules. It enhances systematic design, and adaptation to diverse real-world scenarios. Beyond its structural benefits, this taxonomy also simplifies the understanding of FL by breaking it into core components, offering a clearer pedagogical framework. Ultimately, we hope this perspective facilitates both theoretical analysis and practical implementation, making FL more accessible, adaptable, and intuitive for newcomers.

    \item \textbf{Novel Alignment Operator.} We propose a novel operator \textbf{Alignment}  to explicitly capture and formalize knowledge alignment between clients, distinguishing our approach from prior formulations. We argue that this alignment operator is as fundamental as aggregation in Federated Learning. Aggregation refers to the process of combining model updates from distributed clients to form a global model, ensuring knowledge is shared across the system.

    Alignment, on the other hand, constrains the aggregation process to meet specific objectives, such as fairness, convergence, or robustness. This perspective highlights the dual importance of aggregation and alignment in driving effective learning across distributed systems.
    
    \item \textbf{Historical contextualisation.} We trace the evolution of Federated Learning, highlighting its foundational ties to distributed optimisation.
    
    \item \textbf{Entry-level understanding.} We provide a comprehensive yet accessible introduction to Federated Learning, catering to newcomers and facilitating their engagement with the topic.

    \item \textbf{Federated Learning Python Frameworks Systematisation.} We conduct a comprehensive exploration of existing Python-based Federated Learning frameworks, systematically analysing their features, capabilities, and suitability for different use cases.

    \item \textbf{Challenges in Sub-Fields.} We identify and systematise the key challenges faced within various sub-fields of Federated Learning.
    
\end{itemize}

Given the extensive scope of this survey, we provide a detailed roadmap to ease navigation through the various sections:

\begin{itemize}
    \item \textbf{Section~\ref{sec:preliminaries}: Preliminaries.} This section provides foundational knowledge to understand the FL framework, starting with key mathematical notation. It then formulates the distributed optimization problem, outlines core FL family of algorithms, and presents various architectural paradigms. Finally, it presents an innovative modular Meta-Framework approach, conceptualising FL as a structured blueprint of interconnected modules.

    \item \textbf{Section~\ref{sec:history}: Distributed Optimisation - A Historical Perspective.} This section examines the evolution of distributed optimisation and its relationship with Federated Learning.

   \item \textbf{Section~\ref{sec:fl_in_detail}: Dissecting the FL Modular Meta-Framework.} This section systematically deconstructs Federated Learning through the lens of a modular meta-framework, organizing existing literature across key FL domains. By structuring FL as a composable set of interoperable modules, this taxonomy provides a coherent and intuitive perspective, making it easier to grasp the fundamental principles and interconnections within FL.
   
\item \textbf{Section~\ref{sec:frameworks}: Python FL Frameworks.} This section provides a practical perspective on FL, with an in-depth discussion of the available Python frameworks for implementing distributed learning. We examine their features, capabilities, and limitations.

    \item \textbf{Section~\ref{sec:alternatives}
: Alternatives and Comparisons.} Here, we clarify the distinctions between Federated Learning and other techniques that involve distributed systems and Machine Learning but diverge from the FL paradigm.

    \item \textbf{Section~\ref{sec:discussion}: Real-World Applications and Challenges.} We present real-world examples of FL frameworks in use and propose potential future directions for use cases. We emphasise the significance of open platforms for collaborative training and sharing FL models. Given the complexity of FL frameworks, which integrate distributed systems, mathematical optimisation, and software engineering, we highlight the necessity of advanced methods for monitoring, debugging, and profiling in practical FL systems. The section concludes with insights into FL's challenges, including environmental impact and hardware constraints.

    \item \textbf{Section~\ref{sec:next_steps}: Future Research Directions.} This section motivates and explores potential avenues for future research in the FL domain, considering emerging trends and unresolved issues.
    
    \item \textbf{ Section~\ref{sec:conclusion}: Conclusion.} We conclude by summarising the journey through the Federated Learning landscape and remarking on the importance of a privacy-aware distributed design to tackle critical domain use cases. We further remind that FL is a multidisciplinary field encompassing knowledge from distributed systems, optimisation algorithms, and software engineering.
\end{itemize}

%% file: tables/surveys_comparison.tex
\begin{longtable}{p{4cm} p{3cm} p{8cm}}

\caption{Representative overview of Federated Learning Survey Literature by Focus Area and Contributions\label{tab:surveys_comparison}} \\
\toprule
\textbf{Survey} & \textbf{Focus} & \textbf{Key Contributions} \\
\midrule
\endfirsthead

\toprule
\textbf{Survey} & \textbf{Focus} & \textbf{Key Contributions} \\
\midrule
\endhead

\midrule
\multicolumn{3}{r}{\textit{Continued on next page}} \\
\midrule
\endfoot

\bottomrule
\endlastfoot

\multicolumn{3}{l}{\textbf{Foundational and General}} \\

\citet{fl_open_problems} & FL overview and open challenges & Provides a comprehensive and in-depth overview of the federated learning field, along with a well-founded set of research challenges that guide future work. \\[1.5pt] 
\hdashline 

\rowcolor{green!10} \vspace{0.5cm} \textbf{Ours (2025)}  & \vspace{0.1cm} FL overview from a \textbf{Meta-Framework Perspective} & \vspace{0.1cm}Systematises FL as a modular Meta-Framework with modular components; proposes Alignment operator taxonomy to guide knowledge aggregation; traces historical evolution from distributed learning; surveys current Python frameworks. \\

\midrule 
\multicolumn{3}{l}{\textbf{Privacy \& Security}} \\

\citet{decentralized_fl_privacy} & Privacy methods & Surveys security and privacy threats in decentralized federated learning, examining adversary models, defense mechanisms, and the role of trust and verifiability in the absence of a central server. \\[1.5pt] 
\hdashline

\addlinespace[6pt] 

\citet{robustness_survey} & Robustness & Presents a comprehensive survey of privacy and robustness in FL, offering a structured taxonomy of threat models, attacks, and defenses, while highlighting key techniques, assumptions, and future directions toward secure and resilient FL systems. \\[1.5pt] 
\hdashline

\addlinespace[6pt] 

\citet{membership_survey} & Membership inference attacks & Provides a comprehensive survey on membership inference attacks in FL, introducing a structured taxonomy of attack types and defense strategies, along with a critical evaluation of their effectiveness, limitations, and open research challenges. \\[1.5pt] 
\hdashline

\addlinespace[6pt] 

\citet{blockchain_survey_fl} & Blockchain & Comprehensive survey of Blockchain-based FL, presenting taxonomy architectures, analysing their integration across privacy-sensitive domains, and highlighting blockchain’s role in enhancing FL security and decentralization. \\

\midrule 

\multicolumn{3}{l}{\textbf{Trustworthiness}} \\

\citet{model_aggregation_survey} & Aggregation Strategies & Conducts a systematic review of model aggregation techniques in federated learning, highlighting their impact on global model accuracy and reliability, and analyzing methods designed to handle low-quality or malicious client updates. \\[1.5pt] 
\hdashline

\addlinespace[6pt] 

\citet{li2023interpretable} & Interpretability & Proposes a novel taxonomy for interpretable federated learning, covering methods that enhance prediction explainability, enable model debugging, and attribute contributions to data owners, crucial for fair reward allocation. Includes a comprehensive analysis of current approaches, evaluation metrics, and future research directions.  \\[1.5pt] 
\hdashline

\addlinespace[6pt] 

\citet{BayesianFLSurvey} & Probabilistic approaches & Addresses challenges such as data scarcity, heterogeneity, uncertainty, and limited explainability by covering Bayesian frameworks designed for robust learning in real-world federated settings. \\

\midrule 

\multicolumn{3}{l}{\textbf{Personalisation}} \\

\citet{personalized_survey} & Personalisation methods & Presents a structured taxonomy of personalisation strategies, grounded in key challenges and motivations, and offers a comprehensive analysis of existing techniques. Outlines future directions in architectural design, benchmarking, and trustworthy personalised FL. \\

\midrule 

\multicolumn{3}{l}{\textbf{Applications}} \\

\citet{fed_survey_iot} & IoT and application domains & Covers how FL can be applied across various IoT services and applications. It also offers a taxonomy and key takeaways to guide future FL-IoT integration efforts and addresses ongoing challenges in the field.\\[1.5pt] 
\hdashline 

\addlinespace[6pt] 

\citet{federatedRL_survey_2023} & Reinforcement Learning & Comprehensive review of Federated Reinforcement Learning in IoT applications, analysing current solutions across domains such as security, efficiency, and vehicular systems, while identifying research gaps and future challenges. \\

\bottomrule 
\end{longtable}

%% file: survey_sections/formulation.tex
\section{Preliminaries for Federated Learning}
\label{sec:preliminaries}
\subsection{Notation, Terminology and Key Concepts}
\label{sub_sec:Notation}

In Federated Learning (FL), a federation consists of a set of clients or nodes, denoted as $C \coloneq \{c_1, ..., c_N\}$. Each client is represented as $c$, where $c \in C$. Generally, the core mechanism of FL involves a global optimisation algorithm and local optimisation algorithms operating on locally available data. Typically, only a subset of federation nodes participates in training during each global communication round; this subset is referred to as a \textbf{cohort} of size $S$.

\subsubsection{Data \& Client Partitioning}
In a distributed Machine Learning paradigm, data is obtained from multiple computational nodes, such as Internet of Things (IoT) devices (e.g., CCTV cameras, sensors, smartphones) and organizational servers. The global data of a distributed task is denoted as $D$, while the local data of a client $c$ is represented by $D_c$, containing $M_c$ samples. Formally, this can be expressed as:
\begin{equation}
\label{eq:local_dataset}
    D \coloneq \bigcup_{c=1}^{N} D_{c}, \hspace{1cm}
    D_c \coloneq \left\{\left(x_{i}, y_{i}\right)\right\}_{i=1}^{M_c}.
\end{equation}

where $x_i$ and $y_i$ denote the features and target samples, respectively, from each client. We denote \( |D| \coloneq M \) and \( |D_c| \coloneq M_c \) as the total number of samples across all clients and the total number of samples for a single client, respectively.

In distributed systems, data exhibits certain properties that influence \( D_c \). Unlike the centralized setting, where data is typically homogeneous and independently and identically distributed (IID), FL often deals with heterogeneous (non-IID) data. Each local dataset \( D_c \) consists of data points \( (x, y) \) sampled from a client-specific probability distribution \( P_c \). These distributions vary across clients, meaning that for any two clients \( c_a, c_b \in C \) with \( c_a \neq c_b \), we have \( P_{c_a} \neq P_{c_b} \). This distributional shift across clients is a defining characteristic of FL, reflecting the realistic non-IID nature of distributed data.  

\subsubsection{Distributed Communication}
\label{sub_sub_sec:dist_communication}

Following the literature on distributed systems and Federated Learning, several terminologies are commonly used to describe client nodes. One such term is \textbf{stragglers}, referring to clients that are significantly slower than others in completing their local training during a communication round, often becoming a bottleneck in the aggregation process. Another important concept is \textbf{staleness}, which describes the situation where clients perform local training based on outdated global models, leading to the submission of updates that may no longer be relevant or optimal for the current state of the global model.

\paragraph{\textbf{Communication Rounds and Epochs.}}
In FL, the learning process is organized into discrete \textbf{communication rounds}, where clients update their local models and send them to the central server (or other clients in decentralized settings) for aggregation. Each communication round typically involves:
\begin{enumerate}
    \item \textbf{Downlink:} Broadcasting the latest global model from the server to a subset of clients (a \textbf{cohort}).
    \item Local training on client-specific data for a fixed number of \textbf{epochs}.
    \item \textbf{Uplink:} Uploading the locally updated models from clients to the server for \textbf{aggregation}, where client models are merged into a global model.
\end{enumerate}

\paragraph{\textbf{Convergence.}}
Convergence refers to the process by which the global model reaches a stable state, where further communication rounds lead to negligible performance gains. This process is influenced by several key factors. \textbf{Data heterogeneity}, where clients possess non-iid data distributions, can significantly slow down convergence or lead to suboptimal global models. \textbf{Communication constraints}, such as limited bandwidth or unreliable network conditions, may delay the exchange of updates, thus extending the time required to converge. Additionally, \textbf{client participation} plays a crucial role, as not all clients are available in every training round, introducing variability that can disrupt or delay convergence.

\subsubsection{Model Parameters and Optimisation}
\label{sub_sub_sec:model_parameters}

\paragraph{Model Representation.}
The learnable parameters of a Machine Learning model (e.g., a neural network) are denoted by $\theta$. The parameters of a global model in a centralized setting are represented as $\theta$, while client-specific models are denoted by $\theta_{c_1}$, $\theta_c$, and $\theta_N$, representing the first, any general, and the last client, respectively. A neural network is defined as $f$, a universal approximating function parametrised by input data and parameters $\theta$. For a specific client $c$, the neural network is represented as $f_c(x; \theta_c)$, which, given a data point $(x, y) \in D_c$, predicts $\hat{y} \coloneq f_c(x; \theta_c)$. Differentiating $f_c$ from $f$ acknowledges that clients may learn different models, a scenario termed as \textbf{heterogeneous model learning}~\citep{fed_md}.

\paragraph{\textbf{Optimisation and Loss Functions.}}
In Machine Learning, parameters are optimized to minimise a loss error function $\ell\left( y , f\left(x; \theta\right)\right)$, which evaluates the model's performance. Gradient descent algorithms are typically employed to update the parameters iteratively using gradient information $\nabla_\theta l$ (gradient of the loss with respect to the model parameters $\theta$) and a learning rate \( \eta > 0\), which determines the step size of each update. The learning rate controls how much the parameters are adjusted in response to the gradient, balancing the trade-off between convergence speed and stability. Optimisation can be subject to constraints, denoted as $\Theta$, specific to a given task. To express numerical values that control the enforcement of specific mathematical constraints throughout this survey we use $\rho$ and $\lambda$.

\paragraph{\textbf{Probability and Logits.}}
The probability of a random variable $Z$ taking the value $z$ is denoted as $P(Z=z)$. In classification models, the term \textit{logits} refers to the $\log$ probabilities output by a classifier model.

\subsubsection{Aggregation}
\label{sub_sub_sec:aggregation}

A fundamental operation in Federated Learning is \textbf{aggregation}, which combines information from multiple client nodes to update a global representation. This step is most commonly performed by aggregating model parameters, such as in Federated Averaging, where client updates are averaged, typically weighted by local dataset size. However, depending on the problem formulation, aggregation can also be applied to other learning artefacts like gradients, embeddings, enabling flexible strategies that account for data heterogeneity, communication efficiency, or robustness to outliers.

\subsection{Distributed Objective}
\label{sub:global_objective}

A Federated Learning problem can usually be considered an optimisation problem. The problem formulation reflects on each client's objectives and describes how the local clients' objectives are orchestrated together. Examples of how the client's objectives can be mutually related include consensus optimisation (similar clients' objectives)~\citet{FedAvg}, and competitive optimisation (competitive clients' objectives)~\citet{Tang2023CompetitiveCooperativeMR,Wu2021MarSFLEC}. In consensus optimisation, cooperation can occur individually or group-wise (as in distributed clustering learning~\citet{stochastic_clustering_fl}). 
Generally, a distributed learning problem is minimising a global loss/objective function of the form $\mathcal{G}(\mathcal{L}_1(\theta_1),...,\mathcal{L}_N(\theta_N))$.
We further formalise the process of constraining parameters or knowledge artifacts during aggregation using an \textbf{Alignment operator}, denoted as $\mathcal{A}$. We formalise this taxonomy of alignment operators due to its importance in federated learning, where it appears in various forms across numerous studies (e.g.~\cite{REFC,REFD,REFE}). The objective is to minimize the expected loss across clients while ensuring that the alignment measure $\mathcal{A}$ stays within a predefined threshold $\epsilon$. Additional constraints may affect the parameters ($\theta, \theta_c$) being modelled, as

\begin{equation}
\label{eq:fl_framework}
\begin{aligned}
& \underset{\theta, \{\theta_c\}_{c=1}^N}{\text{minimise }}
& & %
\mathcal{G}\left(\mathcal L_1(\theta_1),...,\mathcal{L}_N(\theta_N), \mathcal{L}(\theta))\right) \\
  & \hspace{5mm}\text{s.t.}
& & \mathcal{A}\left(\theta, \{\theta_c\}_{c\coloneq1}^N\right) \leq \epsilon, \left(\theta, \{\theta_c\}_{c\coloneq1}^N\right) \in \Theta.
\end{aligned}
\end{equation}

Here, $\mathcal{L}_c(\theta_c)$ denotes the expected loss of client $c$ where the expectation is taken for the client $c$ data distribution; and $\mathcal{L}(\theta)$ represents an optional term that corresponds to the expected loss concerning the joint clients' data distribution. Also, $\theta$ may be interpreted either as a full global model of interest to all clients or as a part of the model (e.g. a subset of Neural Network parameters) that is of interest to all clients. Next, $\mathcal{G}$ represents an aggregation function that specifies how the different local losses are mutually combined, and $\mathcal{A}$ represents an alignment function that specifies how the models $\theta_1,...,\theta_N$ and $\theta$ are mutually related. $\mathcal{A}$ returns a vector (multi-dimensional) output, then the inequality $\mathcal{A} \leq \epsilon$ is interpreted component-wise. Finally, $\Theta$ represents an a priori constraint on the space of admissible models/parameters.

Typically, $\mathcal{L}(\theta_c)\coloneq\mathbb{E}_{(x, y) \sim \mathcal{D}_c}\left[\ell\left(y, f_c(x_c ; \theta_c)\right)\right]$ is the expected loss with respect to each client $c$'s data distribution.

Examples of the aggregation function $\mathcal{G}$ include the average and the maximum functions. That is, 
many formulations consider 

\begin{equation}
\mathcal{G}(\mathcal{L}_1(\theta_1),...,\mathcal{L}_N(\theta_N)) \coloneq \frac{1}{N}\sum_{c=1}^N \mathcal{L}_c(\theta_c), 
\end{equation}

or 

\begin{equation}
\mathcal{G}(\mathcal{L}_1(\theta_1),...,\mathcal{L}_N(\theta_N)) \coloneq \max ( \mathcal{L}_c(\theta_c), c\coloneq1,...,N).
\end{equation}

The alignment operator $\mathcal{A}$ can also take many forms, for example:

\begin{equation}
\mathcal{A}(\theta, \{\theta_c\}_{c\coloneq1}^N) \coloneq \|\theta-\theta_c\| \leq \epsilon, c\coloneq1,...,N,
\end{equation}

or 

\begin{equation}
\mathcal{A}(\theta, \{\theta_c\}_{c\coloneq1}^N) \coloneq \|\theta_i-\theta_j\| \leq \epsilon, i,j\coloneq1,...,N,
\end{equation}

\noindent hence enforcing clients' mutually similar models. Note that here $\epsilon$ may also be taken to be zero, thus enforcing full consensus across clients' models. 

It can also take the form of

\begin{equation}
\mathcal{A}(\theta, \{\theta_c\}_{c\coloneq1}^N) \coloneq -\|\theta_i-\theta_j\| \geq \epsilon, i,j\coloneq1,...,N,
\end{equation}

\noindent thus enforcing mutually different models across clients.

Additionally, a combination of the similarity-enforcing and difference-enforcing constraints as above is possible, for distinct sets of client pairs.

Among the many possible formulations of the form~\eqref{eq:fl_framework}, we list some common examples:

\hspace{-10mm}\begin{minipage}{\linewidth}
\begin{tabular}{cc}
\begin{minipage}{0.49\textwidth}
\begin{equation}
\label{eq:A}
\begin{aligned}
& \underset{\{\theta_c\}_{c\coloneq1}^N}{\text{minimise }}
& &
\frac{1}{N} \sum_{c\coloneq1}^N \mathcal{L}_c(\theta_c) \\
  & \hspace{5mm}\text{s.t.}
& & \theta_i=\theta_j, i,j\coloneq1,...,N,
\end{aligned}
\end{equation}
\end{minipage}
&
\begin{minipage}{0.49\textwidth}
\begin{equation}
\label{eq:B}
\begin{aligned}
& \underset{\{\theta_c\}_{c\coloneq1}^N}{\text{minimise }}
& & \max (\mathcal{L}_c(\theta_c), c\coloneq1,...,N) \\
  & \hspace{5mm}\text{s.t.}
& & \theta_i=\theta_j, i,j\coloneq1,...,N,
\end{aligned}
\end{equation}
\end{minipage}
\end{tabular}
\end{minipage}

\hspace{-5.5mm}\begin{minipage}{\linewidth}
\begin{tabular}{cc}
\begin{minipage}{0.48\textwidth}
\begin{equation}
\label{eq:C}
\begin{aligned}
& \underset{\theta, \{\theta_c\}_{c\coloneq1}^N}{\text{minimise }}
& & \frac{1}{N} \sum_{c\coloneq1}^N \mathcal{L}_c(\theta_c) + \frac{\rho}{2}\sum_{c \coloneq1}^N \|\theta_c-\theta\|^2,
\end{aligned}
\end{equation}
\end{minipage}
&
\begin{minipage}{0.48\textwidth}
\begin{equation}
\label{eq:D}
\begin{aligned}
& \underset{\theta, \{\theta_c\}_{c\coloneq1}^N}{\text{minimise }}
& & \frac{1}{N} \sum_{c\coloneq1}^N \mathcal{L}_c(\theta_c) + \frac{\rho}{2}\sum_{c\coloneq1}^N \|\theta_c-\theta\|,
\end{aligned}
\end{equation}
\end{minipage}
\end{tabular}
\end{minipage}

\begin{equation}
\label{eq:E}
\hspace{-3mm}\begin{aligned}
& \underset{\{\theta_c\}_{c=1}^N}{\text{minimise }}
& & \frac{1}{N} \sum_{c\coloneq1}^N \mathcal{L}_c(\theta_c) \\
  & \hspace{5mm}\text{s.t.}
& & \|\theta_i-\theta_j\|  \leq \epsilon,
\end{aligned}
\end{equation}

Formulation~\eqref{eq:A} appears in many references and is perhaps most widely used currently~\citep{FedAvg}. 
Formulation~\eqref{eq:B} appears in distributed robust (model agnostic) Federated Learning~\citep{REFB}. Formulations~\eqref{eq:C}, \eqref{eq:D}, \eqref{eq:E} are examples that arise in Personalised Federated Learning~\citep{REFC,REFD,REFE}.

Equation~\eqref{eq:fl_framework} can also be reformulated as an unconstrained problem, for example, as observed in Eq.~\eqref{eq:C}, e.g. using a quadratic penalty method as
\begin{equation}
\label{eq:fl_framework_math}
\begin{aligned}
& \underset{\theta, {\{\theta_c\}_{c=1}^N}}{\text{minimise }}
& & \frac{1}{N}\sum_{c\coloneq1}^N \mathcal{L}_c(\theta_c) + \rho\sum_{c\coloneq1}^N g(\mathcal{A}\left(\theta, \theta_c\right))\
& \hspace{2mm} 
\\ & \hspace{5mm} \text{where} & & g(\mathcal{A}\left(\theta, \theta_c\right)) \coloneq \max(-(\mathcal{A}\left(\theta, \theta_c \right)-\epsilon), 0)^2.
\end{aligned}
\end{equation}

The solutions \eqref{eq:fl_framework_math} and Eq.~\eqref{eq:C} can be related; Typically they are mutually close or equal when $g$ is large enough.

Locally, at each client, the population loss $\mathcal{L}_c(\theta_c)$ is unknown. Therefore, it is replaced by an approximation based on the available training data $D_c$ using pairs $(x,y)$, through a sample average:

\begin{equation}
    \label{eq:local_opt_function}
    \minimize _{\theta_c} \mathcal{L}_c\left(\theta_c\right),
    \hspace{1cm}
    \mathcal{L}_c(\theta_c)\coloneq\frac{1}{|D_c|}\sum_{(x^i,y^i) \in D_c} \ell\left(y^i_c, f_c(x^i_c ; \theta_c)\right).
\end{equation}

In heterogeneous Federated Learning environments, clients may exhibit diverse data distributions, leading to challenges in global model convergence. To address this, clustering similar clients can ease optimisation by enabling separate global models for distinct client groups. This motivates the Clustered Federated Learning (CFL) problem, which involves optimising multiple global models corresponding to client clusters.

The CFL problem naturally leads to a dual optimisation task involving:

\begin{enumerate}
    \item Determining the optimal cluster memberships \( \{\mathcal{C}_k\}_{k \coloneq 1}^{K} \), where \( \mathcal{C}_k \) denotes the set of clients in the \( k \)-th cluster;
    \item Optimising the global model parameters \( \{\boldsymbol{\theta}_k\}_{k \coloneq 1}^{K} \) for each cluster.
\end{enumerate}

These two components can be optimised jointly or in an alternating fashion.

We define the model parameter optimisation objective as:

\begin{equation}
\label{eq:cfl_param_opt}
\min_{\{\boldsymbol{\theta}_k\}_{k=1}^{K}} \sum_{k=1}^{K} \sum_{c \in \mathcal{C}_k} \frac{1}{|D_c|} \sum_{(x^i, y^i) \in D_c} \ell\left(y^i_c, f_c(x^i_c ; \theta_c)\right)
\end{equation}

In this formulation, \( K \) denotes the number of clusters, and \( \boldsymbol{\theta}_k \) represents the global model parameters associated with the \( k \)-th cluster. The function \( \ell(\cdot , \circ) \) corresponds to the local loss evaluated over each client's data.

The cluster membership assignment objective, given current global models, can be formulated as:

\begin{equation}
\label{eq:client_clustering}
\min_{\{\mathcal{C}_k\}_{k \coloneq 1}^{K}} \sum_{k \coloneq 1}^{K} \sum_{c \in \mathcal{C}_k} d\left( \boldsymbol{\theta}_c, \boldsymbol{\theta}_k \right)
\end{equation}

Here, \( \boldsymbol{\theta}_c \) denotes the local model update of client \( c \), and \( \boldsymbol{\theta}_k \) is the global model representing cluster \( k \). The distance metric \( d(\cdot, \circ) \) can reflect dissimilarity in parameter space, gradient updates, or model outputs, and is often instantiated as the Euclidean distance, cosine similarity, or KL divergence.

Clustering is typically performed using distance-based algorithms such as \textit{k}-means or spectral clustering. By integrating cluster structure into the federated setting, CFL captures non-IID distributions more effectively, leading to improved personalisation and convergence in practice.

\subsubsection{From Problem Formulation to Algorithmic Solution}
\label{sub:local_objective}
Building on problem formulations from the Federated Learning (FL) literature, we formalize how these optimisation problems are practically addressed through algorithmic solutions.

The foundational algorithm of FL is FedAvg~\citep{FedAvg}, an iterative algorithm within a server-client centralised schema (See Fig.~\ref{fig:star_schema_fl}). In this framework, the server aggregates optimal parameters from the local client models $\theta_c$ after each communication round, by averaging them, as shown in the following update rule:
\begin{equation}
    \label{eq:fedAvg}
    \theta \leftarrow \sum_{c\coloneq1}^N \frac{|D_c|}{|D|} \theta_c, \hspace{1cm}
    \theta_c^{s+1} \leftarrow \theta_c^{s}-\eta \nabla_{\theta_c^{s}} l\left(\theta_c^{s}\right).
\end{equation}
Eq.~\eqref{eq:fedAvg} represents both the aggregation on the server and the local optimisation on the clients using Stochastic Gradient Descent, which updates parameters based on their gradient information at each local step $s$.

The server-client Federated Learning algorithm FedAvg can be outlined as follows:
\begin{enumerate}
\item \textbf{Initialisation:} Initialisation of the global model following standard practices or task-related heuristics;

\item \textbf{Client Selection:} Based on predefined heuristics, such as computational resources and communication quality, certain client nodes, referred to as a cohort, are selected to participate in more robust training rounds;

\item \textbf{Server Broadcast:} The global model's parameters, along with relevant statistical data and other forms of knowledge such as pre-trained features or model insights, are shared with selected client nodes to support their local training;

\item \textbf{Client Training:} Local nodes perform $Z$ optimisation steps until convergence or a specific stop criterion;

\item \textbf{Clients' Broadcast:} Local models parameters $\theta_c$, along with any other client-specific knowledge, are shared with the server to enable parameter aggregation;

\item \textbf{Aggregation:} The server collects the local client updates and averages them to create an updated version of the global model;

\item Repeat steps 2-6 for $Y$ rounds or until a specific stop criterion.
\end{enumerate}

To illustrate a more complex scenario, consider the Clustered Federated Learning (CFL) algorithm, which proceeds in iterative communication rounds:

\begin{enumerate}
    \item \textbf{Initialisation:} Initialise the global model for each cluster. Initially, clients may be randomly assigned to clusters.
    
    \item \textbf{Cluster Broadcast:} Each cluster broadcasts its global model parameters, along with any relevant statistics or shared knowledge (e.g. pre-trained features), to its assigned clients.
    
    \item \textbf{Local Training:} Each client performs local training on its private data and computes an updated model \( \boldsymbol{\theta}_c \).
    
    \item \textbf{Client Upload:} Clients send their local model updates \( \boldsymbol{\theta}_c \), and optionally any client-specific statistics or metadata, back to the server.
    
    \item \textbf{Cluster Reassignment:} Clients are reassigned to clusters based on a distance metric \( d(\cdot, \circ) \), typically comparing local updates to cluster centres.
    
    \item \textbf{Cluster Aggregation:} For each cluster, the server aggregates the local model updates from its assigned clients to update the cluster-specific model \( \boldsymbol{\theta}_k \).
    
    \item \textbf{Global Synchronisation:} Cluster centres \( \boldsymbol{\theta}_k \) are iteratively refined to minimise the overall CFL objective.
\end{enumerate}

The FedAvg straightforward training process becomes more complex when incorporating local model alignment penalties, modifying the alignment algorithm, or enforcing stronger privacy and security guarantees. While FedAvg, as defined in Eq.~\ref{eq:fedAvg}, is designed to solve Eq.~\ref{eq:A}, adapting it to solve other relevant formulations, such as Eqs.~\eqref{eq:B}-\eqref{eq:E}, requires non-trivial modifications to the algorithm.

\subsubsection{Convergence}

Convergence in optimisation refers to the process by which an algorithm iteratively refines its solution, ideally approaching a local or global optimum. In classical optimisation, convergence analysis is centered on factors such as the loss landscape, step size (learning rate), and initialization. The speed and accuracy of convergence determine the efficiency of an algorithm, particularly in large-scale, non-convex settings where slow or premature convergence can hinder optimal solutions.

In Federated Learning (FL), convergence is even more intricate due to distributed optimisation, data heterogeneity, and communication constraints. FL aims to optimize a global objective function $\mathcal{G}$ through local updates on distributed clients, and its convergence properties are affected by assumptions on $\mathcal{G}$, client data distributions, aggregation and alignment strategies. Notably, the distinction between independent and identically distributed (IID) vs. non-IID data profoundly impacts convergence behavior, often necessitating algorithmic adaptations to maintain stability and efficiency.

An overview of convergence in Federated Learning is examined in~\cite{convergence_fl_theory}. In this work, we highlight the key aspects of FL algorithms and their convergence properties.

\paragraph{\textbf{Function Classes and Their Convergence Properties.}}  
The convergence properties of FL algorithms depend significantly on the formulation choices, such as Eqs.~\eqref{eq:B}--\eqref{eq:E}, and on the function class of the global objective. If the objective function \( f \) is convex, standard gradient-based methods generally ensure convergence, with sublinear rates depending on smoothness and strong convexity assumptions. When \( f \) is strongly convex, convergence guarantees are stronger, often achieving linear rates under appropriate learning rate schedules. Most modern FL applications involve non-convex objectives, such as those arising in deep neural networks. In this setting, convergence analysis typically focuses on reaching stationary points rather than global optima~\citep{stationary_points_conv_analysis}. Nevertheless, if \( f \) satisfies the Polyak-Łojasiewicz (PL) condition\footnote{A mathematical property that helps ensure that an optimisation algorithm converges efficiently. It states that if a function satisfies this condition, the gap between its current value and the optimal value can be used to bound the gradient norm.}, convergence to a global minimum can still be achieved at a linear rate, despite non-convexity, thereby mimicking the behavior of strongly convex functions.

\paragraph{\textbf{Impact of Data Heterogeneity on Convergence.}}  
A fundamental challenge in FL is data heterogeneity, which significantly influences convergence rates. When data is independently and identically distributed (IID) across clients, FL methods tend to achieve convergence rates similar to centralised settings, as the problem is homogeneous and gradient updates from different clients align well~\citep{FedAvg}. In contrast, practical FL settings often involve non-IID data, which leads to optimisation difficulties. To address this, some works assume bounded local loss minimisers imposing an upper bound on the divergence of local optima across clients to facilitate convergence guarantees~\cite{bounds_fl}. Other approaches model clustered heterogeneity, assuming structured variability where clients belong to underlying clusters, thereby improving optimisation strategies~\cite{clustering_fl_efficient}.

\paragraph{\textbf{Optimisation Techniques for Improved Convergence.}}  
To address the challenges posed by data heterogeneity and improve convergence in FL, various optimisation approaches have been proposed. Federated Averaging (FedAvg) remains the standard baseline method in which clients perform local updates before aggregation. However, FedAvg lacks convergence guarantees under non-IID data~\cite{fed_avg_problem_non_iid}. To mitigate this, optimisation-based methods have been developed that leverage historical information to improve client updates. These include momentum-based FL approaches~\citep{xu2021fedcm, wang2019slowmo, das2022faster}, and proximal methods such as FedProx~\citep{FedProx} and SCAFFOLD~\citep{SCAFFOLD}. Operator splitting methods (e.g. FedSplit~\citep{fedsplit}) has also been suggested to address the issue of incorrect fixed points in earlier optimization methods, ensuring that the iterates converge to the true minimizer in convex federated settings. Client selection techniques have also emerged, strategically choosing well-behaved clients to enhance global model training and reduce divergence issues~\citep{power_of_choice, contra}. Additionally, employing pre-trained models and transfer learning has proven effective in providing better initialisation, thereby improving convergence efficiency.

\paragraph{\textbf{Linear Speedup and Scalability.}}  
An ideal FL algorithm should exhibit linear speedup, where the loss error scales as \( O(1/N) \) with the number of clients \( N \). This is typically achievable when data is IID, as gradient noise diminishes symmetrically across clients, enabling near-optimal speedup. However, in heterogeneous data settings, convergence degrades due to inconsistent updates across clients. To overcome this, several techniques such as adaptive weighting, variance reduction, and personalised FL have been proposed~\citep{stochastic_clustering_fl}. Furthermore, even when theoretical speedup is achievable, practical performance may be limited by communication overhead. This has led to the development of strategies such as local updates, adaptive aggregation, and communication-efficient methods like compression~\citep{no_public_kd, function_space_fl}.

\paragraph{\textbf{Stochastic Gradient Noise Assumptions.}}  
FL convergence is also influenced by assumptions regarding stochastic gradient noise. A common assumption is that of zero-mean noise, ensuring unbiased convergence~\citep{zero_mean_assumption_convergence_analysis}. Techniques like Differential Privacy, which introduce randomly generated zero-mean noise, are used to enhance privacy in FL~\citep{DP_genesys}. Another assumption is that of uniformly bounded variance, which guarantees controlled convergence rates under certain conditions~\citep{uniform_variance}. More recent work has explored FL under heavy-tailed noise distributions, prompting the use of robust estimation techniques such as gradient clipping and adaptive learning rates~\citep{robust_over_the_air_noise}.

\begin{notebox}
For further details, we refer the reader to the optimisation Algorithms subsection in the survey~\cite{fl_open_problems} and to an overview of Federated Learning convergence in~\cite{convergence_fl_theory}.
\end{notebox}

\subsection{Taxonomy of Federated Learning Systems}
\label{sub_sec:fl_flavour}
The structure and capabilities of participating clients define the type of Federated Learning architecture. Federated Learning can be broadly categorized into Cross-Silo, Cross-Device, and Hierarchical Federated Learning, each tailored to distinct use cases. In a \textbf{Cross-Silo} federation, multiple organisations such as hospitals collaborate, each represented by a robust client with reliable communication.

\begin{figure*}
    \centering
    \includegraphics[clip,width=0.7\linewidth]{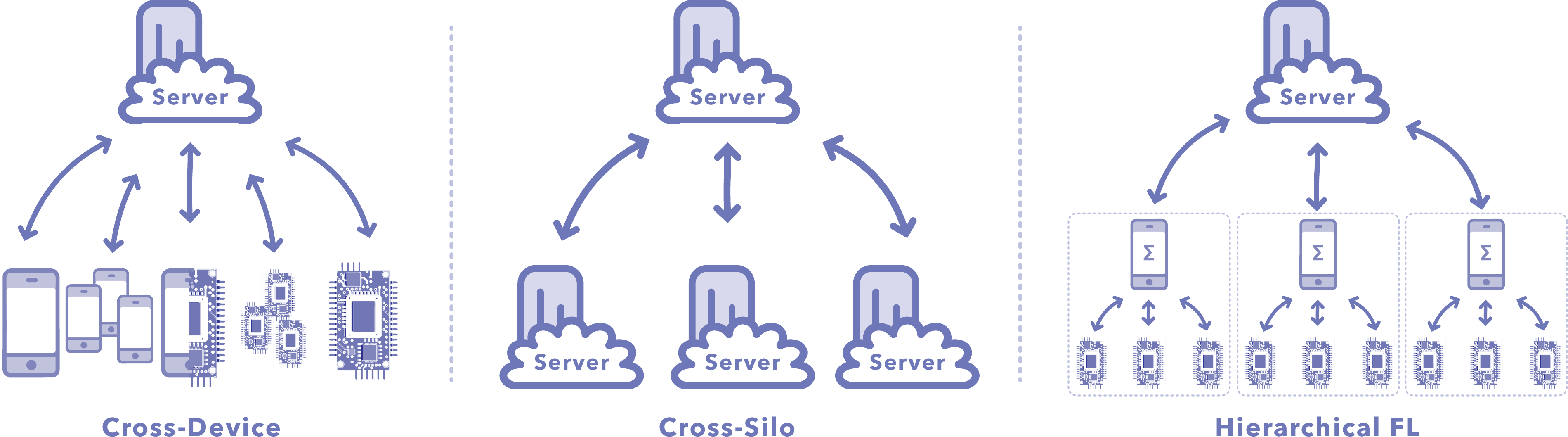}
    \caption{
    Contrasting different infrastructure meta-types of Federated Learning architectures. Cross-Device FL utilises low-resource devices with uncontrollable training availability (e.g. smartphones, microcontrollers). Cross-Silo is based on a reliable infrastructure such as a collection of servers. Hierarchical FL mixes Cross-Device and Cross-Silo FL infrastructure.
    }
    \label{fig:meta_infra_flavours}
\end{figure*}
Conversely, in a \textbf{Cross-Device} federation, a central server orchestrates learning with less powerful clients (e.g. IoT or phone devices), which may be intermittently available or compromised. Training occurs when the device is idle, charging, and connected to a robust network.

\textbf{Hierarchical Federated Learning (H-FL)}, also known as Hybrid Federated Learning or Fog Learning, combines Cross-Device and Cross-Silo FL. Individual silos perform internal aggregations across low-power nodes, with further inter-silo aggregation, avoiding raw data sharing. Fig.~\ref{fig:meta_infra_flavours} illustrates the differences between Cross-Device, Cross-Silo, and Hierarchical FL.

\begin{figure}
    \centering
    \includegraphics[clip,width=0.7\linewidth]{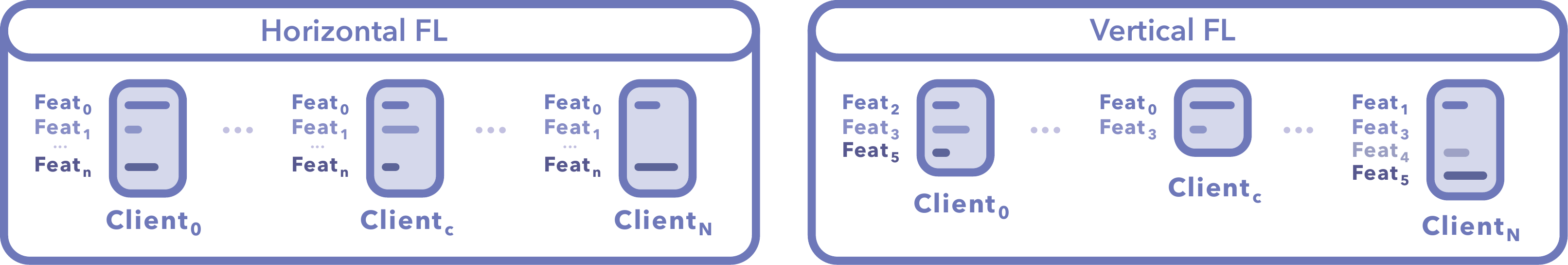}
    \caption{
    Abstract overview of the main differences between Horizontal and Vertical Federated Learning. Horizontal Federated Learning (left) expects the same features across clients, while Vertical Federated Learning (right) involves clients with different features.
    }
    \label{fig:fl_flavour}
\end{figure}
The organization of data across clients determines the type of Federated Learning framework, broadly classified into Vertical (Heterogeneous) and Horizontal (Homogeneous) Federated Learning. In \textbf{Vertical (Heterogeneous) FL}, data is distributed such that different clients hold different features for the same users. For example, in a hospital group, each clinic may provide different clinical features for the same patient.

\textbf{Horizontal (Homogeneous)} FL, on the other hand, involves clients with the same set of features for different users. For instance, different clinics in the same hospital group may produce identical types of features for patients. Fig.~\ref{fig:fl_flavour} contrasts horizontal and vertical FL.

\subsection{Meta-Framework Perspective}
\label{sub_section:mod_toolbox}

In this survey, we propose viewing Federated Learning (FL) from a \textbf{Meta-Framework} perspective, comprising eight key modules (Fig.~\ref{fig:meta_fl_overview}). 

A \textbf{meta-framework} is a higher-order conceptual structure that defines a flexible, modular, and composable system rather than prescribing a rigid methodology. Unlike the traditional framework term, which defines a predefined pipeline or architecture, a meta-framework allows for the systematic composition of multiple interoperable components, enabling customization and adaptation to different requirements and constraints. By defining a structured set of modules that can be configured and extended, a meta-framework serves as both a unifying theory and a practical design guide.

FL aligns with this meta-framework paradigm. Rather than being a singular, fixed approach, FL is an umbrella term encompassing a diverse set of strategies for enabling distributed learning across dispersed data sources. Different FL applications require distinct solutions depending on constraints related to infrastructure, communication, data characteristics, aggregation, security, privacy, and trustworthiness. Viewing FL as a meta-framework allows us to decompose it into a set of fundamental modules, each addressing a specific dimension of FL. These modules can mostly be independently designed, improved, and recombined, making FL more adaptable to various deployment scenarios, from cross-device learning in edge computing to cross-silo collaboration among enterprises.

By embracing this perspective, we shift the understanding of FL from a monolithic paradigm to a structured, modular ecosystem, where different FL techniques and methodologies can be composed like building blocks. This enables scalability, cross-domain applicability, and a systematic approach to analyzing, designing, and implementing FL systems, ultimately leading to more robust federated learning solutions.

We briefly introduce each module:

\begin{figure}
    \centering
    \includegraphics[clip,width=0.7\linewidth]{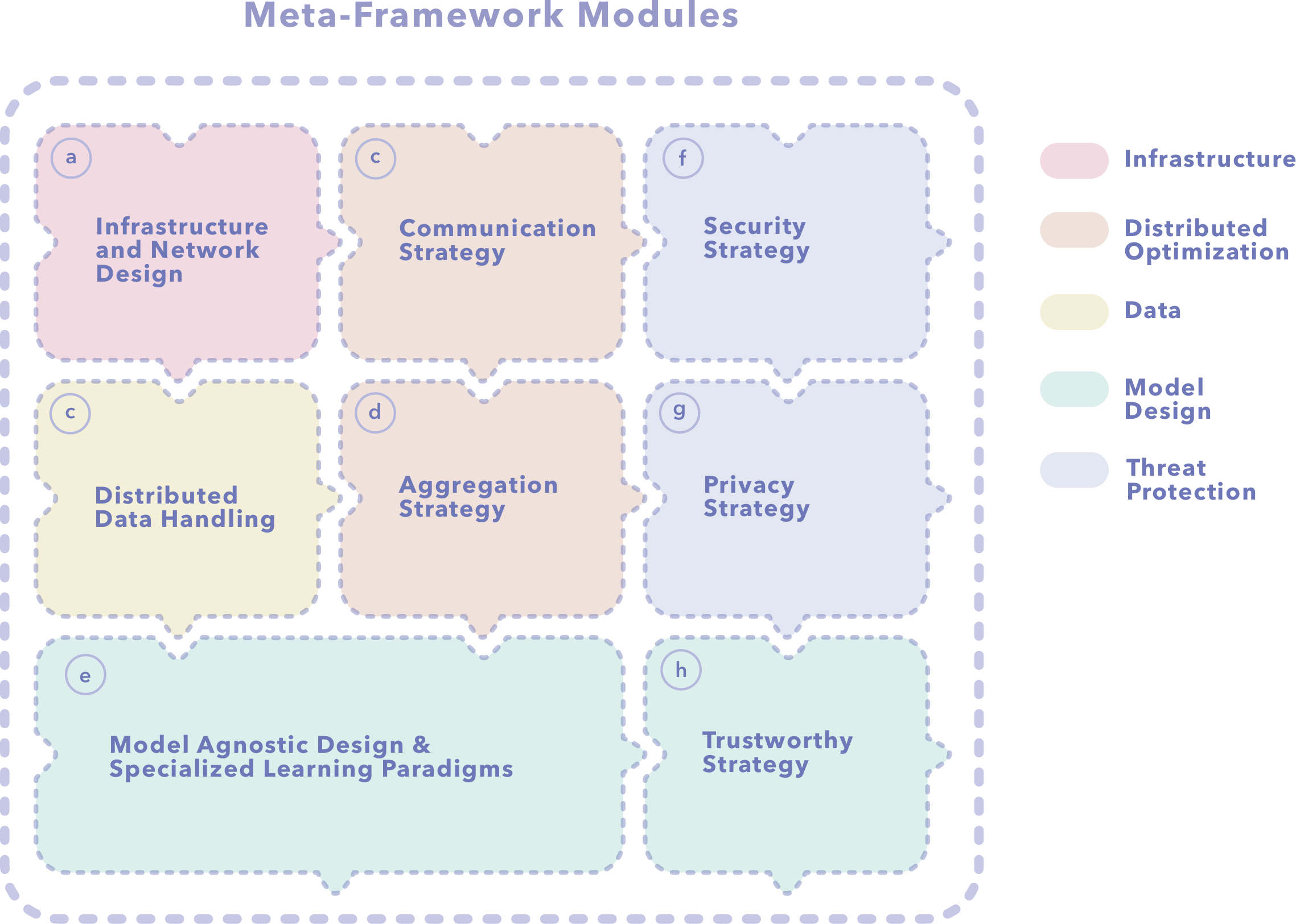}
    \caption{
    Meta-Framework overview of the Federated Learning paradigm as a fusion of key modules categorised into Infrastructure, Data, Threat Protection, Distributed Optimisation, and Model Design.
    }
    \label{fig:meta_fl_overview}
\end{figure}

\begin{enumerate}[a)]

    \item \textbf{Infrastructure and Network Design}: This module focuses on the distributed nature of FL systems, addressing network resources, communication fidelity, and the physical topology of the distributed network. Applications with inherently distributed data, such as healthcare, smart agriculture, or energy routing, necessitate custom infrastructure designs. The network architecture can be modularly adjusted based on FL system types, device constraints and multiplicity, and organisational participation.

    \item \textbf{Communication Strategy}: This module examines the communication protocols underpinning Federated Learning networks, focusing on both \textbf{centralised} (star-schema) and \textbf{decentralised} (peer-to-peer) architectures. In centralised networks, a server orchestrates the coordination and aggregation of updates, while decentralised networks distribute responsibilities among clients, enabling fully collaborative learning without a central authority. Communication artefacts, such as neural network weights, statistics, gradients, and logits, are key to facilitating efficient knowledge exchange and collaboration. Furthermore, the synchronicity of communication is explored, addressing the trade-offs between synchronous and asynchronous approaches in terms of efficiency, scalability, and convergence.

    \item \textbf{Aggregation Strategy}: This module addresses the central role of aggregation in combining the learning of multiple client nodes. Aggregation strategies are critical for balancing contributions from diverse clients, managing data heterogeneity, and ensuring convergence in distributed learning systems. Furthermore, by integrating an \textbf{alignment operator} into the optimisation process, this module introduces a richer and more constrained optimisation landscape, enabling better coordination and knowledge sharing across client models.

    \item \textbf{Distributed Data Handling}: This module addresses the challenges of managing distributed data in Federated Learning. We have categorised datasets and benchmarks commonly used to evaluate FL systems to ease FL evaluation. While real-world distributed datasets are naturally partitioned across clients, benchmark datasets require synthetic partitioning to simulate FL settings. Techniques such as \textbf{Sharding} and \textbf{Dirichlet Distribution-based partitioning} are commonly used to distribute data among clients meaningfully. Additionally, this module incorporates noisy learning considerations, highlighting the importance of benchmarks that account for label corruption and adversarial perturbations to evaluate FL systems under realistic conditions.

    \item \textbf{Model Agnostic Design \&
Specialized Learning Paradigms}: This module highlights the flexibility of Federated Learning (FL) as both a distributed optimisation framework and a distributed systems paradigm, enabling it to adapt to diverse models and tasks. FL’s \textbf{model-agnostic} nature allows for various optimisation strategies tailored to local client models, including probabilistic initialisations, global parameter sharing, and transfer learning. Beyond traditional supervised learning, FL supports specialised paradigms like \textbf{Reinforcement Learning}, which facilitates collaborative policy optimisation in dynamic environments such as autonomous driving, energy management, and robotics. By integrating domain-specific requirements while maintaining privacy and scalability, this module underscores FL’s adaptability and modularity in addressing real-world challenges.

    \item \textbf{Security Strategy}: This module focuses on safeguarding FL systems against a wide range of security threats. These threats include:
\begin{itemize}
    \item \textbf{Data Poisoning:} Altering training data to manipulate the global model.
    \item \textbf{Adversarial Attacks:} Modifying inputs to affect system behaviour during inference.
    \item \textbf{Model Theft:} Hijacking local or global models.
    \item \textbf{Physical Attacks:} Tampering with FL physical nodes to disrupt learning.
\end{itemize}
Mitigation strategies include Blockchain, Trusted Execution Environments, input validation, and Secure Multi-Party Computation, ensuring the robustness of FL systems.

    \item \textbf{Privacy Strategy}: This module emphasises privacy preservation in Federated Learning (FL) systems. While FL inherently offers intrinsic privacy by avoiding raw data transmission, advanced techniques further enhance protection against potential exploitation of communication artefacts. 

    Privacy threats include:
\begin{itemize}
    \item \textbf{Model Inversion Attacks:} Attempts to reconstruct raw data used for training.
    \item \textbf{Membership Inference Attacks:} Attack designed to check whether some given data was used for training.
\end{itemize}
    
    Notable privacy defensive methods include \textbf{Differential Privacy}, which introduces carefully calibrated noise to shared data to prevent individual information leakage, and \textbf{Homomorphic Encryption}, enabling computations to be performed directly on encrypted data without decryption. Additionally, \textbf{Zero-Knowledge Proofs} provide a robust mechanism for verifying the correctness of computations without revealing the underlying data, ensuring privacy and trustworthiness in sensitive applications. Together, these techniques form a comprehensive framework for safeguarding data privacy in FL systems, even in adversarial settings.

    \item \textbf{Trustworthy Methodologies}: This module addresses the need for trustworthiness~\citep{mucsanyi2023trustworthy} in real-world Federated Learning (FL) deployments. Ensuring \textbf{fairness}, \textbf{robustness to noisy labels}, \textbf{uncertainty estimation} and \textbf{explainability} in predictions are critical to fostering confidence in FL systems and making them viable for practical applications. For example, \textbf{domain-knowledge injection} or \textbf{informed machine learning} techniques can be employed in FL to incorporate expert knowledge or contextual insights into the learning process, improving both interpretability and performance, particularly in specialised fields such as healthcare or finance.

\end{enumerate}

%% file: survey_sections/history_context.tex
\section{Distributed Optimisation - A Historical Perspective}
\label{sec:history}

In this section, we establish a connection between Federated Learning and distributed optimisation/learning techniques, requiring a time travel to the past for historical contextualisation.

The distributed optimisation problem consists of
\begin{equation}\label{eq:distrOpt}
  \minimize_{w \in \mathbb{R}^m} \frac 1N \sum_{n\coloneq1}^N f_n(w)
  \end{equation}
where $N$ is the number of agents or data and computing centres, and for each $n\coloneq1, \dots, N$, function $f_n$ is a local function internal to the agent $n$ and determined by data private to the agent~\citep{Chen2012,nedicReview2018}.

\paragraph{\textbf{The Consensus problem and (sub-)gradient optimisation methods.}}

The consensus problem can be found in many areas of computation, like control and game theory or computer systems~\citep{nedicReview2018}. In a peer-to-peer network, a common task is to compute aggregate measurements, say average file size or CPU usage. In a network consisting of $N$ nodes, each computing node $n$ can easily estimate its average file size $w_n$, and it is connected to other peers over a network without a central coordinator node~\citep{diff_rec_ls}. These nodes have to compute the global file size $w \coloneq \frac 1N \sum_{n\coloneq1}^N w_n$~\citep{deGroot1974,borkar1982asymptotic,tsitsiklis1984problems,Tsitsiklis1984DistributedAD,jadbabaie2003coordination,kempe2003gossip}.This problem can be generalised to the distributed optimisation problem~\eqref{eq:distrOpt}, by considering a consensus objective functions of $w$, such as $f_n(w) \coloneq \frac{1}{2} (w - w_n)^2$, instead of simply computing the mean $w$. Minimizing this function encourages $\omega$ to be close to $\omega_n$, which aligns with the idea of reaching a consensus among multiple agents in a system.

Several (sub-)gradient methods were proposed for problems of the form~\eqref{eq:distrOpt} where $f_n$ are convex functions, with convergence relying on diminishing learning rates~\citep{distributed_subgradient_multi_agent}. In \citet{jakovetic2014fast}, two distributed gradient algorithms were proposed, inspired by the centralised Nesterov gradient descent, well-known for its accelerated convergence properties. The first method introduced was the Distributed Nesterov Gradient, which significantly enhanced the efficiency of communications and computations across the network. The second method, Distributed Nesterov Gradient with Consensus Iterations, added another layer of sophistication but required strongly convex functions, knowledge of the Lipschitz constant of the gradient of the loss, and knowledge about communication graph connectivity.

\citet{Vogels2022BeyondSG} concentrates on studying how sparse averaging in decentralised learning speeds up training, by enabling larger learning rates in situations where clients share the same data distribution.

Decentralised Exact First-Order Algorithm (EXTRA,~\citet{shi2015extra}) achieves exact convergence to the solution of the consensus problem. EXTRA is designed to operate with a fixed large step size, which remarkably does not depend on the network size, and it ensures synchronised iteration across all nodes. Each local variable at agent $n$ uniformly converges to the precise minimiser of the global function. Distributed gradient descent (DGD) methods require diminishing step sizes to converge to an exact minimiser. Both EXTRA and DGD utilise similar mixing matrices and share comparable complexities per iteration. However, EXTRA leverages the gradients from the last two iterations, unlike DGD, which uses only the gradient from the last iteration. This modification in EXTRA not only enhances stability but also improves convergence rates.

Parallel to developments in decentralised optimisation strategies like \citet{jakovetic2014fast} and EXTRA, from \citet{shi2015extra}, significant advancements were also made in the field of incremental gradient methods. One notable contribution was the introduction of SAGA~\citep{defazio2014saga}, an optimisation algorithm that refines the concepts of earlier methods like SAG~\citep{schmidt2017minimizing}, SDCA~\citep{shalev2013stochastic}, and SVRG~\citep{johnson2013accelerating}. These algorithms are known for their efficient incremental gradient updates and fast linear convergence rates.
SAGA stands out due to its comprehensive theoretical framework, which not only improves upon the convergence rates of SAG and SVRG but also extends the applicability of these methods to more complex optimisation scenarios. One of the key features of SAGA is its support for composite objectives, where a proximal operator can be applied to the regulariser part of the objective function. This makes SAGA particularly useful for problems involving regularisation. Further, SAGA can directly handle non-strongly convex problems. It is also adaptive to any inherent strong convexity in the problem, optimising its performance based on the specific characteristics of the objective function. This adaptability makes SAGA a versatile tool in both theoretical and practical settings.

The robust framework introduced by \citet{komodakis2010mrf} for optimising discrete Markov Random Fields (MRFs) through dual decomposition provided a powerful basis for handling complex optimisation problems in computer vision. This method effectively decomposed large optimisation problems into smaller subproblems, which were then solved individually and combined to achieve a comprehensive solution~\citep{rush2012tutorial}. The strength of this approach lies in its ability to tailor subproblem selection and use efficient inference techniques, such as graph-cut methods, to enhance overall algorithm performance and applicability.
Building upon this foundation, the 2022 paper by \citet{su2022convergence} explored the nuances of implementing dual decomposition in more dynamic and less controlled environments typical of real-world distributed systems. Their research addressed the critical aspects of asynchrony and computational inexactness—elements that are often inevitable due to communication delays, packet drops, and the inherent limitations of computational accuracy within agent-based systems.

Primal-dual algorithms are a class of optimisation methods used to solve problems that can be formulated as saddle-point problems. The goal is to find a point that simultaneously minimises a primal objective function and maximises a dual objective function. These algorithms have become essential tools in various fields, such as convex optimisation and Machine Learning. Primal-dual methods are particularly well-suited for Federated Learning due to their ability to naturally handle decomposable structures and constraints. In many cases, primal-dual algorithms can decompose the global optimisation problem into smaller, more manageable subproblems. Each node can independently solve its subproblem, typically involving only its local data, while dual variables are used to enforce consistency across the nodes. Specifically, consider the problem
\begin{equation} \label{eq:problemPD}
    \begin{aligned}
\minimize _{\mathbf{w} \in \mathbb{R}^m} & \sum_{n\coloneq1}^N f_n(\mathbf{w}),\\
\text{s.t.} \hspace{5mm}& g(\mathbf{w}) = b
\end{aligned}
\end{equation}
where $f$ is a possibly non-convex function, $g$ is a function from $\mathbb{R}^m$ to $\mathbb{R}^d$, and $b \in \mathbb{R}^d$. To handle the constraints, we introduce the Lagrangian function $L(\mathbf{w},\lambda) \coloneq \sum_{n\coloneq1}^N f_n(\mathbf{w})+ \langle \lambda,g(x) - b \rangle$, where $\lambda \in \mathbb{R}^d$ are the Lagrange multipliers, also called dual variables. In primal-dual methods, we want to find saddle points, i.e., a pair $(x^\star, \lambda^\star)$ satisfying the inequalities
\begin{equation*}
    L(x^\star,\lambda) \leq L(x^\star, \lambda^\star) \leq L(x, \lambda^\star),
\end{equation*}
for all $x$ and $\lambda$. The inequalities entail that the $x^\star$ maximises the Lagrangian with respect to the primal variable $x$ and $\lambda^\star$ minimises it with respect to the dual variable $\lambda$. 
Chambolle and Pock introduced the Primal-Dual Hybrid Gradient method~\citep{chambolle2011first}. It combines primal and dual updates with additional extrapolation steps to improve convergence properties.
The stochastic version of the Primal-Dual Hybrid Gradient algorithm~\citep{Chambolle2018} introduces randomness in the update steps, either by selecting random subsets of data or variables to update, which can significantly reduce computation time for large-scale problems.
A method covering second-order stationary solutions for non-convex distributed optimisation was presented by \citet{hong2018gradient}. This paper investigates two primal-dual-based algorithms to solve linearly constrained non-convex optimisation problems. These algorithms utilise first-order information and are shown to compute second-order stationary solutions (ss2) with probability one. This is significant as it extends existing results beyond first-order primal-only algorithms, achieving global convergence to ss2 for unconstrained distributed non-convex learning problems over multi-agent networks.
A family of randomized primal-dual block coordinate algorithms named DSCOVR (Doubly Stochastic Coordinate Optimisation with Variance Reduction,~\citet{xiao2019dscovr}) was presented for asynchronous distributed optimisation of large-scale linear models with convex loss functions. The primary goal is to enhance computational and communication efficiency in distributed Machine Learning environments.
RandProx~\citep{condat2023randprox} uses randomized proximal updates. It introduces randomness in updating dual variables or calling the proximity operator, which reduces computational load while maintaining accuracy.

\paragraph{\textbf{Alternating Direction Method of Multipliers (ADMM).}}
\label{par:admm}

The ADMM consists of a distributed optimisation method that tries to solve both a primal and dual problem. The primal problem consists of many local sub-problems characterised by local variables that condition the solution of the local problem. The dual problem involves regularisation variables (constraints) distributed across multiple local sub-problems, which are optimised in parallel by each local optimisation step.

Historically, the ADMM method evolved through some convoluted mutations, sparkling into existence as a result of an inexact implementation of the Augmented Lagrangian method (ALM), which is further known as the method of multipliers~\citep{admm_history}.

Roughly speaking, ADMM is an approximation method to the ALM method, which instead of jointly optimising the primal and dual variables, takes the block coordinate variation and alternately optimises in blocks respectively the primal and dual variables. Mathematically speaking, the classical ADMM optimisation problem can be described as:

\begin{equation}
\begin{aligned}
\minimize _{\mathbf{w} \in \mathbb{R}^n, \mathbf{z} \in \mathbb{R}^a} \{\mathcal{L}(\mathbf{w}, \mathbf{z}, \boldsymbol{\pi}) & :=f(\mathbf{w})+g(\mathbf{z}) +\langle A \mathbf{w}+B \mathbf{z}-\mathbf{b}, \boldsymbol{\pi}\rangle\\
& +\frac{\sigma}{2}\|A \mathbf{w}+B \mathbf{z}-\mathbf{b}\|^2\}
\end{aligned}
\end{equation}

where $\pi$ represents a dual variable (Lagrange multiplier), to enforce the constraint suggested by the equation $Aw+Bz=b$. The $w$ and $z$ represent the variables being optimised, subject to the $f$ and $g$ objectives. The term $\frac{\sigma}{2}\|A \mathbf{w}+B \mathbf{z}-\mathbf{b}\|^2$ penalises deviations from the constraint defined, where $\sigma$ consists of an hyper-parameter controlling the penalty level.

Throughout many decades, the ADMM method has been presented in many flavours: Symmetric ADMM; Fast ADMM, Generalised ADMM, Linearised ADMM, and Stochastic ADMM.

Regarding convex problems over distributed optimisation problems ADMM, has also been proven to have some well-behaved properties (e.g. linear convergence:~\citet{Shi2013OnTL}).

If we eliminate the dual constraints in the ADMM formulation, the resulting optimisation problem closely resembles a standard FL setting. Specifically, the global model in FL can be viewed as a regulariser that guides the local models toward consensus. This perspective highlights how ADMM provides a principled way to enforce agreement between local models while allowing for personalization, a key characteristic of many FL approaches.

In particular, consider the following formulation Eq.~\eqref{eq:admm_fl}, where \( \mathbf{w} \) represents the global model and \( \mathbf{W} \) contains the local models across clients:

\begin{multline}
\label{eq:admm_fl}
\minimize _{\mathbf{w} \in \mathbb{R}^n, \mathbf{W} \in \mathbb{R}^{n \times C}} \left\{\mathcal{L}(\mathbf{w}, \mathbf{W}, \cancel{\Pi}):=\sum_{i=1}^C \ell\left(\mathbf{w}, \mathbf{W}_i, \cancel{\boldsymbol{\pi}_i}\right)\right\} 
\\
\ell\left(\mathbf{w}, \mathbf{W}_i, \cancel{\boldsymbol{\pi}_i}\right) := f_i\left(\mathbf{W}_i\right) + \cancel{\left\langle\mathbf{W}_i - \mathbf{w}, \boldsymbol{\pi}_i \right\rangle} + \frac{\sigma_i}{2} \left\|\mathbf{W}_i - \mathbf{w}\right\|^2.
\end{multline}

By removing the dual variables (denoted by the cancellations), the optimisation objective retains a structure similar to FL, where each client optimises its local objective \( f_i(\mathbf{W}_i) \), and the global model \( \mathbf{w} \) acts as a soft constraint via the quadratic regularization term. This term encourages local models to remain close to the global model but still allows deviation based on local data heterogeneity.

Consequently, ADMM can be seen as a generalized framework that unifies FL with constrained optimisation principles. By introducing or removing dual constraints, one can transition between explicit enforcement of consensus (via Lagrange multipliers) and a more flexible regularization-based approach, as commonly used in FL. This connection provides a valuable perspective through which FL methods can be analyzed and extended using optimisation techniques from ADMM.

\begin{notebox}
For further exploration of ADMM, we invite the reader to read~\citep{admm_survey}.
\end{notebox}

\paragraph{\textbf{Block Coordinate Descent.}}
\label{par:bcd}
Classic Block Coordinate Descent (BCD) scheme, introduced by \citet{block_coordinate_descent}, explores an optimisation algorithm where the variables get dispersed in different nodes, and the iterative progression of its solving can be described as an alternate optimisation step on each individual or block of variables, each block at a time, while the others are locked. In a sense, Federated Learning follows this pattern if we encapsulate the several varieties and flavours it builds upon. In FL, each client is solving a subset of variables (a block) of variables and later on, in the aggregation step, a consensus is obtained.

\paragraph{\textbf{Token/random-walk Optimisation.}}

The Distributed Token/random-walk optimisation method is historically described as an iterative process which permits the optimisation of a set of variables which hop between a graph of distributed nodes and is solved locally using each node's data (\citet{iterates_token} uses the \textit{iterate} term instead of the \textit{token} term:~\citet{incremental_gradient_distributed_token}). The simplest and classical version of the distributed token optimisation method utilises only one token per federation network, meaning that no parallel global model learning is performed. Having this in mind, multiple token approaches to increase learning convergence have been tried using semi-decentralised communication topology ~\citep{multi_token_fl} and using the augmented lagrangian method~\citep{multi_token_fl_augmented}.

\paragraph{\textbf{Clustering.}}

Before Federated Learning was introduced, numerous research had focused on the study of clustering techniques within distributed networks. These studies cover from optimised distributed communication to distributed learning. HEED~\citep{heed} dives into the conceptualisation of a hierarchical clustering technique to reduce energy consumption, whilst producing well-distributed cluster heads.

Distributed Optimisation techniques often resort to a collaboration between all clients/agents, since it is assumed all agents are interested in an equal objective. However, this is not always the case~\citep{mtt_survey} and this cooperation might even damage the optimisation process (consider non-iid data for instance). Clustering can help mitigate this issue by allowing agents to be grouped according to a similar objective, hence helping the convergence~\citep{clustering_diffusion_old}.

\paragraph{\textbf{Federated Learning.}} The FedAvg~\cite{FedAvg} algorithm, marked a significant step in distributed machine learning. However, its design prioritized simplicity and scalability for real-world deployment, overlooking foundational principles from mathematical distributed optimisation. This pragmatic focus led the field to naturally diverge into two complementary directions: one emphasizing the rigorous mathematical underpinnings of distributed optimisation, and the other centering on privacy-aware distributed learning, addressing concerns like user data security and regulatory compliance.

This divergence reflects the dual demands of Federated Learning: the need for efficient solutions to real-world constraints and the importance of theoretical guarantees for optimisation. Over time, subsequent works have increasingly bridged these perspectives, enriching Federated Learning with both practical innovations and mathematical rigor. Notable contributions include FedBCD~\citep{fedBCD}, which adapts block coordinate descent methods to federated settings; Inexact Federated ADMM~\citep{InexactFedADMM}, which integrates Alternating Direction Method of Multipliers (ADMM) into FL; Multi-Token Federated Learning~\citep{multi_token_fl}, which enhances communication efficiency; and Stochastic Clustering FL~\citep{stochastic_clustering_fl}, which introduces clustering techniques to improve training dynamics.

%% file: survey_sections/toolbox.tex
\section{Dissecting the Federated Learning Meta-Framework}
\label{sec:fl_in_detail}

In this chapter, we present the Meta-Framework modules that structure the Federated Learning (FL) paradigm (Section~\ref{sub_section:mod_toolbox}). Consisting of eight key modules, this Meta-Framework taxonomy not only enhances conceptual understanding but also serves as a foundational blueprint for designing FL systems and frameworks.

\subsection{Infrastructure and Network Design}
\label{sub_sec:infrastructure}

The design of a Federated Learning system is shaped by its underlying infrastructure and network architecture, which influence its efficiency, scalability, and robustness. FL operates across diverse environments, from resource-constrained edge devices to high-performance servers, requiring careful consideration of computational resources, connectivity, and communication overhead.  

Key challenges include accommodating heterogeneous devices with varying capabilities, managing resource constraints in low-power systems, and mitigating communication bottlenecks that impact training efficiency. Optimised system design must balance these trade-offs while ensuring seamless coordination and reliable model convergence.  

This section examines the infrastructural foundations of FL, exploring strategies to enhance performance and scalability in real-world deployments.

\subsubsection{Heterogeneous Network Node Requirements}
\label{sub_sub_sec:heterogeneous_requirements}

\paragraph{\textbf{Resource Limitations of Edge Devices.}}
Edge devices, such as microcontrollers, lack the powerful resources available in cloud devices, personal computers, or mobile phones. These devices are typically composed of low-capacity CPUs, limited memory, and low power, often lacking GPUs. RAM is often limited to kilobytes, and CPU speeds are bound to megahertz.

Embedded systems, composed of hardware and specific embedded OSs, can be categorised into three types~\citep{embedded_devices}:
\begin{enumerate}
    \item Type I: Full-functional hardware with a general-purpose OS.
    \item Type II: Hardware with a non-general-purpose OS and computational limitations.
    \item Type III: Slave devices without an OS, such as USB storage devices.
\end{enumerate}

\paragraph{\textbf{Heterogeneity of Embedded Systems.}}
FL solutions must account for the resource heterogeneity of embedded processors, which range from ARM architectures to x86, FPGAs, and AI accelerators. Devices like Arduinos, Raspberry Pis, Nvidia Jetsons, and Google Edge TPUs vary significantly in resource capabilities. Standardised frameworks, such as Edge Impulse~\citep{edge_impulse} and ONNX Runtime~\citep{onnxruntime}, help bridge this gap by enabling model deployment across diverse devices.

\subsubsection{Memory and Resource Constraints}
\label{sub_sub_sec:memory_constraints}

\paragraph{\textbf{Low-Memory Devices.}}
Training deep learning models on devices with stringent memory limitations requires innovative solutions. For instance, \citet{OnDeviceT_low_memory} demonstrates techniques for training models within 256 KB of memory. Traditional frameworks like PyTorch or TensorFlow are unsuitable for such environments due to their significant memory overhead.

\paragraph{\textbf{Bare-Metal Frameworks.}}
Bare-metal frameworks, such as LiteRT~\citep{tensorflow_lite}, enable low-memory neural network inference but lack support for continuous learning. These frameworks are optimised for pre-trained model deployment rather than iterative training, which is essential for FL.

\paragraph{\textbf{Strategies for Resource Efficiency.}}
To address these constraints, several methods are employed:
\begin{itemize}
    \item \textbf{Model Quantisation:} Reduces precision levels (e.g., from 32-bit to 8-bit) to minimise memory consumption, enabling efficient model deployment on resource-constrained devices. Quantisation is particularly effective for reducing communication overhead in FL scenarios~\citep{quantizationRFL}.
    
    \item \textbf{Model Pruning:} Removes non-essential parameters while preserving performance, focusing on simplifying the model structure without significantly impacting accuracy~\citep{attention_prunning}. Pruning can dynamically adjust model complexity to match the capabilities of edge devices~\citep{FedTiny}.
    
    \item \textbf{Sparse Representations:} Utilises sparsification to retain only critical features, reducing the density of model weights and gradients. This approach decreases transmission size during communication rounds and improves scalability in FL settings~\citep{sparsification}.
    
    \item \textbf{Early-Exit Training:} Allows intermediate layers to produce predictions, reducing computational costs and enabling inference to terminate earlier when confidence thresholds are met~\citep{branchynet_early_exit}. This technique is well-suited for time-sensitive applications on low-power devices.
    
    \item \textbf{Split Learning:} Distributes the training process between devices and servers, lowering resource requirements on edge devices by offloading computationally intensive tasks~\citep{split_learning_origin}.

    Split Learning~\citep{split_fl_comparison} is a paradigm to train and infer over Machine Learning models, whilst computing entry and end points of neural network layers on the client side and further moving the intermediate heavy computation to the server side (See Fig~\ref{fig:split_learning}). 
SplitNet~\citep{Kim2017SplitNetLT} extended the neural network splitting idea to a tree-like neural network, where the model is layer-wise split concerning its classes/features to reduce the parameters usage and computation overhead. Split Learning also enhances privacy by transmitting intermediate representations instead of raw data and can also be combined with privacy-security mechanics, such as homomorphic encryption to preserve client data better~\citep{Khan2023SplitWP}.

\begin{figure}[!htbp]
    \centering
    \includegraphics[clip,width=0.5\linewidth]{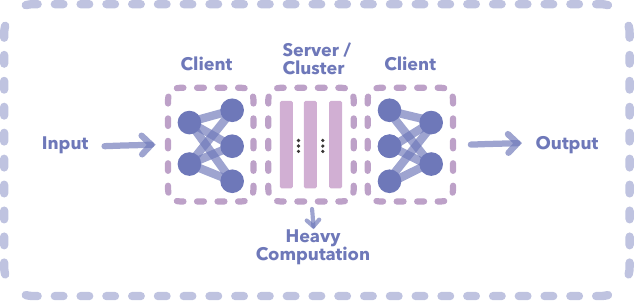}
    \caption{Abstract depiction of Split Learning, which shares a neural network model computational execution between clients and more computational-powered servers.}
    \label{fig:split_learning}
\end{figure}

Even though Split Learning and Federated Learning are different learning paradigms, they share some principles (e.g. privacy-preserving, distributed computing) and can be composed. SplitFed~\citep{Thapa2020SplitFedWF} presents a solution to deal with the downside of the heavy computation of ML on the client side, by relieving the clients' heavy computation to server computation nodes, but still relying on a Federated Learning network of nodes who collaborate their learning.

On the health domain, \citet{split_fl_comparison} utilises Split-NN to showcase the surprisingly better efficiency and the accuracy possible to achieve when compared with default Federated Learning~\citep{FedAvg} and synchronously distributed SGD~\citep{distributed_synchronous_SGD} solutions.

    \item \textbf{Knowledge Distillation:} Facilitates the transfer of knowledge from a larger, more complex model (teacher) to a smaller, simpler model (student)~\citep{ProxyDecentralizedFL}. This technique is particularly effective for addressing data and model heterogeneity in FL, enabling lightweight models to benefit from collaboratively learned knowledge without directly accessing raw data.
\end{itemize}

\subsubsection{Communication Challenges and Protocols}
\label{sub_sub_sec:communication_protocols}

\paragraph{\textbf{Heterogeneous Communication Networks.}}
Devices in FL often connect through diverse networks, including fibre optic, 3G, 4G, 5G, and Wi-Fi~\citep{Niknam2019FederatedLF}.In some cases, the complexity is further increased by the use of analogue over-the-air communication~\cite{over_the_air_ml}. This heterogeneity impacts communication cost and complexity, introducing delays and potential failures in client-server artefact exchanges.

\paragraph{\textbf{Communication Protocols.}}
FL frameworks employ various communication protocols to address these challenges:
\begin{itemize}
    \item \textbf{gRPC:} Designed for real-time streaming and large data loads, supporting synchronous and asynchronous communication~\citep{flower,MetisFL,NVIDIA_FLARE}.
    \item \textbf{AMQP:} Message-oriented middleware enabling features like queues and publish-subscribe mechanisms, suitable for distributed computing~\citep{ibm_fl}.
    \item \textbf{MPI:} A parallel computing protocol ideal for large-scale, CPU-oriented tasks in homogeneous infrastructures~\citep{fedML}.
\end{itemize}

\subsubsection{Learning Scheduling}
\label{sub_sub_sec:learning_scheduling}

FL can employ either \textbf{offline} or \textbf{online} learning schedules:
\begin{itemize}
    \item \textbf{Offline Learning:} Infrequent training sessions with potential model architecture updates between sessions.
    \item \textbf{Online Learning:} Continuous updates to adapt to real-time data changes and counteract data drift, ensuring robustness in production environments~\citep{OnlineFL}.
\end{itemize}

\subsection{Aggregation Strategy}
\label{sub_sec:alignemnt_methods}

Aggregation lies at the core of FL, serving as the mechanism to combine knowledge from multiple clients into a global model. This process determines how local updates are merged and plays a pivotal role in ensuring the efficiency, robustness, and fairness of the FL system. Various aggregation methods have been developed to address challenges such as data heterogeneity, communication constraints, and model convergence. Popular strategies, like FedAvg~\citep{FedAvg}, rely on simple averaging, but more sophisticated techniques introduce client weighting, regularization, or feature-based aggregation to enhance performance and adapt to diverse client environments.

To further refine the aggregation process, additional constraints or guiding principles may be applied. We explore this concept in a dedicated subsection, introducing the alignment operator $\mathcal{A}$ as a complementary tool to constrain and guide aggregation for specific objectives.

The original and most popular algorithm in Federated Learning is the FedAvg algorithm~\citep{FedAvg}, which can be depicted as an optimisation problem solved by minimising the average aggregation function:

\begin{equation}
    \label{eq:aggregation}
    \minimize _{\theta, \{\theta_c\}_{c=1}^N} \mathcal{G}\left(\mathcal{L}(\theta_{1}), \ldots, \mathcal{L}(\theta_N)\right).
\end{equation}

FedAvg averages the local optimised parameters, obtained from each client at round $t$ of the global learning procedure. The method's popularity, despite good results, disregards client data heterogeneous distributions. Consequently, in non-homogeneous data scenarios, standard FedAvg works poorly~\citep{fed_avg_problem_non_iid}.

Relying primarily on the aggregation function $\mathcal{G}$ to optimise a Federated Learning problem is suboptimal, as there is no basis to assume that the average of local client solutions will yield a better model for all clients~\citep{fed_avg_problem_non_iid}. Consequently, the field has begun to explore various aggregation techniques and the communication of diverse knowledge artefacts, beyond local losses and model weights, initially investigated by FedAvg~\citet{FedAvg}.

FedProx~\citep{FedProx} is a generalisation of the FedAvg algorithm to address the heterogeneity of the distributed data. It adds a proximal term to the local client loss to regularise the learning model, encouraging the local model to remain close to the global model.

SCAFFOLD~\citep{SCAFFOLD} stands as an extension to the FedAvg algorithm with a better convergence rate, especially on heterogeneous data and it is invariant to sampling techniques. SCAFFOLD work shows that training for many epochs without communication rounds can deteriorate the global model performance by losing the sensibility to the client's data, also known as client drift. In its essence, SCAFFOLD battles the same issues as FedProx and can be seen as an upgrade to the distributed optimisation algorithm DANE~\citep{DANE}, using stochastic control variables. SCAFFOLD experiments with the idea that commonly clients are disjoint in their distributions, but there can be similarities in some of them, therefore different step sizes should be accountable for better global and local optimum convergence.

Power-of-Choice~\citep{power_of_choice} studied how biasing the selection of clients through their learning loss can improve error convergence. 

Inexact Federated ADMM~\citep{InexactFedADMM} explores a practical federated implementation of the Alternating Direction Method of Multipliers (ADMM) algorithm considering the small distributed sub-problems as the local models being optimised by clients in a federation and overcomes inefficient computations by utilising a cohort of clients instead of the full federation.

Concerned with the client drift phenomenon, FedFish~\citep{function_space_fl} focuses on aggregating local approximations to the functions learnt
by the respective clients, using an estimate based on their Fisher Information, whilst defending the idea that the model aggregation technique used in FL plays a major role in the client drift. They present a genuine metric to evaluate the impact of their aggregation procedure: the Client-Server Barrier.
Given a performance metric, this metric measures on average the distance between the evaluation of the function over the parameter space of the federation local models and the server global model.

All these previous methods mentioned resort to a single server for the aggregation phase, however, there is also an aggregation type that distributes the aggregation phase into several servers (e.g. Prio~\citep{prio}, Prio+~\citep{prio_plus}).

Studying the impact of different federated aggregation algorithms, FedDecorr~\citep{dimensionality_collapse} has shown that under data heterogeneous scenarios FL aggregation schemes lead to greater dimensional collapse. Therefore, FedDecorr introduced a loss regularisation term, which encourages different dimensions of representations to be uncorrelated.

\subsubsection{Personalised Aggregation}
\label{sub_sub_sec:personalization}

In a Federated Learning setting, each client learns a local independent model, which can be seen as a personified compression of local client data. Personalised Federated Learning harnesses the intrinsic nature of individual data while exploring methods for a model to gain a comprehensive understanding of clients' data on a global scale through model sharing. At the local level, however, the approach involves incorporating insights from other clients, ensuring a balance between leveraging collaborative knowledge and prioritizing the integrity of the local data and the local model.

FedFish~\citep{function_space_fl} demonstrates promising post-personalisation results when data is majorly heterogeneous and locally fine-tuned for a few steps after the Federated Learning procedure finishes. They also suggest a way to evaluate personalisation performance: Client Personalisation Performance.
Client Personalisation Performance consists of a metric for assessing personalisation quality, defined by taking the global model previously trained using an FL procedure followed by a fine-tuning and evaluation process over a set of clients over local unseen data. The metric then calculates the average performance of the federated fine-tuning evaluation.

pFedMe~\citep{moreau_envelopes}, similar to FedProx~\citep{FedProx} develops a regularisation method to soften the parameter sharing between clients, by utilising a Moreau Envelope term in the loss function, which controls the weight of the influence of the global model over the local models.

\citet{amazon_noisy_feedback} leverages user feedback (pseudo-labels) to customise local models for individual users while maintaining robustness against noisy feedback.

FedPac~\citep{fed_pac}, uses a feature-alignment procedure to align the local models, where a weighting of local versus a global model is learnt for each client to better adjust locally to the heterogeneous data distribution.

kNN-Per~\citep{knn_personalized_fl} proposes a personalisation
mechanism based on local memorisation, which uses a standard FedAvg algorithm for federated aggregation, but after each local training procedure stores in a local store database the (key: value) pairs of the data samples with the corresponding local model learning representations. Finally, in inference mode, it queries a specific client with an input and based on it, the client computes the k-nearest neighbours over the local dataset and interpolates between the global model and these local intermediate results to obtain the final prediction. Through their experiments, they state that the memorisation technique's usefulness lies greater when distribution shifts coexist between clients.

\paragraph{\textbf{Parameter Decoupling.}}
\label{par:parameter_decoupling}

One technique found in the penalisation take of Federated Learning separates a model into two parts: the knowledge encoding representation sub-model (the body) and the classifier (head). Some research papers suggest training locally the body whilst maintaining the classifier head globally for all clients, others do the opposite.

FedRep~\citep{shared_representation_personalized_fl}, for instance, shares an authentic view of heterogeneous data settings. They transmit the idea, that labels may have different distributions over clients, but the underlying features which encode those labels share similar patterns, a common representation. Consequently, they deeply explore the low-dimensional space understanding as the bare bones of the features, to learn a global learning representation structure, which is personalised at the local/client level. The clients and the server focus on learning a
global representation together, while each client learns their final layer locally conserving the personal properties of the local client data.

Federated Averaging with Body Aggregation and Body Update (FedBABU,~\citet{fedBABU}) explores the parameter decoupling method where the head is locally personalised (finetuned) and the body (representation learning part) is aggregated from the multiple clients.

\subsubsection{Feature-Based Aggregation}
\label{sub_sub_sec:feature_level_alignment}

Heterogeneous environments, where clients in a federation have non-iid data distributions hinder some aggregation methods (e.g. FedAvg). Furthermore, deep learning applications have proven that network bottlenecks are great for encoding meaningful feature representations. Therefore, instead of blindly aggregating network parameters, which not only showcases limitations to non-iid data but also limits models to homogeneous architectures, what if we could map weights meaningfully when aligning client models?

This question sparked several works to try the collaboration of clients' models in between feature representations. FedPac~\citep{fed_pac}, builds upon the FedAvg algorithm but adds a regularisation term penalising the distance between the global feature centroid of a class $y_i$ and the client local feature embedding of the respective data point $x_i$. 

Differently, $fed^2$~\citep{Fed2_alignment} presented a feature-aligned method, where the parameters are combined considering the semantic alignment between them. Each model's weights are grouped by their similar features and are consequently combined through an average operation. In a general sense, $fed^2$ minimises the total feature-level parameter variance among the federation collaborative nodes.

\subsubsection{Knowledge Distillation Aggregation}
\label{sub_sub_sec:knowledge_distilation}

Knowledge Distillation (KD) is a kind of meta-learning algorithm, which assumes the existence of two model types (the teacher, and the student). The teacher normally corresponds to a larger model trained on a task and the student corresponds to a weaker model, which is trained by absorbing the wisdom of the larger model using fewer parameters (model compression,~\citet{model_compression_llm}). Even though knowledge distillation was initially designed as a model compression mechanism, the notion of cooperatively sharing knowledge between homogeneous models to improve performance has gained track (Codistillation,~\citet{co_distillation}).

Applied to FL, the cooperation of teacher-student learning can be thought of as an assembled distillation, in which a global model is a student and several clients are teachers passing on their knowledge to the global model.

The model aggregation operation, typically encountered in the distillation process is the Kullback–Leibler Divergence (KLD), which makes use of the models' prediction logits to approximate the learnt distribution of one model (local model) and the other (global model). This conceptualisation is important since it means that the model predictions must be present in the aggregation server during the models alignment procedure. Furthermore, as the local data is heterogeneous and private (non-present in the aggregation server) most works use a public dataset to ensure the alignment is possible in heterogeneous systems~\citep{fed_md,RobustHFL}.

Since the availability of public datasets, in the FL domain, is scarce, some works have tried to use KD without them~\citep{no_public_kd,fedRad_kd}.

FedRad~\citep{fedRad_kd} introduces relational knowledge~\citep{rel_kd} and single-sample knowledge into the loss function of cooperative distillation optimisation. During the local distillation alignment, it weights each homogeneous model, through predicted entropy to control the penalty of each model loss. The global aggregation can be described as the standard FedAvg, averaging an updated version of each client's global models.

Considering heterogeneous models, proxy model sharing work~\citep{ProxyDecentralizedFL} follows a different paradigm, first by being a decentralised FL solution, secondly by utilising a special homogeneous proxy model (global model: all the clients share the same proxy architecture, but each one has a unique version of it), which is shared directly between clients. Hence, in every training iteration, the local model of each client utilises KD to incorporate knowledge from the proxy model of the other client. In parallel, as the proxy model assimilates knowledge from the local model before returning to its original client, the local model also absorbs the knowledge from the proxy model. This establishes a reciprocal exchange of knowledge between the two models.

\subsubsection{Token-Based Aggregation}
\label{sub_sub_sec:token_alignment}

We have explored techniques to align the shared federation knowledge into one global understanding of the localised data/task. The Token aggregation method is a method where the global model (the token) periodically hops between the federation network, according to some heuristic (e.g. random walk:~\citet{token_random_fl}) and either aggregates the node's local model with the global model or trains the global model directly over the client's data. No server is required, as the token communicated between the nodes acts as the global model; thus, this alignment technique can be categorised as a decentralised method~\citep{multi_token_fl}.

\subsubsection{Attentive Aggregation}
\label{sub_sub_sec:attention_alignment}

In many real-world FL scenarios, clients exhibit heterogeneous data distributions, varying reliability, or differing contributions to model performance. 

Attentive Aggregation addresses this challenge by dynamically weighting client updates based on their relevance, or reliability to the global model. Inspired by attention mechanisms commonly used in deep learning, these techniques enable FL systems to focus more on valuable updates, leading to improved model convergence and robustness.

FedAtt~\citep{attention_aggregation} builds on the FedAvg algorithm but enhances it with an attentive aggregation algorithm, allowing the global model to prioritize updates from more informative or reliable clients. Further enhancements and variations are proposed in~\cite{attention_aggregation_2} and~\cite{transformers_attention_aggregation_fl}, integrating transformer-based attention models to refine the aggregation process.
\subsubsection{Concatenation Aggregation}
\label{sub_sub_sec:concatenation_alignment}

In Vertical Federated Learning, where clients hold complementary features for the same set of users, a common approach to combining knowledge is Concatenation Aggregation. This method capitalizes on the distributed nature of feature sets by computing embeddings independently on each client. These embeddings are then transmitted to a central server, where they are concatenated to form a unified latent representation for downstream processing~\citep{vfed_communication}. This concept builds on the principles of split neural networks, where model layers are distributed across different network nodes~\citep{split_learning_origin}.

Concatenating the clients' latent transformations of their data is particularly effective in scenarios where clients contribute distinct and non-overlapping features. By integrating these embeddings at the server, the system achieves a holistic representation of the data without requiring direct data sharing, thereby ensuring privacy. The server processes the concatenated embeddings to perform tasks such as prediction or classification, leveraging the collective knowledge embedded in the distributed features.

\begin{figure}[!htbp]
    \centering
    \includegraphics[clip,width=0.7\linewidth]{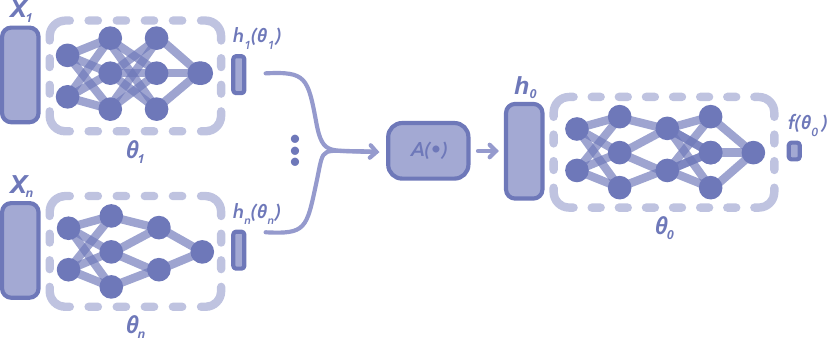}
    \caption{Split-NN architecture overview. Distributed Learning network where each client trains an intermediate representation and shares it with a server, which then fuses them, applies backpropagation, and defuses the update with the clients.}
    \label{fig:split_nn}
\end{figure}

Split Neural Networks~\citep{SplitNNdrivenVP,split_learning_origin} also follow a concatenation aggregation operation. They are designed as a distributed model learning framework, consisting of a local client feature embedding process and a server-side end-tail of the model. The server-side leverages the several sub-model embeddings being produced at the client level and combines them as sub-representations for the tail heads of the server model. After this distributed feedforward process ends, back-propagation is performed on the server and the server returns the jacobians to the client so they can perform their respective local back-propagation (See Fig.~\ref{fig:split_nn} for an abstract depiction of a Split-NN architecture). Split Neural Networks also slightly distance themselves from the split learning framework, since only the initial feature mapping is computed on the client side when using Split Neural Networks.

However, while Concatenation Aggregation enables collaboration within Vertical Federated Learning, it poses challenges related to communication overhead and client synchronization. Efficiently optimising embedding dimensions and minimising redundancy in the concatenated representation are crucial to enhancing both the scalability and performance of this approach.

\subsubsection{Alignment Operator as a Complement to Aggregation}
\label{sub_sub_sec:alignment_op}

In this section, we have explored aggregation as a family of operators for combining local updates in Federated Learning. Building on this, we introduce the notion of an \textbf{alignment operator}, $\mathcal{A}$, as a complementary mechanism to the aggregation operator, a generalised taxonomy inspired by numerous works~\citep{REFC,REFD,REFE}. While aggregation focuses on merging knowledge across clients, alignment provides a framework to constrain and guide this process to meet specific objectives, such as fairness, convergence, or robustness.

As discussed in Section~\ref{sec:preliminaries}, the alignment operator $\mathcal{A}$ imposes constraints on how knowledge is shared and aggregated within the federation (see Eq.~\ref{eq:fl_framework_2}). This additional layer of control enables FL systems to better handle challenges arising from data heterogeneity, diverse client capabilities, and varying optimisation goals.

We recall the problem formulation, introducing the alignment constraining operator:

\begin{equation}
\label{eq:fl_framework_2}
\begin{aligned}
& \underset{\theta, \{\theta_c\}_{c=1}^N}{\text{minimise }}
& & 
\mathcal{G}\left(\mathcal{L}_1(\theta_1), \ldots, \mathcal{L}_N(\theta_N), \mathcal{L}(\theta)\right) \\
  & \hspace{5mm}\text{s.t.}
& & \mathcal{A}\left(\theta, \{\theta_c\}_{c=1}^N\right) \leq \epsilon, \quad \left(\theta, \{\theta_c\}_{c=1}^N\right) \in \Theta.
\end{aligned}
\end{equation}

The alignment operator plays a pivotal role in refining the aggregation process by introducing targeted constraints and objectives. For example:
\begin{itemize}
    \item \textbf{Client Prioritisation:} Alignment can prioritise updates from specific clients based on their data quality, importance, or relevance to the global objective.
    \item \textbf{Balancing Data Heterogeneity:} By guiding aggregation to account for non-iid data distributions, alignment can mitigate biases and ensure fair contributions across clients.
    \item \textbf{Feature Representation Alignment:} Alignment can synchronise feature representations across models, promoting consistency and enhancing collaborative learning outcomes.
\end{itemize}

By complementing the aggregation operator, alignment strategies allow Federated Learning systems to achieve more targeted and efficient outcomes. This synergy between aggregation and alignment expands the optimisation landscape, providing additional flexibility to address the complex demands of real-world FL scenarios.

\begin{notebox}
For a comprehensive review of Model Aggregation Strategies, refer to the survey by~\cite{model_aggregation_survey}.
\end{notebox}

\subsection{Communication Strategy}
\label{sub_sec:communication_strategy}

The strategy to communicate different content between federation clients can be segregated into 5 main categories: \textbf{1.} Communication Topology; \textbf{2.} Learning Periodicity; \textbf{3.} Content; \textbf{4.} Compression. \textbf{5.} Client Selection.

\subsubsection{Communication Topology}
\label{sub_sub_sec:communication_topology}

The design of different topology solutions for a distributed system is driven by multiple factors:  

\begin{itemize}
    \item \textbf{Simplicity and ease of deployment/debugging:}  A well-structured topology simplifies implementation and maintenance.  
    \item \textbf{Robustness to single points of failure:} Decentralised or redundant structures prevent system-wide disruptions.  
    \item \textbf{Faster convergence:} Optimised topologies can accelerate model synchronization and training efficiency.  
    \item \textbf{Mitigating trust issues between entities:} Certain topologies help address concerns related to data privacy and inter-organizational collaboration.  
\end{itemize}

\begin{figure}[!htbp]
    \centering
    \includegraphics[clip,width=0.7\linewidth]{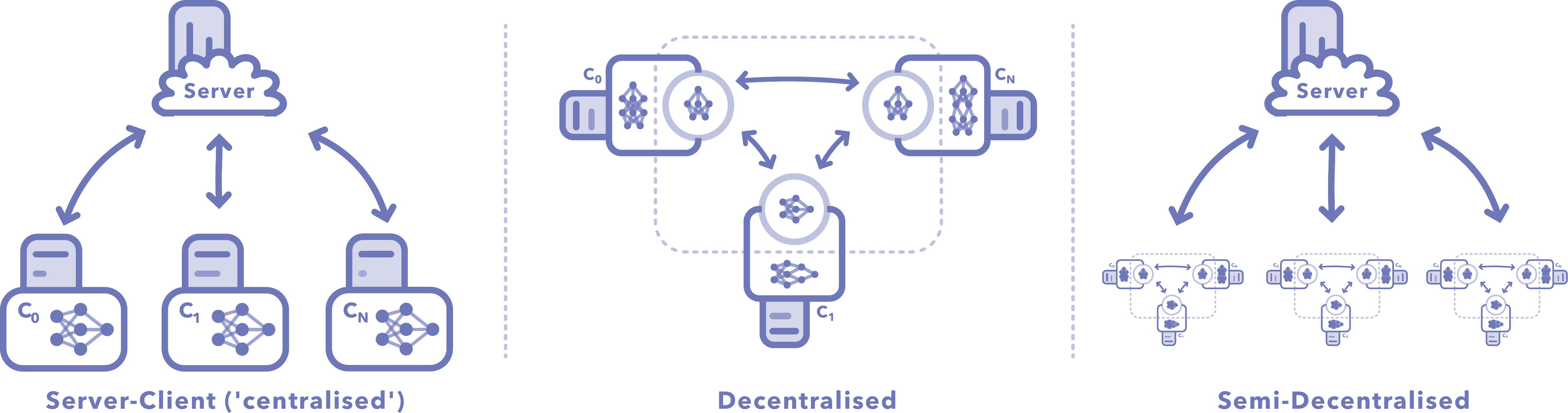}
    \caption{Comparison of distributed network topologies: (Left) Centralised server-client star schema with a single control point. (Middle) Decentralised peer-to-peer system with direct client communication. (Right) Semi-decentralised topology combining peer-to-peer and server communication.}
    \label{fig:communication_topology}
\end{figure}

The most popular architecture topologies for Federated Learning are centralised (star schema) networks, decentralised networks and semi-decentralised networks, which represent a hierarchical mixture of both server-dependent (centralised) network and peer-to-peer (decentralised) communication (See Fig.~\ref{fig:communication_topology}).

\paragraph{\textbf{Decentralised Learning.}}
\label{par:communication_topology}
Cross-silo Federated Learning (FL) combines resources from different organisations, which can raise trust concerns on centralised FL since the organisations must cooperate on central server behaviour.

Decentralised FL can diminish these concerns, by utilising Peer-2-Peer (p2p) communication, removing the need for a central server. This is beneficial since it reduces bandwidth communication and makes the system more reliable to infrastructure failures.

The work of FL Proxy Model Sharing~\citep{ProxyDecentralizedFL}, for example, proposes a mesh of nodes, which learn both a private model and a proxy one which is shared and aggregated each round with a peer node. They control the distillation of knowledge by adjusting a weight between the knowledge transfer of the private to the proxy, and the inverse.

Sharing model parameters faces the risk of saturating a network, (the Large Language Model, Llama-2, for instance, accounts for 70 Billion of parameters~\citep{Llama2}), therefore being able to exchange a distillation of knowledge instead of the actual model parameters is communication-wise beneficial. Not only does this approach lighten the communication bottleneck, but it also better protects the privacy of the raw data used for training.

BrainTorrent~\citep{descentralized_BrainTorrent} proposed a simple p2p Federated Learning setup. In this setting, each client maintains a local model and a data structure that records the model versions of other network clients. At each communication step, a client requests model updates and local dataset size from other clients with a more advanced model version to perform a standard weighted aggregation locally.

Gossip-based protocols are also widely explored in p2p federated networks~\citep{gossip_learning,gossip_feddual}. FedDual~\citep{gossip_feddual} explores the use of a pair-wise gossip protocol to extract knowledge from neighbouring nodes. Additionally, it addresses privacy and security concerns commonly associated with distributed communication by introducing a Gaussian mechanism to perturb each weighted local gradient.

Additionally, token-based methods are frequently used in decentralised learning to optimise communication bandwidth. A token model circulates between nodes, progressively aggregating knowledge before periodically synchronizing with other nodes to enhance global learning~\cite{token_random_fl}.

Peer-to-peer networks also play a fundamental role in blockchain technology, serving as the underlying infrastructure that enables decentralization. In a blockchain system, nodes communicate and validate transactions directly with one another in a distributed p2p network. This architecture enhances fault tolerance, security, and censorship resistance, making it a crucial component of blockchain-based applications, including FL~\cite{blockchain_survey_fl}.

\begin{notebox}
Regarding privacy and security concerns in decentralised federated learning, ~\citet{decentralized_fl_privacy} provides a comprehensive summary in a survey format.
\end{notebox}

\paragraph{\textbf{Semi-decentralised Learning.}}
\label{par:semi_decentralized}
Straggler nodes deteriorate the convergence of FL as the computed local updates become stale/delayed. Therefore, fully trusting client nodes to deliver their learning progress hinders the progress of the global learning process. Semi-decentralised FL is designed to be a hybrid approach between centralised and decentralised communication systems. More precisely, as a whole the system is distributively centralised (relies on a server), but in extreme edge cases, it uses peer-to-peer communication to encompass network issues fairly~\cite{SemiDecentralizedFL}. As expressed by \citet{SemiDecentralizedFL}, there are two types of stragglers: Computation-limited stragglers 
 (type I) and communication stragglers (type II). They suggest and develop a method to utilise neighbour clients to surpass straggler (type II) issues, where local updates are conveyed to the central server with
the help of neighbouring nodes, when the direct connection with the central server is not optimal.

\subsubsection{Synchronous and Asynchronous Coordination}
\label{sub_sub_sec:communication_periodicity}

Standard Federated Learning methods (e.g. FedAvg~\citep{FedAvg}, FedProx~\citep{FedProx}), can be described as Synchronous Federated Learning since they follow an iterative and incremental recipe, where the server waits for a client agglomeration Cohort to finish local training and consequently, the clients wait for the server to aggregate the global model and share with the clients again. This synchronous behaviour makes the system agnostic to client/server resources, being dependent on Stragglers, which are the slowest clients to perform training on a communication round.

\begin{figure}[!htbp]
    \centering
    \includegraphics[clip,width=0.6\linewidth]{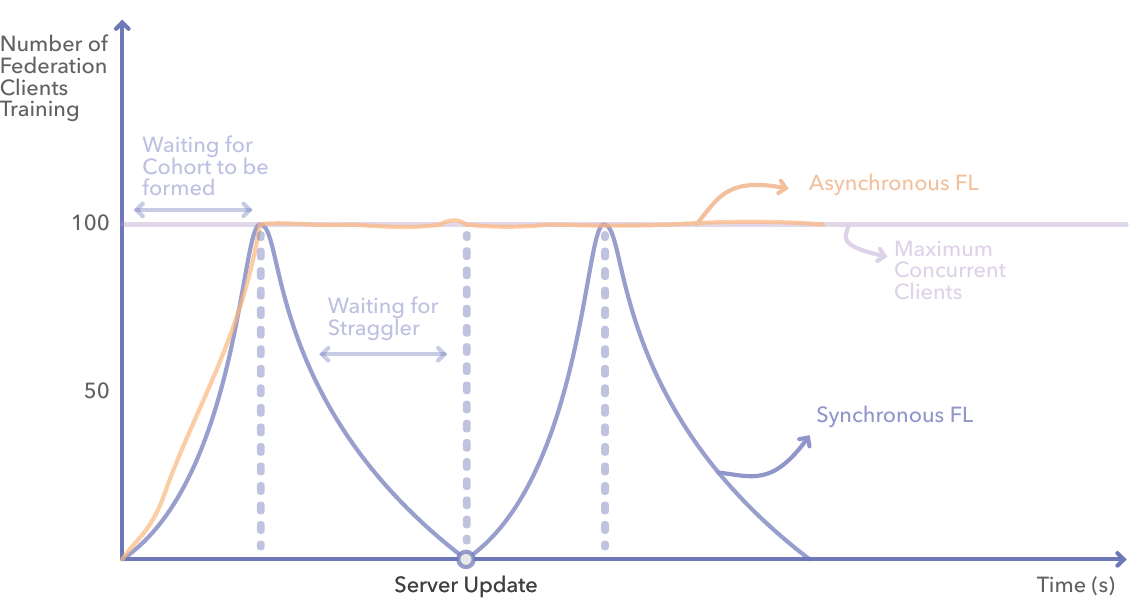}
    \caption{Impact of stragglers on Synchronous vs Asynchronous Federated Learning. Initially, both approaches behave similarly, but after cohort formation, asynchronous servers/clients progress faster by not waiting for stragglers, leading to quicker convergence in fewer training rounds.}
    \label{fig:async_fl}
\end{figure}

\paragraph{\textbf{Synchronous Federated Learning.}} Synchronous Federated Learning is designed for low-concurrency use cases and is preferred due to ease of analysis/debugging and privacy guarantees. However, in high-concurrent cases, where the system follows a cross-device FL architecture, characterised by the clients having different compute power and intermittent availability, an Asynchronous approach can help with training speed (See Fig.~\ref{fig:async_fl}).

\paragraph{\textbf{Asynchronous Federated Learning.}} Asynchronous Federated Learning deviates from the synchronous process in an attempt to accelerate the convergence of the learning procedure, by reducing the effect of Stragglers.

FedBuf~\citep{async_fl} assumes the existence of Stragglers in the learning communication process and attempts to mitigate this effect by utilising buffered asynchronous aggregation. The server contains a buffer that awaits $K$ clients to send their local updates and aggregates these model parameters. The learning process and communication are performed asynchronously between the clients and server and utilising privacy-aware and secure techniques like Differential Privacy and Secure Aggregation.

Papaya~\citep{papaya} extends FedBuf's work using a Trusted Execution Environment for secure aggregation and demonstrates that AsyncFL can achieve more fair
models than SyncFL with over-selection. Over-selection is a technique used in SyncFL, corresponding to dropping slow clients' train results. Ultimately, this behaviour can develop a bias towards faster-performant clients.

It is worth pointing out that Asynchronous FL, inevitably, introduces a phenomenon named staleness. Since clients independently start training new models without global convergence, some clients continue training over previous server model versions. One solution to this staleness factor is establishing a timeout limit, after which a pre-defined amount of time a straggler model is removed from an update phase~\citep{papaya}.

\paragraph{\textbf{Semi-Synchronous Federated Learning.}}
\label{par:semi_sync_fl}

To mitigate the stillness of fast learners in Synchronous policies, the Semi-Synchronous FL policy was introduced~\citep{SemiSynchronousFL} allowing every learner to stop, not in terms of convergence metrics but according to pre-defined synchronisation time point. The selection of this time threshold is based on the maximum time it takes for the federated straggler to finish training. Semi-synchronous FL, is therefore dependent on network statistics, requiring an initial training round to collect the respective cohort statistics.

Examine Table.~\ref{tab:sync_comparison} for a comparison summary of different policies for scheduling training sessions.

\begin{table}[!htbp]
\centering
\caption{Learning Scheduling Policies comparison. Adapted from \citet{SemiSynchronousFL}.}\bigskip
\input{tables/sync_comparison}
\label{tab:sync_comparison}
\end{table}

\subsubsection{Communication Content}
\label{sub_sub_sec:communication_content}

Since the inception of Federated Learning, numerous approaches have been developed to enable effective communication between clients and servers. Within this scope, a variety of communication content alternatives have emerged, each tailored to address specific challenges like privacy, efficiency, and scalability.

\paragraph{\textbf{Early Approaches.}}
The foundational methods, \textbf{FedAvg} and \textbf{FedSGD}~\citep{FedAvg}, introduced the concept of learning a global model by sharing:
\begin{itemize}
    \item \textbf{Neural Network Parameters (FedAvg):} Clients share their locally trained model parameters for aggregation into a global model~\cite{FedAvg}.
    \item \textbf{Loss Gradients (FedSGD):} Clients transmit gradient updates computed from their local data~\cite{FedAvg}.
\end{itemize}

\paragraph{\textbf{Advanced Communication Content.}}
To improve privacy, or communication efficiency, or communication heterogeneity, several alternative communication payloads have been proposed:
\begin{itemize}
    \item \textbf{Proxy Models:} Proposed in~\citet{ProxyDecentralizedFL}, these act as knowledge compression mechanisms, enabling efficient transmission of compressed model knowledge between clients and servers.
    \item \textbf{Model Parameters with Noise Masks:} Introduced in~\citet{original_se_fl}, this approach mitigates model inversion attacks by adding noise masks to shared parameters, ensuring privacy while maintaining accurate model aggregation.
    \item \textbf{Token Models:} In this client-to-client sharing mechanism~\citep{multi_token_fl}, global model parameters are transmitted across clients to either train on local data or extract knowledge from pre-trained local models.
    \item \textbf{Model Predictions:} Instead of sharing model parameters, clients share the outputs (predictions) of their local models~\citep{RobustHFL}. These predictions can be used by other nodes to improve learning without exposing raw data.
    \item \textbf{Embeddings:} As seen in~\citet{split_learning_origin}, clients focus on learning and sharing sub-representations (embeddings) instead of performing end-to-end training. This reduces communication overhead while preserving task-specific learning.
    \item \textbf{Statistics:} Clients transmit metadata such as data distribution hints, performance metrics, and the number of local data points.
\end{itemize}

\subsubsection{Compression}
\label{sub_sub_sec:model_compression}

One of the key challenges in distributed learning is the communication bottleneck caused by bandwidth-heavy transmission and inefficient communication rounds. This issue is exacerbated by several factors, including large model sizes, limited network resources, data heterogeneity among clients, and high federation participation.

To address these challenges, various model compression techniques are employed, such as \textbf{pruning}, \textbf{quantisation}, \textbf{sparsification}, and \textbf{knowledge distillation}. These methods aim to reduce the size of transmitted updates, improve communication efficiency, and minimise resource usage while maintaining model performance.

\paragraph{\textbf{Model Pruning.}}
As Neural Networks are universal approximators of non-trivial functions, they are flexible to specific structural manipulations without significant performance degradation. For instance, studies on Transformer networks have shown that some attention heads contribute minimally to final predictions~\citep{attention_prunning}. This insight suggests that \textbf{pruning} unimportant components can reduce model size and computation time while maintaining comparable performance. In Federated Learning (FL), \textbf{FedTiny}~\citep{FedTiny} dynamically prunes network components during training based on their relevance to final predictions, making it suitable for scenarios with intensive computational and memory constraints.

\paragraph{\textbf{Model Quantisation.}}
Model Quantisation involves reducing the number of bits used to represent the numerical values encoding model weights, thereby decreasing the size of transmitted updates. \citet{quantizationRFL} introduced a federated quantisation technique using Kurtosis Regularisation (KURE) and Additive Pseudo-Quantisation Noise (APQN) to create robust quantised models. Their approach ensures that the model remains tolerant to varying bit widths while retaining full-precision accuracy. Additionally,  Lazily Aggregated Quantisation (LAQ)~\citep{lazily_aggr_quant} compresses gradient updates and skips non-informative communication rounds, reducing communication overhead. By sharing gradients with inherent noise, this method also enhances privacy by mitigating potential exploitation of communication artefacts.

\paragraph{\textbf{Sparsification.}}
Sparsification is another effective compression strategy that reduces the size of updates by transmitting only the most significant components of the model parameters or gradients. For example, sparse gradient updates focus on retaining only a small fraction of the most informative elements, discarding less significant values. This approach significantly reduces communication overhead while maintaining learning efficacy. Furthermore, sparsification inherently enhances privacy, as fewer model details are shared during communication rounds~\citep{sparsification}.

\paragraph{\textbf{Knowledge Distillation.}}
Knowledge Distillation is a popular method for transferring knowledge from a large and complex model (teacher) to a smaller and simpler model (student). In the context of FL, this technique is particularly useful for addressing challenges such as data and model heterogeneity. For instance, works like~\citet{ProxyDecentralizedFL} and~\citet{RobustHFL} employ knowledge distillation to harmonise learning across diverse client models while ensuring robust global performance.

\subsubsection{Client Selection}
\label{sub_sub_sec:client_selection}

Distributed Learning systems face both physical and digital constraints. During federated training, devices may be turned off unexpectedly or the network connectivity might drop during the parameters-sharing phase. Furthermore, the training and communication speeds might differ due to the internet physical components of the federated client, creating progress bottlenecks in the system. Such situations make it necessary to design FL frameworks robust to these learning inconsistencies. One way to mitigate it is by instituting the notion of Client Selection in such a way that the learning procedure is mildly affected and such that it can be enhanced based on some heuristics (e.g. segregating clients into clusters to align clients in groups in terms of data homogeneous characteristics).

The simplest client selection technique is based on the Cohort mechanism, where a subset of available clients is chosen to be part of the Federated Learning process. One further customisation step is to weigh the client selection based on heuristics (e.g. client number of samples, noise rate, speed performance).

FedRN~\citep{FedRN} assumes clients have noisy data, and therefore consults K-reliable neighbour clients' models to assess the health of its data and trains its local model using the not noisy filtered data. These noise-aware auxiliary training models are bi-modal Gaussian Mixture Models, which express the probability distribution of noisy versus clean examples.

TiFL~\citep{tifl} is an FL framework designed to mitigate the effect of stragglers by segregating clients into tiers depending on their performance.

\subsection{Distributed Data Handling}
\label{sub_sec:data}

The simplistic assumption that distributed data behaves similarly to centralised, curated data is unrealistic. Statistically speaking, most Machine Learning mechanisms rely upon the assumption that the data/evidence supporting a specific task (e.g. regression, classification, generative modelling) is Independent and Identically Distributed (iid).

In reality, however, distributed data is either not identically distributed or not independent; thus, federated data is generally non-iid. We present the four core types of non-iid data observed in real-world scenarios~\citep{stochastic_clustering_fl}:

\begin{itemize}
    \item \textbf{Feature Distribution Skew.} The marginal distribution of the data features $P(x)$ varies across federated clients. Example: Imagine two equal CCTV cameras pointing at the respective garden of a house. One of the gardens has trees, the other has a tree and many gnomes over the grass. The distribution of items/objects and the features are different.
    \item \textbf{Label Distribution Skew.} The marginal distribution of label data $P(y)$ varies across federated clients. Example: Imagine two equal CCTV cameras recording a garden. One CCTV camera is in Country X and the owner populates it with cactuses, the other CCTV is recording a Country Y house filled with gnomes. The Country Y house will probably never have a cactus so for the Country Y's CCTV there is no label for cactus.
    \item \textbf{Feature Concept Skew.} The features conditioned on the input labels $P(x|y)$ vary across federated clients, whereas $P(y)$ remains constant. Example: Imagine two CCTV cameras recording the same object from two different perspectives. Even though the object $y$ remains the same, the encoded understanding of each object will vary deeply since the captured features are not the same.
    \item \textbf{Label Concept Skew.} The labels conditioned on the input features $P(y|x)$ vary across federated clients, whereas $P(x)$ remains constant. Example: Imagine two different CCTV cameras recording the same object. The raw input is the same, but since the cameras are different, the perception of the object $y$ can vary.
\end{itemize}

\subsubsection{Heterogeneous Data}

Distributed Learning systems by nature imply different data characteristics over the different nodes composing the distributed network. This heterogeneous non-iid data behaviour hinders the learning mechanism (e.g. FedAvg:~\citet{FedAvg}) and makes several standard/classical Machine Learning approaches not viable. 

\paragraph{\textbf{Clustering Federated Learning (CFL).}}
\label{sub_sub_sec:clustering}

\begin{figure}[!htbp]
    \centering
    \includegraphics[clip,width=0.5\linewidth]{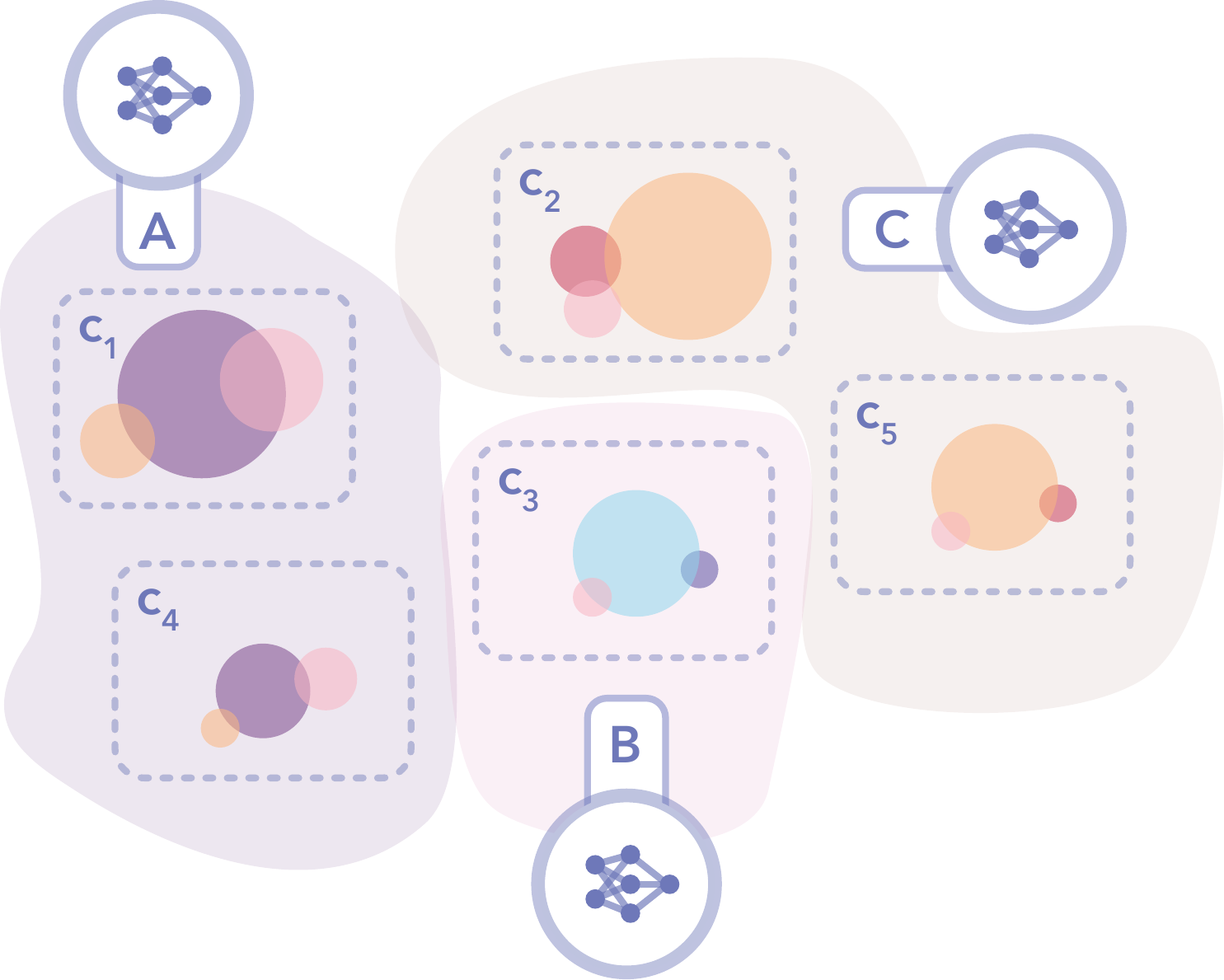}
    \caption{Abstract overview over a federated clustering algorithm, which groups together clients in clusters (A: client 1,4; B: client 2,5; C: client 3) according to how similar their data distributions are. The circles represent the class distributions when considering a classifying problem (circle colour: the respective class; circle size: the amount of data points corresponding to the class).}
    \label{fig:clustering}
\end{figure}

One way to mitigate the non-iid factor is finding client nodes that share similar data distributions and grouping them. Mathematically, we try to find $k$ clusters, optimising the best client-cluster fit for each client data distribution. This process can be considered a meta-learning mechanism since it paves the way for a more insightful model alignment protocol, by leveraging the information of likewise client nodes when selecting clients to perform the distributed learning alignment~\citep{clustering_fl_efficient,stochastic_clustering_fl} (See Fig.~\ref{fig:clustering} to view a federated clustering abstract example, where clients are grouped so the learning is more robust to heterogeneous data distributions). 

In a sense, CFL exploits the intrinsic geometric properties of FL by grouping clients through the usage of measurement distances between client nodes. One first work in the CFL realm, \citet{cluster_fl_privacy} recursively performs a top-down FL bipartition over the training federation. Until convergence, an FL training round is executed, followed by the server performing a bipartition over its clients to understand whether this new partition better separates congruent from incongruent clients. If the new partition is worse (evaluation metric smaller than a threshold hyperparameter) the system falls back to the previous clustering structure.

StoCFL~\citep{stochastic_clustering_fl} is a Bi-Level clustering approach to FL, agnostic to the total amount of clusters. It estimates the similarity of any two clients to assign \textit{close} clients to the same cluster and eases the data distribution divergences during the model alignment phase. Models are aligned in the same cluster space and then further aligned globally using each cluster model for regularisation purposes.

\subsubsection{Unlabeled Data}
\label{sub_sub_sec:u_fl}

Most Federated Learning Literature tackles supervised learning use cases, however, the presence of unlabeled data over Internet of Things (IoT)/edge devices is more prevalent than explicitly labelled one. 

Regarding cyber-attack vulnerabilities, some works (e.g.~\citet{AE_U_ensemble}) have proposed federated auto-encoder architectures to detect anomalies in unlabelled data, which intrinsically learn a codification of the data to spot data distribution shifts.

FedUL~\citep{UData} studies how clients can effectively learn from situations where no data is labelled, but assuming prior knowledge of the class distribution present at each client.

Federated Contrastive Averaging (FedCA:~\citet{FedCA}) undertakes an unsupervised representation learning approach utilising contrasting learning techniques on the client side and server aggregation to align the local samples logits database and the local models onto a global one. FedCA extends SimCLR~\citep{simCLR} to a federated setting with two key modules: a (local and global) database containing the logits of the client nodes, which is shared with the server and a representation learning alignment model.

FedX~\citep{fedX}, concerned with data sharing among federation clients (e.g. FedCA:~\citet{FedCA}), explores a different take on unsupervised contrastive Federated Learning through the usage of local and global knowledge distillation.

\paragraph{\textbf{Feedback-based Implicit Annotated Data.}} 

\citet{amazon_noisy_feedback} provide a realistic take on utilising natural human behaviour as pseudo labels (Feedback Noise Modeling) in the domain of smartphone usability, assuming no labelled data is available on the device, rather they resort to intrinsic positive/negative feedback from behaviour cues. This is an interesting take since it provides a platform for an active learning deployment. Furthermore user feedback signals represent a robust source of knowledge to reduce the effect of concept and data drifts, especially in scenarios where data is majorly unlabelled.

\subsubsection{Streaming Data - Online Learning}

Online Learning assumes that data instances we are interested in learning come in a streaming fashion and may not be available after training/inference. In Online Learning no target label is known beforehand, therefore the learning process is continuous, through an iterative process: \textbf{1.} At time stamp $t$, a new batch or data point $X_t$ arrives; \textbf{2.} A model makes a prediction $\hat{y} = M_t(X_t)$; \textbf{3.} Ground truth $y_t$ is revealed, with or without a time delay $\delta t$; \textbf{4.} The model $M_t$ learns from data pair $(X_t,y_t)$ and, thus, updates to $M_{t+1}$; \textbf{5.} This process repeats on the arrival of a new data element. For neural networks, online stochastic gradient descent is normally used. If there are no imminent shifts in data distributions, the online learning scheme can be conceived as an offline learning procedure by collecting and storing the streaming data and only dispatching the learning process as needed.

Online FL is the expansion of Online Learning methods to a distributed setting, where each client continually receives a stream of data (e.g. wearable monitoring devices). It is deeply explored in real-time use cases like Fraud Detection, Online Services (Music/Movie streaming) Recommendations, etc. 

FedOMD~\citep{OnlineFL} recognises that data may not be available offline and designs a collaborative method to minimise a regret metric over a convex problem.

\citet{decentralized_personalised_online_fl} concerned with large-scale recommendation systems extends Online FL to personalisation and a decentralised setting. They introduce a novel learning model aggregation scheme, which performs a learning step over a given arriving data element and a learnt weighted combination of the local and peer models to update the local model.

Some works also consider the hyper-parameters tuning of Federated Learning within the optimisation problem (e.g. learning rate, batch size). \citet{FederatedHypergradientDescent}, for instance, leverages an online procedure to update the hyper-parameters dynamically.

\subsubsection{Datasets and Benchmarks}
\label{sub_sub_sec:datasets_and_benchmarks}

Datasets and benchmarks play a pivotal role in Federated Learning (FL), enabling researchers to simulate realistic scenarios, compare algorithms, and address key challenges such as non-iid data, noisy labels, and privacy preservation. Below, we provide a consolidated overview of commonly used datasets and benchmarks in FL, categorised by domain and task:

\input{tables/datasets}

The table combines datasets and benchmarks, categorizing them by their primary domain or task. It highlights the diversity of datasets used in FL, as well as specialised benchmarks designed to address FL-specific challenges like non-iid data distributions, noisy labels, and privacy concerns.

\subsubsection{Data Processing}
\label{sub_sub_sec:data_processing}

\paragraph{Data Partitioning.} In real-world Federated Learning (FL) applications, data is inherently partitioned across clients, often reflecting their individual contexts (e.g., user devices or organizations). However, when evaluating FL algorithms using benchmarks, data is often synthetically partitioned to either simulate a federated scenario or test the robustness of algorithms under varying degrees of heterogeneity.

A common and simple partitioning technique, initially used by FedAvg~\citep{FedAvg}, is \textit{data sharding}. In this method, training examples are sorted by class labels and divided into shards, with each client assigned a predefined number of shards (e.g., two shards per client). While effective for basic simulations, this method is limited in its ability to represent realistic non-iid scenarios.

\begin{figure}[!htbp]
    \centering
    \includegraphics[clip,width=0.6\linewidth]{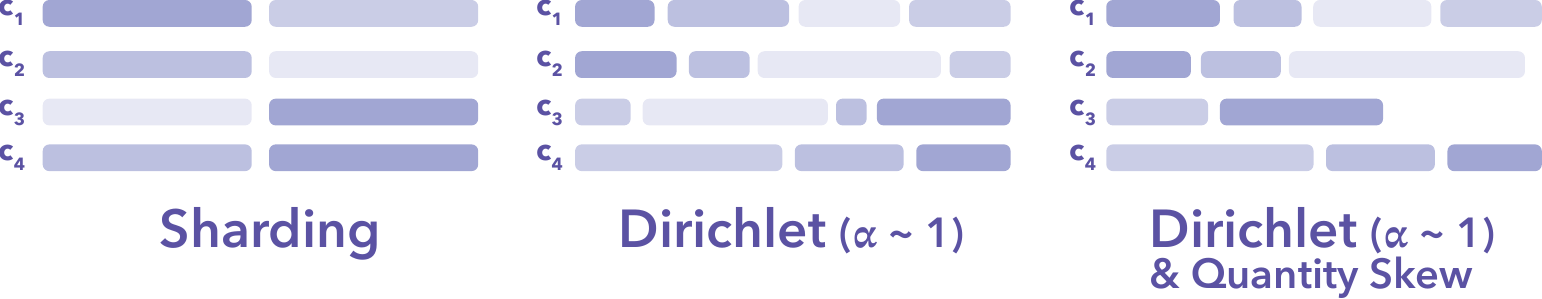}
    \caption{Popular data preparation techniques involve using Sharding (left), the Dirichlet distribution (middle), and a combination of the Dirichlet distribution and manipulating the number of elements in the datasets (right). The different colors represent distinct features, while the size of each feature indicates its distribution scale.}
    \label{fig:data_distribution}
\end{figure}

To address this, a more probabilistic and systematic partitioning approach was introduced: the \textit{Dirichlet Distribution-based partitioning}~\citep{FL_Dirichlet}. This method allows for fine-grained control over the heterogeneity of client data. By tuning the $\alpha$ parameter of the Dirichlet distribution, one can control the level of non-iidness among clients: smaller $\alpha$ values produce greater heterogeneity, while larger $\alpha$ values result in more homogeneous data distributions (see Fig.~\ref{fig:data_distribution}). This flexibility makes the Dirichlet-based method widely used for benchmarking FL systems under diverse non-iid settings.

Regarding heterogeneous data scenarios (inconsistent class label distributions among clients), some researchers~\citep{RobustHFL} have proposed the use of a publicly available dataset shared among all clients, in addition to their private data. The underlying rationale is that this common dataset serves as a shared foundation to align the models, mitigating the effects of shifts in local data distributions.

However, the use of public datasets in a federated setting has been met with criticism, as such datasets are often unrealistic in real-world scenarios. Public datasets—especially large, well-annotated ones—are rarely available, particularly for highly specialised or private domains. Nevertheless, we argue that the concept retains relevance in two distinct forms:
\begin{enumerate}
    \item \textbf{Common Knowledge Sources:} Even if the original data of interest is unavailable, there may exist a shared set of non-confidential metadata or general information common to all parties. Such data, even if in a different modality, can provide guidance for the learning process.
    \item \textbf{Open Public Datasets for Transfer Learning:} In some cases, publicly available datasets similar to the task at hand can be leveraged through techniques like transfer learning or knowledge distillation. These approaches allow models to generalize better across clients by pretraining on public data and fine-tuning on private data.
\end{enumerate}

\paragraph{\textbf{Noisy Learning.}} 

While clean public datasets are often used in research, real-world data is frequently corrupted by adversarial attacks, machine-generated errors, or human annotation mistakes. Unfortunately, most popular datasets do not capture such imperfections. To address this gap, researchers typically simulate noisy annotations by injecting artificial perturbations into datasets. This process allows for controlled evaluation of robustness against noisy labels, but it falls short of fully replicating the complexities of real-world noisy data.

In the noisy learning realm, two widely used noise perturbation styles of injecting label
noise are Symmetric Label Flipping~\citep{symmetricNoise2,symmetricNoise} and Pair Label Flipping~\citep{CoteachingRT_pair_flipping}. Symmetric Label Flipping involves altering the labels of a dataset so that each label has an equal probability of being changed to any other class label. In contrast, Pair Label Flipping is characterised by swapping labels between two specific predefined classes, rather than randomly changing to any class.

\subsection{Model Agnostic Design \&
Specialised Learning Paradigms}
\label{sub_sec:model_agnostic}

Similar to traditional monolithic machine learning, Federated Learning enables the execution of numerous tasks, expanding the possibilities of distributed processing. Locally, on the clients/devices, these tasks can be addressed using various models, such as Convolutional Neural Networks (CNNs), Transformers, etc. This flexibility allows for tailored solutions to specific challenges, leveraging the strengths of different model architectures to enhance performance and efficiency across diverse applications.

To an extreme extent, the model on each client can be a black box concerning other clients on a federation~\citep{fed_md}.

\subsubsection{Model Initialisation}

In Federated Learning, model parameters can be tailored using global or local heuristics. Typically, global model parameters are initialised using normal distributions, in line with standard machine learning practices. However, when the task at hand shares similarities with a previously addressed task, Transfer Learning is often employed, leveraging a pre-trained foundational model. In the context of client-side learning, initialization is generally based on the most recent global model version, which in turn depends on the specific FL aggregation or alignment algorithm in use. For instance, the FedAvg algorithm~\citep{FedAvg} produces a global model that represents the average of the local models, whereas FedPSO~\citep{park2021fedpso} initializes local models based on the best-performing client model from the previous optimisation iteration.

\subsubsection{Neural Architecture Search (NAS)}

Manually designing a neural network architecture to solve a task may be suboptimal, especially in a Federated Learning setting where local data distributions differ and even the tasks/features may not be identical. NAS attempts to dynamically find an optimal neural model architecture balancing a defined search space, search strategy and performance estimation~\citep{nas_2016}. In the realm of FL, NAS was initially proposed by ~\citet{nas_fl} and ~\citet{fednas} tackling online Federated NAS. Online federated NAS optimises both the model and its architecture design simultaneously, in contrast to offline approaches that optimise them sequentially. Two popular techniques for Online Federated NAS are gradient-based methods~\citep{fednas} and Evolutionary Algorithm-based methods~\citep{zhu2021real}.

\begin{notebox}
To further explore the topic of NAS in a Federated Learning setting we invite the reader to read the survey~\cite{fed_nas_survey}.
\end{notebox}

\subsubsection{Federated Reinforcement Learning}
\label{sub_sub_sec:alignemnt_methods}

Reinforcement Learning (RL) represents a distinctive Machine Learning paradigm that conceptualises the learning journey as a Markov Decision Process. In simpler terms, RL adheres to a ``learn by doing'' doctrine, wherein the learning system continually refines its environment understanding by striking a delicate balance between adapting to errors based on the rewards received for actions taken in a given state and exploring uncharted territories within the (state, action, reward) space. This iterative process, characterised by the dynamic interplay of exploitation and exploration, forms the essence of Reinforcement Learning. It is a powerful approach for training intelligent systems to make optimal decisions in complex, dynamic environments. 

In physical real-world cases, however, the concept of learning from real experience is impractical as it would be extremely costly and time-consuming. Therefore to combat this gap, software simulations are often used to substitute real-world prior experiences to fasten RL applications in the real world.

\begin{figure}[!htbp]
    \centering
    \includegraphics[clip,width=0.6\linewidth]{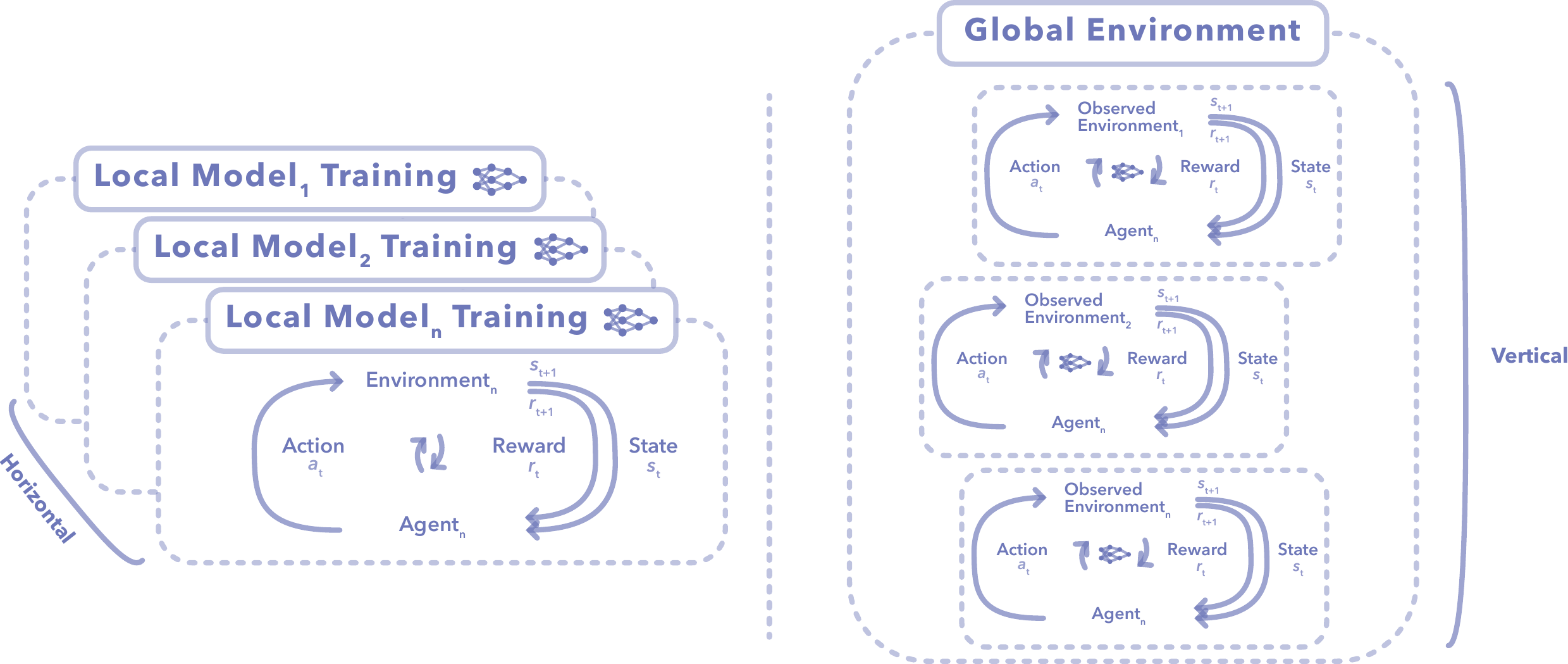}
    \caption{Overview of Horizontal vs. Vertical Reinforcement Federated Learning. (Left) Horizontal FL, where each node explores a different environment. (Right) Vertical FL, with multiple clients observing a shared environment. Adapted from ~\citet{federatedRL_survey_2021}.}
    \label{fig:rl_fl}
\end{figure}

In essence, applied to FL, reinforcement learning, like normal Federated Learning can be segregated into Horizontal and Vertical FL (See Fig.~\ref{fig:rl_fl} for a visual guide and brief overview of the differences).

In the automotive industry, the usage of reinforcement learning is prominent, due to the intrinsic nature of the dynamism of the environment that a vehicle encounters. In this sector, we often find a Horizontal FL, since each vehicle drives in different locations, encountering different obstacles and weather factors. Vertical Federated Reinforcement Learning is less commonly found in the literature, however, works such as \citet{federatedDRL_vertical} exist exploring agents collaborating in a finite space.

\citet{FederatedRL} proposed a Reinforcement Learning system, a Home Energy Management System (HEMS), to estimate the energy consumption of multiple Smart Homes consisting of multiple energy devices (e.g. Air Conditioner). Generally, in a Federated setting each local entity optimises a single local model, whereas, in this HEMS system, a set of energy devices are independently optimised considering user preferences in their reward functions.

Federated Learning applied to Autonomous Driving has gathered interest as it can help speed up the gap between the simulation performance and real-world scenarios, by transferring the knowledge of other vehicle agents when acting on a given state~\citep{toy_cars_frl}. More recent work on Connected Autonomous Driving (Federated Learning applied to Autonomous driving methodologies) has emerged~\citep{simulated_autoD_frl}.

\begin{notebox}
For further in-depth detail about the state of applications of reinforcement learning in a federation scheme, visit the following surveys~\citep{federatedRL_survey_2021,federatedRL_survey_2023}.
\end{notebox}

\subsection{Security Strategy}
\label{sec:security}
Federated Learning (FL) mitigates raw data exposure by transmitting, for instance, model weights or gradients, instead of raw data across network nodes. However, this alone does not guarantee security, as FL remains vulnerable to various attacks and adversarial threats. With an estimated 60 billion Internet of Things (IoT) devices expected by 2025~\citep{IoTAmmount}, FL is positioned to target low-resource applications due to these devices' interconnectivity, abundance of local data, and advancements in embedded AI acceleration. However, this massive and diverse ecosystem presents critical security challenges. A less favourable factor is that this myriad of machines does not always have the ideal security design, possessing low-quality/cheap components facilitating adversarial attacks to devices and endangering FL's applicability~\citep{embedded_devices}.
Furthermore, a key characteristic concerning the dissemination of Machine Learning models over embedded devices is the possible increase of interest in the theft or destruction of these devices carrying valuable and powerful specialised ML models.

Federated Learning is susceptible to various security threats that can compromise model integrity and system reliability~\citep{survey_privacy_security,decentralized_fl_privacy,embedded_devices}. This module focuses on safeguarding FL systems against a wide range of security threats, including:

\paragraph{\textbf{Data Poisoning Attacks.}}
Data poisoning involves adversarial modification of training data to manipulate the global model. Attackers may inject misleading samples to degrade model accuracy or introduce hidden backdoors \cite{backdoor}. Defense mechanisms such as robust aggregation methods \cite{original_se_fl} and anomaly detection~\cite{wu2022fl} can mitigate such risks.

\paragraph{\textbf{Adversarial Attacks.}}
Adversarial attacks involve carefully crafted perturbations to input data, causing incorrect model predictions \cite{contra}. Federated models are particularly vulnerable as attackers may exploit updates to generate adversarial examples. Defense strategies, for instance, include adversarial training \cite{li2023federated}.

\paragraph{\textbf{Model Theft and Inference Attacks.}}
Model theft occurs when adversaries attempt to reconstruct or steal the global model \cite{moda}. Inference attacks further exploit access to model outputs to infer sensitive training data \cite{membershield}. Countermeasures include \textbf{Differential Privacy}~\citep{diff_privacy_google_library} and \textbf{Secure Multi-Party Computation}~\citep{PracticalSMPC}.

\paragraph{\textbf{Physical Attacks.}}
FL relies on distributed nodes that can be physically compromised. Attackers may tamper with hardware to extract sensitive model parameters or disrupt training. Secure enclave technology \cite{nvidia_confidential_computing, ml_tee} and hardware-based attestation can reduce the impact of such attacks.
Security strategies in FL require a holistic approach, combining cryptographic techniques, anomaly detection, and robust optimisation methods to ensure trustworthiness in distributed learning environments.

\subsubsection{Distributed Byzantine Attacks}

Malign attacks are prominent within Machine Learning systems, where the training data, training process, or inference procedures are exploited. In distributed systems, assaults on the established behaviour of distributed infrastructure are classified as \textbf{Byzantine Attacks}. They are a subset of \textbf{Byzantine Failures}, a broader class of distributed issues (e.g. hardware failures, communication failures, adversarial attacks, protocol inconsistencies). 

On a different line of attack, there is the Data Poisoning Attack. It is a kind of byzantine attack, where a third-party entity tries to infect nodes in a learning network to either alter the local dataset, manipulate the labels, or impersonate a benign node to infiltrate the central learning system at a later stage. One popular way to contaminate target labels is using the Label-Flipping technique where according to some probability distribution, labels in a dataset get altered with different ones. CONTRA~\citep{contra}, for instance, tries to combat label poisoning by ranking network nodes depending on their suspected behaviour.

Apart from poisoning attacks, there are other styles of adversarial learning attacks: Inner Product Manipulation~\cite{xie2020fall}; Sign Flipping~\cite{probabilistic_sign_flipping}; and A Little is Enough~\citet{AL_is_Enough}).

\begin{figure}[!htbp]
    \centering
    \includegraphics[clip,width=0.9\linewidth]{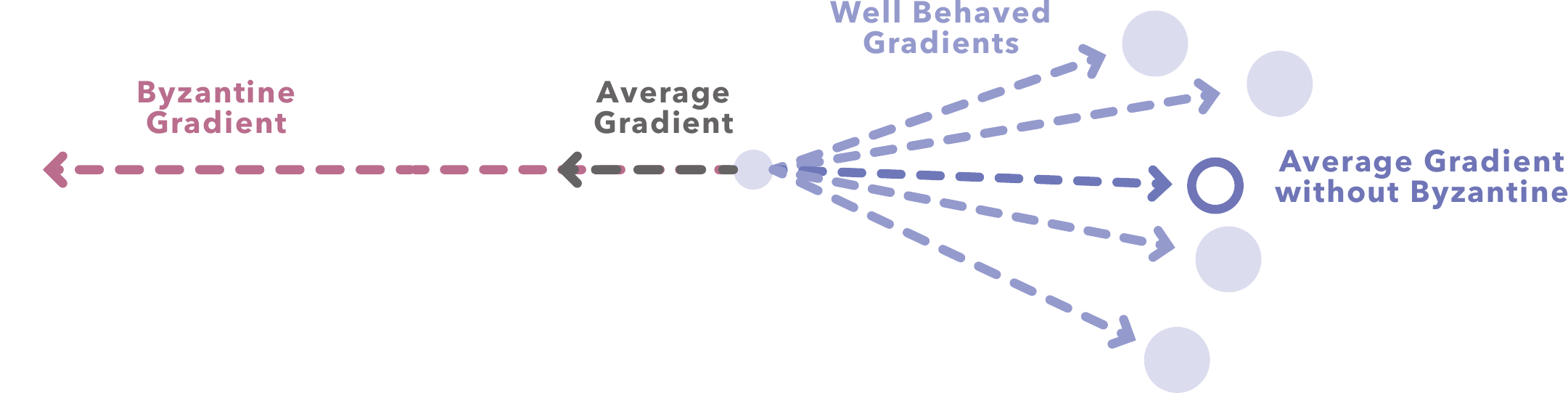}
    \vspace{1mm}
    \caption{Byzantine Gradient Attack. Showcasing how a gradient attack can affect the update of the FedSGD algorithm.}
    \label{fig:gradients_byzantine}
\end{figure}

Focused on building robust distributed learning optimisation methods, some works (e.g. GeoMedian~\citep{GeoMedian}, Krum~\citep{krum}) have
concentrated on gradients' Euclidean norm guarantees. However, the norm of a vector is arguably less important than its direction. In this sense, Zeno~\citep{Zeno} was proposed to enforce gradient direction constraints in the learning process. Some clustering techniques~\citep{Li2021ByzantinerobustFL} group similar clients based on their gradient updates, relying on cosine similarity. In isolation, cosine similarity might not be ideal as it solely accounts for relative directions while disregarding the magnitude of each vector. See Fig.~\ref{fig:gradients_byzantine} to visualise how a byzantine gradient can negatively affect an FL algorithm's learning process.

\subsubsection{Trusted Execution Environment (TEE)}
Secure and Privacy-preserving Federated Learning frameworks have been proposed using Trusted Execution Environments to hide model/gradient updates from malign adversaries~\citep {PPFL_tee}. 

\begin{figure}[!htbp]
    \centering
    \includegraphics[clip,width=0.6\linewidth]{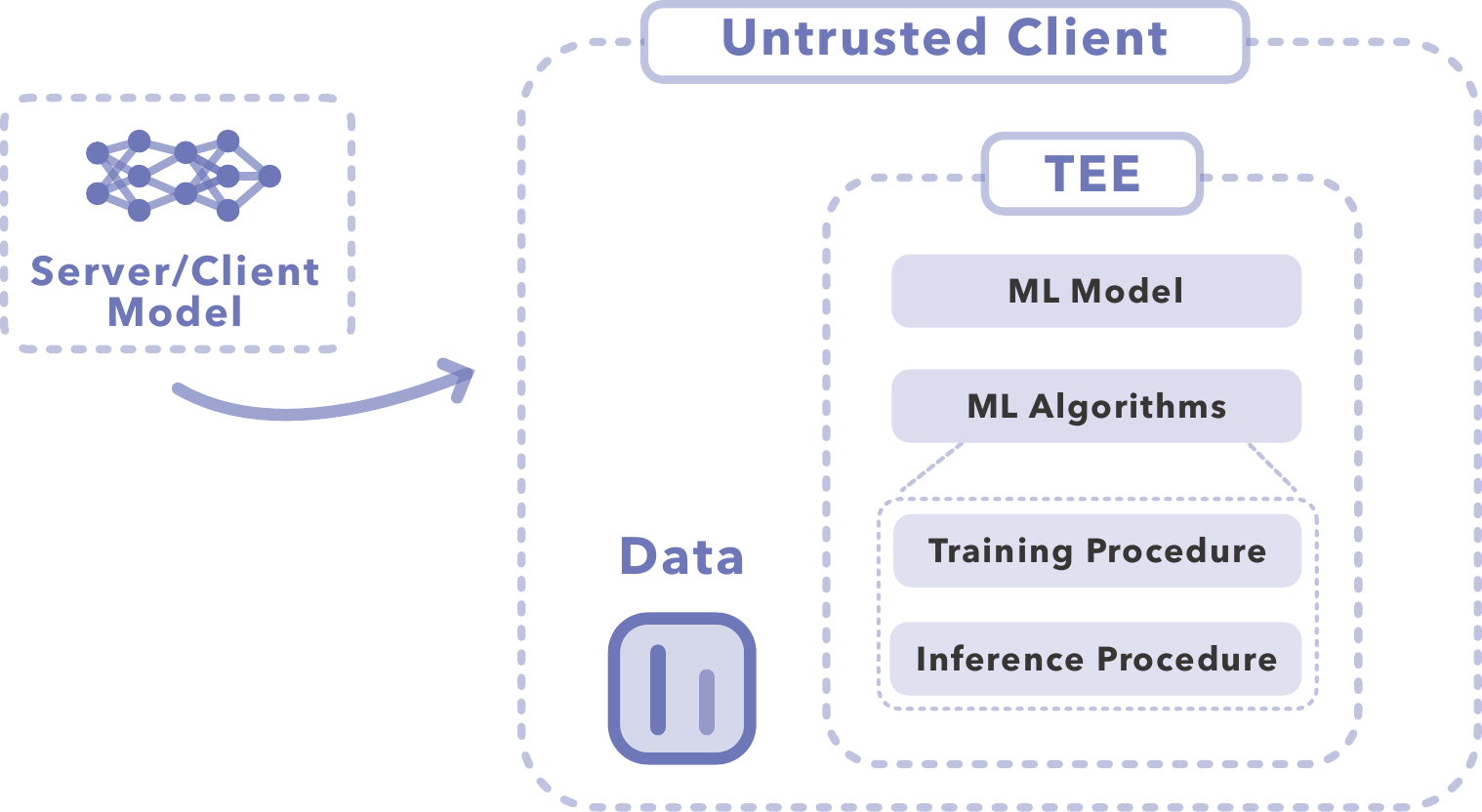}
    \vspace{1mm}
    \caption{We depict a general case where a TEE is used on federation clients to protect the contamination of ML models and the execution of the learning inference procedures. Adapted from \citet{ml_tee}.}
    \label{fig:tee}
\end{figure}

The core idea behind TEE is that along a computer system's Trusted Computing Base (TCB), there is a hardware and application/software enclave where specific computations and resources are protected. Systematically, TEE is powerful because it provides a flexible and secure platform for processing Federated Learning computation (see Fig.~\ref{fig:tee}). Unfortunately, since they run over a TCB (the secure components of a system), which must be audited and remain secure, the hardware enclave limits computational Machine Learning possibilities: \textbf{1. Secure Memory Limitations.} TEE's TCB physical memory is typically constrained to a few megabytes. This limitation restricts operational throughput and the number of parameters that can be accommodated; \textbf{2. Privileged Instructions.} Some TEEs, for security reasons, do not provide access to privileged instructions (e.g. filesystem access, multi-threading) hindering some use cases; \textbf{3. Software Development Kits (SDKs).} TEEs are often populated with simple low-level SDKs.

The TEE, as depicted in Fig.~\ref{fig:tee}, assumes that its host might be compromised, however, assuring that the execution of the procedures assigned to it is not affected. The TEE cannot, however, guarantee that the local data on the client was not corrupted.

Melon~\citep{melon_tee}, focusing on mobile devices, has developed a Secure Learning Framework using a TEE to train deep neural networks distributively.

\begin{notebox}
We invite the reader to explore further the Machine Learning possibilities and limitations of TEE~\citep{ml_tee}.
\end{notebox}

\subsubsection{Multi-party Computation}

To extend privacy guarantees in Federated Learning, secret sharing between parties can be used, in what is called: \textbf{Secure Multi-Party Computation (SMPC)}.

SMPC techniques enable different entities to collaboratively compute a function while ensuring the confidentiality of their respective inputs. In the context of Machine Learning, we can imagine this function being the loss during model training.

While SMPC generally incurs notable communication overhead, its key advantage lies in preserving the privacy of input data, as long as a significant portion of the parties involved are not maliciously coordinating. Using secret sharing, both the parameters of models and the data used in training or inference can be safeguarded. This guardrails that, even a malicious adversarial agent with unlimited time and resources, cannot crack the confidentiality within the input data.

In comparison, Fully Homomorphic Encryption is more computationally expensive than SMPC, even though both allow the execution of mathematical operations over encrypted data.

Google pioneered SMPC implementation over the FL framework~\citep{PracticalSMPC}. They resort to t-out-of-n Secret Sharing~\citep{sharing_a_secret} to ensure that the masking of each client's inputs does not reveal unwanted information. They also introduce a notion of double masking to protect against cases where the server can reconstruct some specific client mask. They use Diffie-Hellman public keys to reduce communication by having the parties agree on common seeds for a pseudorandom generator.

\subsubsection{Secure Aggregation (SE)}
\label{par:se}
\begin{figure}[!htbp]
    \centering
    \includegraphics[clip,width=0.7\linewidth]{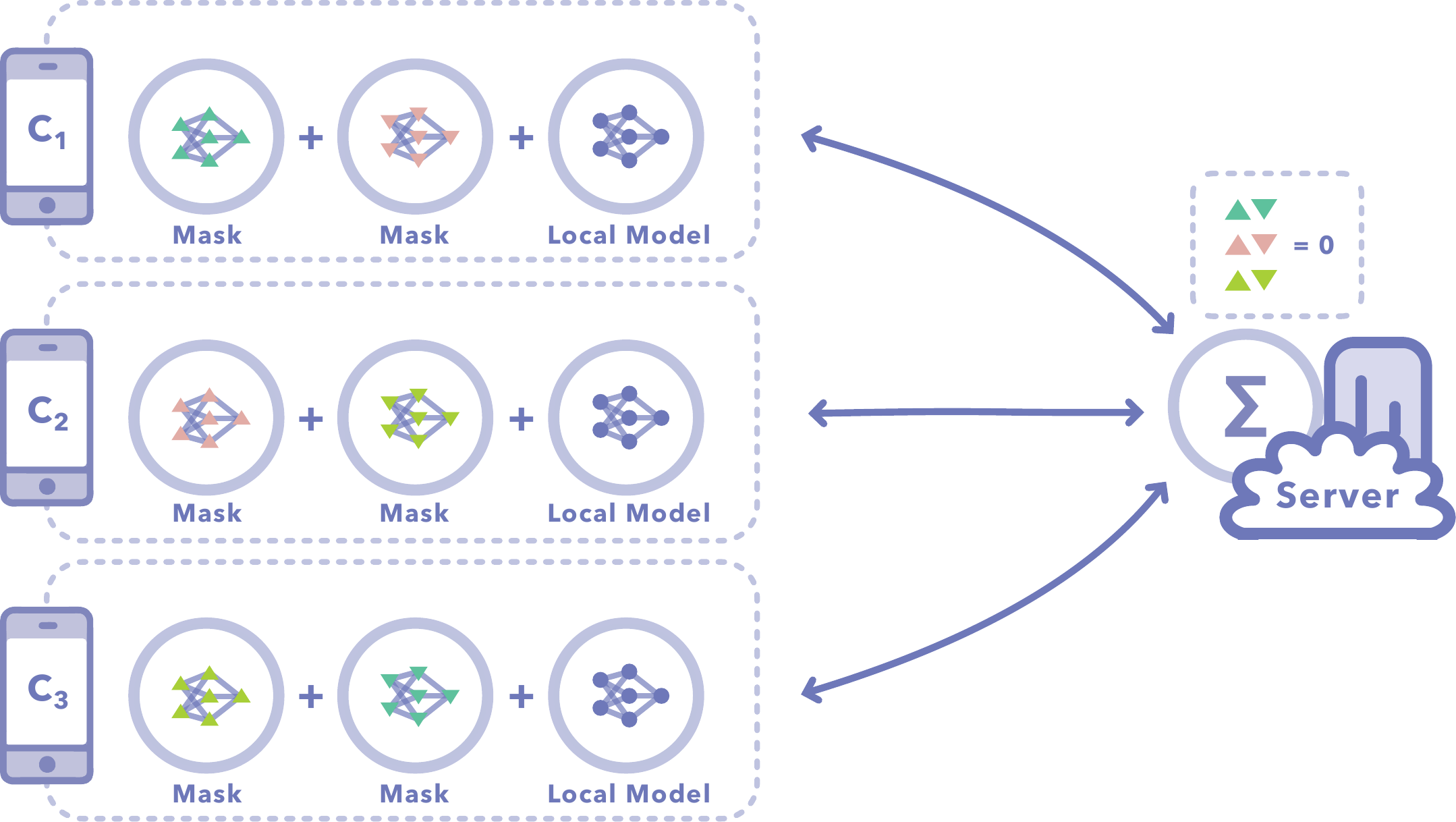}
    \caption{Abstract example of how Secure Aggregation mechanism can be applied to Federated Learning. Each client holds one or more masks which are added to the model weights, conditioned that the sum of all masks from the different clients must add up to 0, so they cancel out.}
    \label{fig:secure_aggregation}
\end{figure}

The notion of Secure Aggregation is based on the idea that malicious agents can try to corrupt a Federated Learning process. 
The Secure Aggregation technique employs additive masking to ensure privacy protection of the model updates shared by clients~\citep{original_se_fl}. Each client adds a uniform random mask to their update, ensuring that each masked update is statistically indistinguishable from the other clients' values. At the time of aggregation, the protocol cancels out all the masks, ensuring that the final model update is an accurate representation of all the client updates while maintaining their privacy (see Fig.~\ref{fig:secure_aggregation}).

SE can be applied through different techniques: \begin{enumerate}
    \item Software-based protocols (e.g. Multi-Party Computation~\citep{PracticalSMPC});
    \item Hardware implementations of Trusted Execution Environments~\citep{PPFL_tee}.
\end{enumerate}

ELSA~\citep{elsa}, for instance, introduces a Secure Aggregation protocol, which utilises two servers to maintain trust (distributed trust) in the learning system. The system guarantees privacy properties as long as at least one server is honest. Unlike RoFL~\citep{RoFL}, EIFFeL~\citep{EIFFeL}, and Prio~\citep{prio}, ELSA stays efficient when dealing with strict bandwidth constraints.

Dropout clients during the training/model aggregation process make the algorithm design of Secure Aggregation considerably more complex. Standard techniques face this problem when the number of dropout clients increases since they rely on the secret sharing of random seeds to further reconstruct/cancel the masks belonging to the dropped clients. To face this malicious issue, LightSecAgg~\citep{LightSecAgg} presents a protocol consisting of moving the masking responsibility to the local clients, which is then followed by the respective mask encoding and its sharing with the other clients, whilst making sure the server can recover the masks upon clients dropouts. It can be used in an asynchronous setting and if the majority of clients persist active in the federation it can increase the resiliency to dropping out clients, whilst reducing the aggregation time.

In Federated Learning settings, the model aggregation process may repeat several times, which may consist of multiple communication overhead if the random-seed usage for masking is not well designed. LERNA~\citep{lerna}, for instance, established a particular initial round for setting up the needed Secure Aggregation configurations to reduce communication costs. Upon a specific pre-defined period, this setup is reset to ensure private-secure guarantees. This method is specially designed for a large-scale interconnected network ($\geq$ 20K clients).

\subsubsection{Blockchain}
Traditional centralised FL systems suffer from inherent infrastructure vulnerabilities, as they rely on a single main server controlling the learning process to remain operational and, thus, they are vulnerable to unauthorised tampering or failures. Apart from classical peer-to-peer decentralised networks, Blockchain can provide properties like tamper-proof, collective maintenance, and traceability in a decentralised setting.

In essence, a Blockchain consists of a chain of blocks, where each block contains a list of transactions, a timestamp, and a reference to the previous block. These blocks are linked and secured using cryptographic hashes, creating an immutable and tamper-resistant system. With the help of consensus mechanisms like \textbf{proof-of-work} or \textbf{proof-of-stake}, participants in a blockchain network validate and agree on the state of the ledger, preventing fraudulent activities. The ledger is the record-keeping system that stores a chronological and transparent transaction history, distributed between multiple blocks.

In the context of FL, the blocks are often designed to have the local training rounds of each client, where each transaction within the block represents each client model parameters. The edge devices assume the responsibility to maintain the Blockchain as ``miners'', receive and store the model parameters, and finally authenticate the parameters by a consensus protocol.

\citet{blockchain_medical_iot} suggests the combination of Blockchain technology with the use of differential privacy, along with model gradients verification step as a move towards a more secure and reliable FL framework.

\begin{notebox}
We encourage the reader to explore the research on Blockchain-based Federated Learning in depth~\cite{blockchain_survey_fl}.
\end{notebox}

\subsubsection{Anomaly Detection}
One effective approach for assessing the status of a FL pipeline involves leveraging anomaly detection techniques. This approach proves invaluable in real-time monitoring, enabling swift identification of any deviations within specific local deployments. Employing anomaly detection facilitates the detection of potential label-poisoning or adversarial attacks.

A Cumulative Sum Control Chart (CUSUM:~\citet{cis}), is a rather simple technique that establishes a threshold of maximum acceptable change in specified weights/model parameters/data distributions.

In the domain of image-based tasks, \href{https://www.microsoft.com/en-us/photodna}{PhotoDNA}, originally developed to combat Child Sexual Abuse Material (CSAM), is a tool that converts images into unique hashes, enabling the identification of similar images within a database. This technology can also be leveraged to detect and track alterations in datasets.

FL-MGVN~\citep{wu2022fl} presented an anomaly detection classification model that uses Mixed Gaussian Variational Self-encoding Networks. It is designed to effectively handle network attacks and sample dissimilarity, addressing key challenges in anomaly detection, including low detection accuracy, high false alarm rates, and the scarcity of labeled data.

For a comparison of threshold-based anomaly detection algorithms, we invite the reader to refer to Table~\ref{tab:anomaly}, where we evaluate multiple approaches. These methods differ in the statistical measures used, data characteristics considered, and techniques applied for threshold calculation.

\begin{table}[htbp]
    \centering
    \caption{Anomaly Detection Federated Methods Comparison. Adapted from ~\cite{anomaly_threshold}.}\bigskip
    \scalebox{0.9}{
    \input{tables/anomaly}
    }
    \label{tab:anomaly}
\end{table}

\subsubsection{Physical Unclonable Functions (PUFs)} PUFs~\citep{ml_systems_tinyml,puf} is a physical/hardware device that can be used to obtain a cryptographic key, which uniquely identifies a specific device. This is possible through the micro-scale physical discrepancies in the fabrication of computer devices, which enables the obtainment of a fingerprint upon specific input stimuli.

A PUF key is intrinsically bound to an individual device excluding any possibility of cloning the key, reverse engineering it or, provided access to it, extracting it. They provide means to watermark an ML model or a model prediction preventing cases where an attacker tries to steal or proclaim ownership.

\begin{notebox}
We invite the readers to review the work of \citet{security_iot_fl} and refer to the Privacy and Security chapter of the Machine Learning Systems
with TinyML book~\citep{ml_systems_tinyml}.
\end{notebox}

\subsection{Privacy Strategy}
\label{sec:privacy}

The world is driving more concern in the ownership and control of one's data. Examples of this are the development of regulations and privacy principles like: \href{https://artificialintelligenceact.eu/ai-act-explorer/}{EU AI Act},
EU General Data Protection Regulation (\href{https://gdpr.eu/tag/gdpr/}{GDPR}), the California Consumer Privacy Act (\href{https://oag.ca.gov/privacy/ccpa}{CCPA}) and the Health Insurance Portability and Accountability Act (\href{https://www.cdc.gov/phlp/publications/topic/hipaa.html}{HIPAA}).

Federated Learning, in its intrinsic form, guarantees a basic level of privacy since raw data is not openly shared between network parties. Conversely, it has been shown that Neural Networks memorise training data~\citep{nn_memorization,text_embeddings_reveal_text} and consequently, this data can be exploited through adversarial attacks on models~\citep{fl_InvertingGradients,DeepLeakage,federated_asr_privacy_issues_examples}. Furthermore, \citet{cluster_fl_privacy} reveals that information about client similarity can be inferred from their weight updates, upon model alignment. Intellectual property is also at risk in Federated Learning, leading to increased interest in \textbf{model watermarking} as a protective measure~\cite{tekgul2021waffle}.

\begin{figure}[!htbp]
    \centering
    \includegraphics[clip,width=0.9\linewidth]{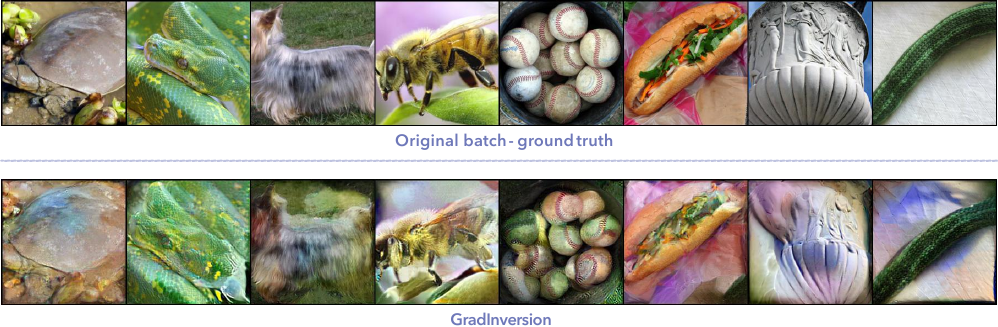}
    \vspace{1mm}
    \caption{Showcasing the capabilities of the Gradient Inversion technique extracted from gradients obtained from a Federated Learning setting. Adapted from GradInversion~\citep{gradients_privacy_issue}.}
    \label{fig:grad_inversion}
\end{figure}

Next, we discuss common privacy-related attacks on Federated Learning systems:

\paragraph{\textbf{Model Inversion Attacks.}} These attacks attempt to reconstruct the raw training data by exploiting access to the model outputs or model parameters. By analysing gradients or model predictions, an adversary can approximate the original input data, potentially revealing private information in the training set. One popular exploit vulnerability is the gradient update vectors since they encode information about the learning function and data. For example, \citet{gradients_privacy_issue} developed a method (GradInversion) to reconstruct the original batches of data from gradients shared from the client to the server (see Fig.~\ref{fig:grad_inversion}).

We can categorise the gradient inversion attack as a \textbf{Data Reconstruction Attack}. Typically, associated with this attack, we can have ``honest-but-curious'' clients/servers, which will behave towards the goal established in the federation, but have a third-party goal like collecting statistics or data theft.

\paragraph{\textbf{Membership Inference Attacks.}} These attacks aim to determine whether a specific data sample was included in the training set. By exploiting divergences in the model behaviour, such as confidence scores or gradient updates, an adversary can infer the presence of individual data points, potentially compromising privacy. MemberShield~\citep{membershield}, for instance, mitigates Membership Inference Attacks risks by training local models with soft-encoded labels to regulate overconfident predictions. It further enhances privacy by incorporating early stopping to prevent unintended memorization. 

\begin{notebox}
We invite the reader to explore in more detail Membership Inference Attacks in the following survey~\citep{membership_survey}.
\end{notebox}

As previously discussed, Federated Learning does not inherently guarantee data privacy, making it susceptible to adversarial attacks. To address these vulnerabilities, a range of techniques has been developed to enhance privacy at different stages of the learning process. Next, we explore these techniques and their role in fortifying privacy-preserving FL systems.

\subsubsection{Differential Privacy} 
\begin{figure}[!htbp]
    \centering
    \includegraphics[clip,width=0.7\linewidth]{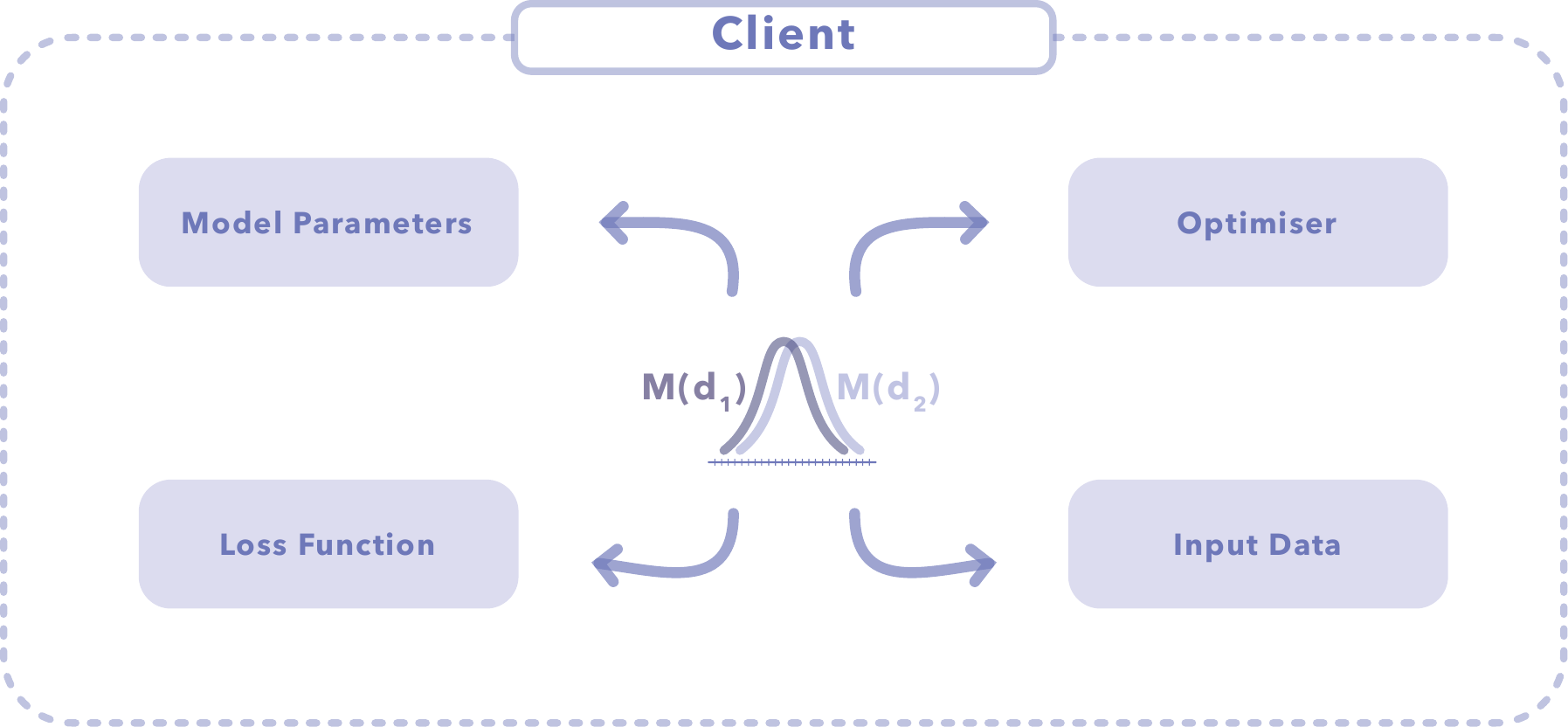}
    \caption{Differential Privacy mechanism and its possible applications (e.g. Model Parameters, Loss Function, Optimiser, Input Data).}
    \label{fig:diff_privacy}
\end{figure}

Differential Privacy (DP)~\citep{DP_genesys} helps obfuscate the danger of exposing raw data through a noise injection procedure (See Def.~\ref{def:dp}) at the cost of forcing the Machine Learning practitioner to trade-off accuracy with data anonymity. 
The injection of noise can be performed over many possible targets: \textbf{1.} Model Parameters; \textbf{2.} Loss Function; \textbf{3.} Optimiser; \textbf{4.} Input Data (See Fig.~\ref{fig:diff_privacy}). 
The main types of differential privacy are $\epsilon-DP$ and $(\epsilon,\delta)-DP$, the latter being a generalisation of the former. $\epsilon$ is a real number defining a threshold of privacy tolerance, whilst $\delta$ is an option to control the probability of the extreme event of data breach occurring. A smaller $\epsilon$ will yield better privacy constraints but result in less accurate models.

\begin{definition}[Differential Privacy]
\label{def:dp}
A noise injection randomised function M: D $\rightarrow$ D' with domain D (e.g. the space of possible training datasets) and codomain D' (e.g. the space of the possible resulting datasets with noise transformation) satisfies ($\epsilon$, $\delta$)-differential privacy provided that for any two adjacent datasets $d_1$, $d_2$ $\in$ D and any subset of outputs S $\subset$ D' it holds that $P[M(d_1) \in S] \leq e^\epsilon P[M(d_2) \in S]+ \delta$ (P: probability; M: randomisation function (e.g. query(dataset) + noise, query(dataset + noise)); $\epsilon$: a positive real number representing the privacy budget; $\delta$: Probability of data accidentally having been leaked).
\end{definition}

The rise of speech-powered companions, such as Siri, Alexa, or Cortana has led to the interest in designing robust Automatic Speech Recognition (ASR). The presence of these assistants on ubiquitous devices makes them a great target for (cross-device) FL applicability. A team by Apple concerned with privacy guarantees proposed training a Transformer with DP, in a Federated Learning setting, proving that a large-scale FL ASR take can have relatively good results when compared with a standard ML approach~\citep{apple_asr_dp}. It also discusses the positive correlation between the cohort sizes and DP success, which suggests that to find a good utility for DP we need a rather big amount of federated clients $>100k$. \citet{clipping_dp_fl} also explores a relevant analysis of the operation of clipping models and gradients, which lies as a base operation to the usage of DP in FL. DP has also been shown to badly impact fairness when a model is faulty already~\citep{dp_vs_accuracy_fairness}.

The concept of Local Differential Privacy~\citep{LDP_FL} has also been introduced in decentralised FL domains to eliminate the need for a central server to manage the privacy budget.

Some popular libraries with off-the-shelf implementations of Differential Privacy are Opacus~\citep{opacus} and Google's DP library~\citep{diff_privacy_google_library}.

\begin{notebox}
For an in-depth exploration of the DP domain, we invite the reader to visit Ted's compilation of articles related to Differential Privacy~\citep{ted_dp_blog}.
\end{notebox}

\subsubsection{Homomorphic Encryption} Homomorphic Encryption (HE)~\cite{he_survey} has been proposed as a privacy-preserving technique in FL to safeguard local raw data by encrypting it, allowing learning computations to be performed directly on encrypted inputs without revealing the underlying data. This ensures that data remains confidential throughout the learning process, even during transmission or aggregation.

Compared to other privacy-preserving approaches, HE introduces a trade-off between computational efficiency, model accuracy, and privacy guarantees. One practical challenge is that most HE schemes operate over integer domains, necessitating quantization of model artefacts (mapping floating point values to integer ones) to enable compatibility with machine learning models.

Moreover, the use of HE in neural networks incurs significant computational overhead due to the sequential nature of neural computations and the requirement for bootstrapping, a costly re-encryption process to manage noise accumulation between operations.

\begin{figure}[!htbp]
    \centering
    \includegraphics[clip,width=0.8\linewidth]{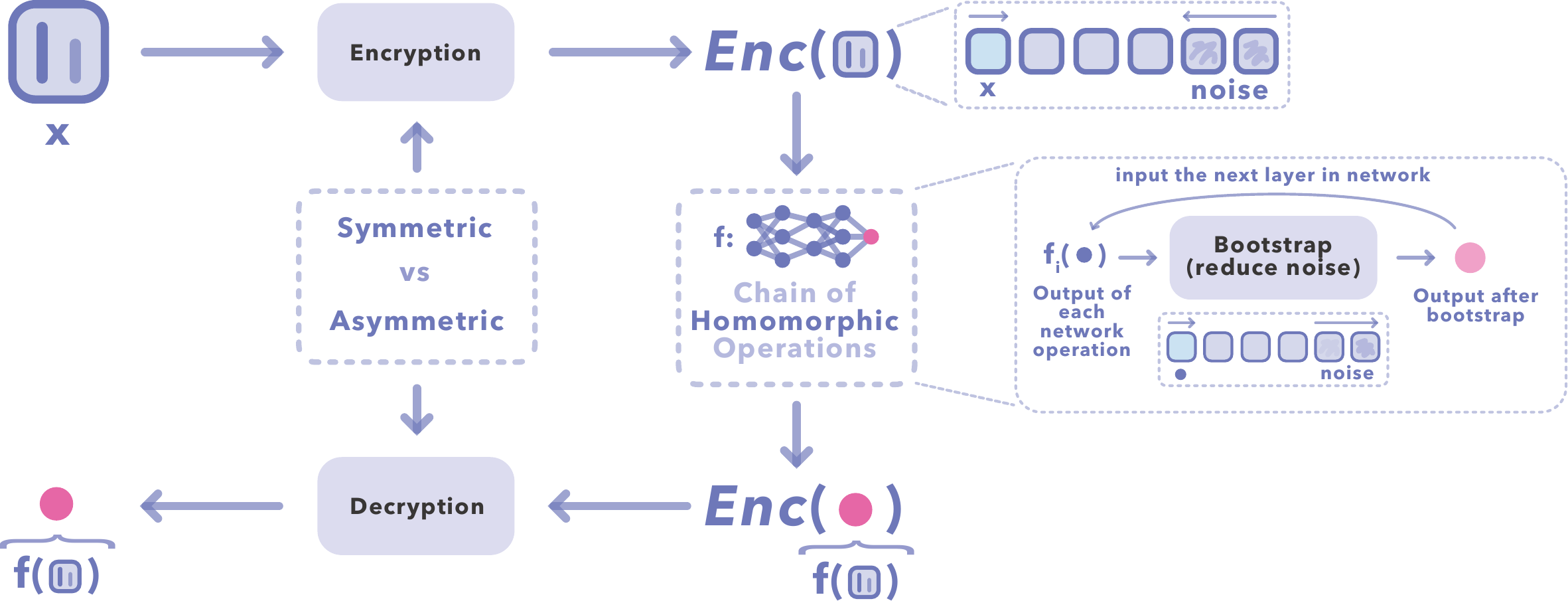}
    \vspace{1mm}
    \caption{We present a FHE framework applied to a neural network use case. Notice that after a homomorphic operation, noise is additionally added to the encrypted input. Above a certain threshold of noise, FHE execution starts deteriorating the correctness of the decryption process, since the input of the operations starts getting corrupted by the increasing levels of noise. The bootstrap method mitigates this issue, by decrypting and encrypting again the input content after an operation or a set of operations.}
    \label{fig:fhe}
\end{figure}

Homomorphic Encryption can be categorised into the following types:

\begin{itemize}
    \item \textbf{Partially Homomorphic Encryption (PHE)}: Allows only a single, predefined operation (either addition or multiplication) to be performed on ciphertexts — but without limits on how many times that operation can be applied. One widely known example of a PHE scheme is the Paillier encryption system.

    \item \textbf{Somewhat Homomorphic Encryption (SHE)}: Extends PHE by supporting both addition and multiplication operations on ciphertexts, but only for a limited number of consecutive operations. This limitation arises due to noise accumulation, which eventually hinders correct decryption.

    \item \textbf{Leveled Homomorphic Encryption (LHE)}: Further extends SHE by allowing users to specify the maximum number of sequential operations (circuit depth) at encryption time. While it removes the fixed limit of SHE, it still does not support unlimited computations.

    \item \textbf{Fully Homomorphic Encryption (FHE)}: The most general and powerful form of HE, supporting arbitrary computations (combinations of additions and multiplications) on ciphertexts. This is achieved using a technique called bootstrapping, which "refreshes" ciphertexts to reduce accumulated noise, enabling unlimited operation depth. Some FHE schemes may impose constraints such as quantization or specific circuit representations (see Fig.~\ref{fig:fhe}).
\end{itemize}

\citet{homomorthic_encryption_ffl} suggests a new weighted aggregation scheme, which leverages Homomorphic Encryption by encrypting the gradients, preventing the server from inferring private information, while still providing comparable accuracy with reasonable computational costs. The algorithmic complexity in this methodology lies in a single server, in contrast with previous works where 2 servers are often needed~\citep{two_server_he}. It also seems to assure better security characteristics, while holding less communication overhead than previous works ~\citep{ShieldFL}.

One popular and current standard implementation zoo for Fully Homomorphic Encryption is OpenFHE~\citep{OpenFHE}. Concrete~\citep{Concrete} is an FHE Compiler, which compiles operations to perform homomorphic evaluation over them.

\subsubsection{Zero-Knowledge Proofs}

Zero-Knowledge Proofs (ZKPs)~\cite{zkp_genesys} are cryptographic protocols that allow a party (the prover) to convince another party (the verifier) that a certain statement is true without revealing any additional information. In the context of FL, ZKPs can be used to verify that a client has correctly trained a model or followed the agreed-upon protocol, without requiring them to expose their local data or intermediate model updates. A ZKP must satisfy three fundamental properties: \textbf{completeness}, \textbf{soundness}, and \textbf{zero-knowledge}. They can be interactive or non-interactive, with zk-SNARKs and zk-STARKs being widely adopted non-interactive schemes that balance efficiency and scalability~\citep{ernstberger2024you}.
While ZKPs enhance FL privacy and security, they introduce computational overhead, particularly in proof generation and verification. Efficient ZKP schemes such as zk-STARKs help address this, but scalability remains an issue, particularly with large-scale FL models.

\paragraph{\textbf{Privacy-Preserving Model Updates.}}
Clients train local models and share updates (gradients or weights) with a central aggregator. However, these updates may inadvertently leak sensitive data. ZKPs can be used to prove correctness of updates without exposing the underlying gradients, mitigating gradient inversion attacks \citep{wang2024zkfl}.

\paragraph{\textbf{Decentralised Federated Learning.}}
Traditional FL relies on a centralised aggregator, which may act dishonestly. ZKPs can enable decentralised FL, where clients submit model updates along with zero-knowledge proofs. This removes the need for a trusted server, making FL more robust to single points of failure~\citep{decentralized_reputation}.

\paragraph{\textbf{Secure Aggregation with ZKPs.}}
Secure aggregation ensures that individual model updates remain hidden while allowing their aggregation to be used for training. ZKPs can be combined with secure aggregation protocols to:
\begin{itemize}
\item Prove that an individual update was computed correctly before aggregation.
\item Ensure aggregation is performed honestly without exposing raw data.
\end{itemize}
This improves end-to-end privacy guarantees in FL~\citep{decentralized_reputation}.

\paragraph{\textbf{Fair Model Contribution and Reward Mechanisms.}}
In collaborative FL, different participants may contribute varying levels of effort. ZKPs can be used to prove that each client has performed a minimum amount of computation before receiving updates. This prevents free-riding and ensures fairness in incentive mechanisms \cite{fairproof}.

\input{tables/privacy_security_comparison}

\begin{notebox}
We encourage the reader to explore the topic of federated learning privacy in the survey by ~\cite{decentralized_fl_privacy}. Furthermore, we direct the reader to Fig.~\ref{tab:frameworks_comparison_data_privacy}, which provides a comprehensive overview of several widely used Federated Learning algorithms, concerning privacy.
\end{notebox}

\subsection{Trustworthy Federated Learning}
\label{sub_sec:trustworthy}

Federated Learning (FL) is mostly considered in critical domains, with concerns about the privacy feasibility characteristics. The bare consideration and usage of FL is a step towards Trustworthy Machine Learning (ML). Trustworthy ML is a paradigm in the Machine Learning community to advance the field considering the real applicability of the designed models. This means that Trustworthy ML normally considers: how the model behaves with Out-Of-Distribution (OOD) data; Leverages simplicity and explainable models to help reasoning and support decision-heavy use-cases; Explores the uncertainty distributions under model predictions; Meditate on the effect of noisy contamination to the data and how we can build robust models to such adverse conditions; Are constructed having in consideration they can have a real supportive impact on the decision making of a critical process. Real-world applications need this thoughtful and accountable design. In the European Union, we already observe a direction towards a regulation on trustworthy Machine Learning (\href{https://www.europarl.europa.eu/news/en/press-room/20231206IPR15699/artificial-intelligence-act-deal-on-comprehensive-rules-for-trustworthy-ai}{Trustworthy AI EU Act}).

\subsubsection{Dynamic Learning}
\label{sub_sub_sec:dynamic_fl}

FL is often studied under the assumption of a homogeneous and static environment. However, in real-world scenarios, FL systems must adapt to dynamic and evolving conditions, leading to the concept of \textbf{Dynamic Federated Learning (DFL)}. This paradigm addresses key challenges such as out-of-distribution (OOD) data, client data drift, and operational shifts in client participation.

DFL research explores methods to improve model robustness against these variations. For instance, \citet{DynamicFL} proposes an adaptation of the FedAvg algorithm that dynamically adjusts local epoch sizing and normalization while incorporating a random-walk parameter drifter, enhancing resilience to data distribution shifts.

\begin{figure}[!htbp]
    \centering
    \includegraphics[clip,width=0.9\linewidth]{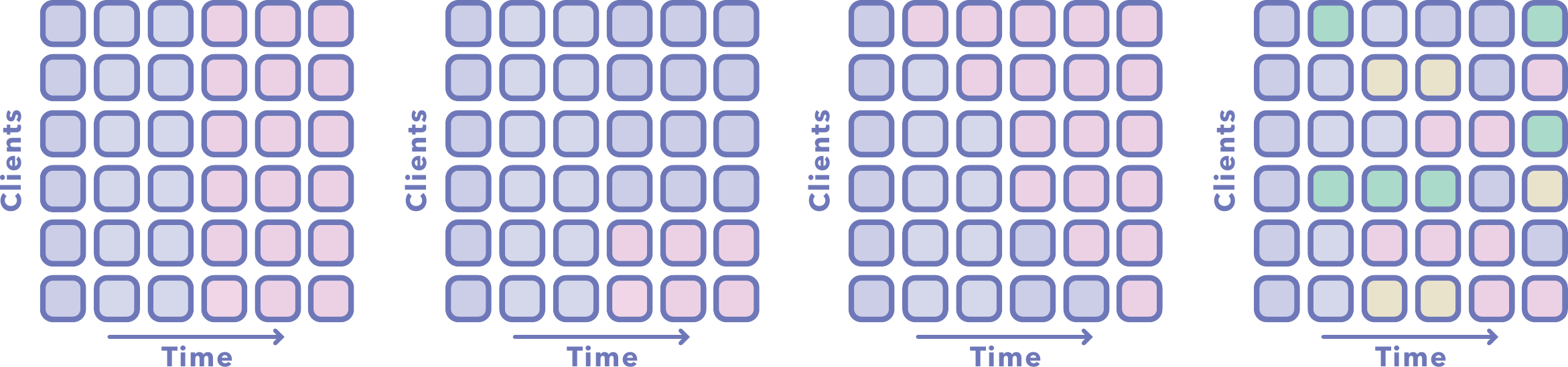}
    \caption{We present four different time-evolving types of data distribution shifts occurring in a Federation, also known as Clients Concept Drift. Each colour represents a different concept distribution. Adapted from \citet{fl_dist_concept_drift}.}
    \label{fig:concept_drift_types}
\end{figure}

The notion of client's concept drift can be defined as a change in the client's data distributions (e.g. new target labels in a client appear and others oscillate in magnitude. See Fig.~\ref{fig:concept_drift_types} for an illustrative evolution of four types of client concept drift).

Consequently, \textbf{drift detection mechanisms} are important either to prompt the FL infrastructure to incrementally learn the new data distribution and avoid decay in performance or to monitor and debug possible problems in the FL infrastructure. Concerned with clients' data drifts,  \citet{fl_dist_concept_drift} introduce FedDrift and FedDrift-Eager, which are FL methods to train groups of models which share the same concepts distribution. In a sense, this work is similar to clustered FL, since it combines similar clients and each group trains a global model.

\begin{figure}[!htbp]
    \centering
    \vspace{0.5cm}
    \includegraphics[clip,width=0.7\linewidth]{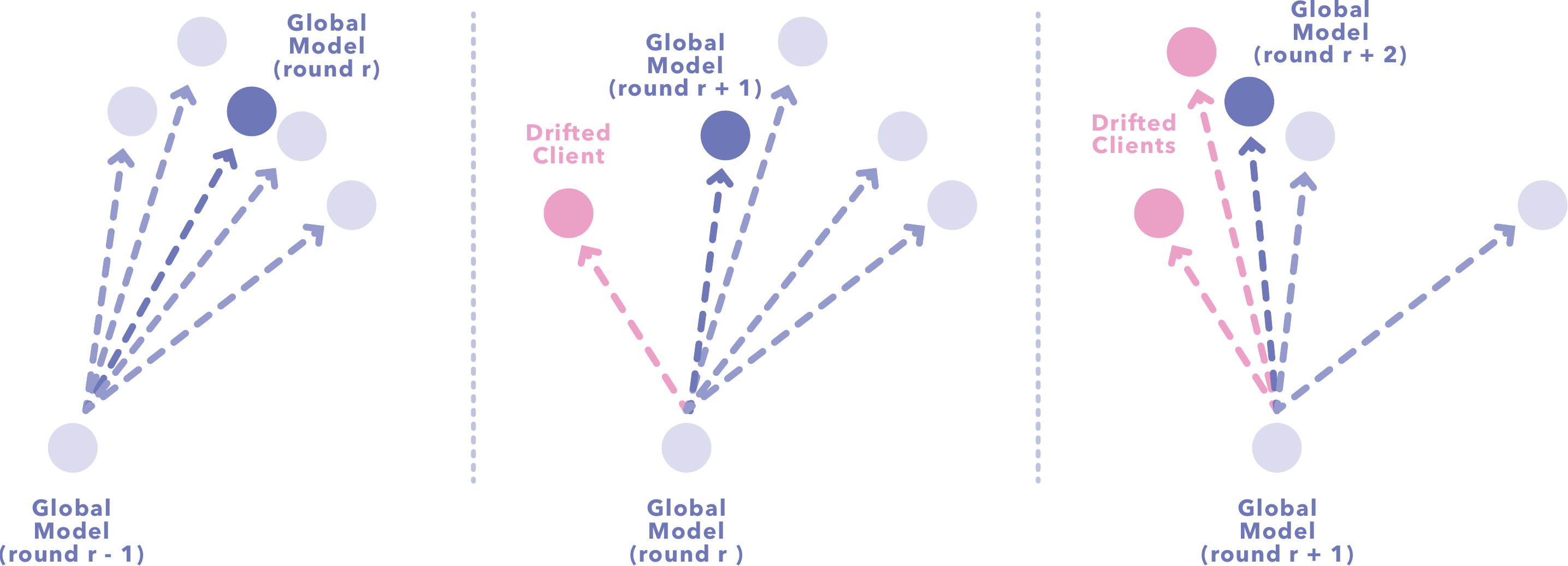}
    \caption{Clients Models drifting when their intrinsic concept distribution changes.}
    \label{fig:concept_drift}
\end{figure}

Understandably, these client's distribution shifts cause fluctuation in the overall behaviour of the Federated Learning mechanism (see Fig.~\ref{fig:concept_drift} to observe the effect on the global model, when clients start having distribution shifts).

\subsubsection{Bayesian Learning}
\label{sub_sub_sec:bayesian_fl}

Bayesian methods offer a principled approach to uncertainty estimation and model interpretability, making them particularly relevant in multi-client cooperative learning settings. In the context of Federated Learning (FL), Bayesian techniques have been explored to enhance robustness and address overfitting issues arising from data scarcity at client nodes.

Bayesian Neural Networks (BNNs) have been integrated into FL frameworks by representing model parameters as probability distributions rather than fixed values. While this approach helps mitigate overfitting, its effectiveness diminishes in the presence of non-iid data, where parameter distributions struggle to generalize across heterogeneous client datasets~\citep{BayesianNF_fl}.

The pFedGP~\citep{PersonalizedFL_GP} was proposed as adapting Gaussian Processes (GPs) to multiple clients, where a shared deep kernel function~\footnote{In Regression problems, features normally map well with common kernel functions, like RBF or Materne. However, when dealing with unstructured data like images on Bayesian classification problems, Deep Kernel Learning~\citep{DeepKL} helps in creating a better compressed and pattern-aware representation of an image, which is fundamental in the good behaviour of a GP.} is learnt globally through the clients and inducing point methods are introduced to reduce the computational overhead. They implement multi-class classification GPs as a pipeline of binary decision trees, where each tree represents a binary GP classifier. These GP models are locally built at every communication round and averaged at the server. They prove this method has non-vacuous guarantees and that it is advantageous when clients lack in data quantity and follow heterogeneous data distributions.

Instead of assuming priors to all model parameters, like in a standard BNN, pFedBayes~\citep{PersonalizedFL_VBI} uses the global distribution being learnt as a probabilistic prior for the local client models.

\begin{figure}[!htbp]
    \centering
    \includegraphics[clip,width=0.9\linewidth]{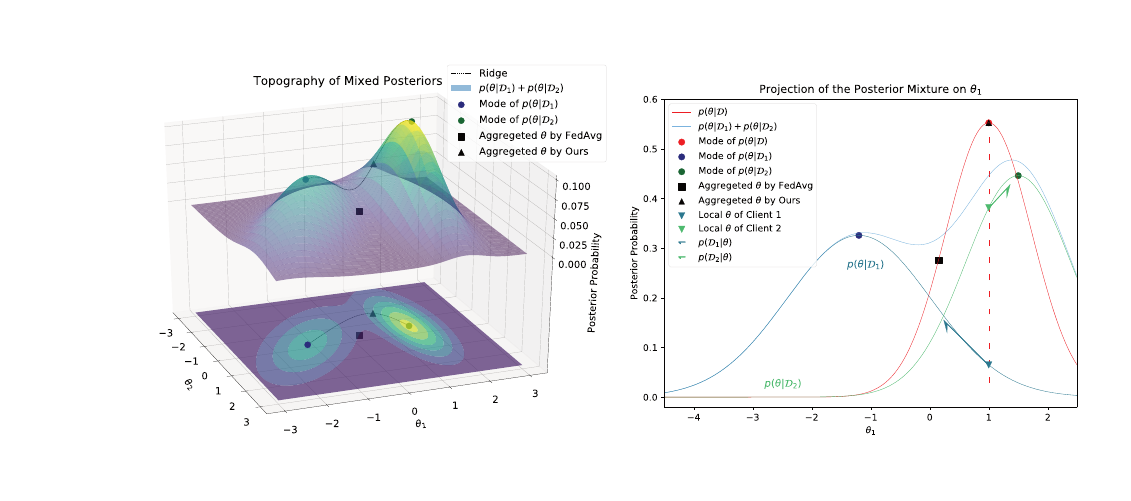}
    \caption{Two heterogeneous clients being modelled with Gaussians, depicting how poorly the FedAvg algorithm can be for aligning heterogeneous clients, in comparison with the FL Bayesian framework~\citep{bayesian_fl_framework}.}
    \label{fig:bayesian}
\end{figure}

\citet{bayesian_fl_framework} presents a Bayesian FL framework, which models clients using Gaussian distributions and showcases how aligning clients' knowledge with the FedAvg algorithm can move the global model to a non-optimal parameter space (see Fig.~\ref{fig:bayesian}).

\begin{notebox}
We invite the reader to explore a comprehensive overview of Bayesian methodologies in Federated Learning in ~\cite{bayesian_fl_survey}.
\end{notebox}
\subsubsection{Noisy Learning}
\label{sub_sub_sec:noisy_labels}
Literature on FL, generally, assumes the data present at the local computation silos are already labelled and additionally that they show no anomalies in their intrinsic distribution. However, in real scenarios, seldom is the data labelled and even when it is, it is reasonable to assume that some noise artefacts may be affecting their quality.

Realistic data normally assumes the form:

\begin{equation}
\label{eq:noisy_local_dataset}
    D_c=\left\{\left(x_{i}, \tilde{y}_{ i}\right)\right\}_{i=1}^{M_c},
\end{equation}

 where labelled annotations have suffered a transformation through a composed function over the original labels/target outputs.

Noisy labels, the result of a composed function which is intrinsically applied to $y$ noted as $\tilde{y}$, can normally be characterised in three distinct ways:
\begin{enumerate}
    \item Random process. An uncontrolled and unknown function inevitably alters the Labels, independently of the corresponding data point;
    \item Y-Dependent Noise $P(\tilde{y} \mid y, x)=P(\tilde{y} \mid y)$. There is the assumption that a perturbation function gets applied to the data, where the new label is conditioned on the previous $y$ label:
    \item XY-Dependent Noise. The most complex assumption relies upon a noise function where there is an influence of both the features and label value conditioning the new label $P(\tilde{y} \mid y, x)$.
\end{enumerate}

The intrinsic nature of distributed learning across different devices involves varying computational resources. The annotated data may be obtained from machine-generated pseudo-labels or distinct human interactions, explicit or implicit, with varying levels of expertise. This diversity can lead to an increase in faulty annotations: noisy labels.

An undesirable issue often observed in Machine Learning processes is the presence of noisy labels in the training data. Their presence in the learning mechanism invokes the following behaviour: First, typically, a model learns the standard/unadulterated data, and lastly, after many learning steps the model starts overfitting to encompass the local noisy space deformations~\citep{noisy_lid}.

\textbf{Local Intrinsic Dimensionality} is a score attributed to data that models the complexity of dimensionality of samples, which consequently helps segregate noisy labels from standard ones~\citep{noisy_lid}.

Systematically, Noisy Label learning literature applies the following techniques: \textbf{1.} Study Robust Learning, where a Robust Loss function accounts for noise errors \textbf{2.} Directly manipulate the labelled data based either on a parametric/non-parametric model, while assuming specific noise rates.

These works target the creation of Robust mechanisms to overcome the effect of Noisy Labels in the Machine Learning process. However, it is also possible to identify, to a certain point the noisy labels present in a dataset and correct them to fix unwanted bias and fairness in the original data~\citep{analsyis_noisy_labels_correction}.

In the FL setting, for instance, FedCorr~\citep{FedCorr} has been presented to deal with heterogeneous environments and correct local client noisy labels.

RoFL~\citep{RoFL} bases their work on a star-schema FL setting, where each client node shares both the model weights and the centroids of each label class and is also inspired by noisy label correction techniques. The aggregation step is performed through a similarity-based summation of the local centroids method. Though, RoFL, assumes and focuses on clients who share different noise distributions, they lack the exploration of inevitable non-iid data.

Symmetric Cross Entropy Learning (SL:~\citet{SymmetricCE}) is a robust loss mechanism, which weights the correction of the prediction versus the target label with a term that helps to learn hard classes in a noise-tolerant way.

RHFL~\citep{RobustHFL} assumes no model architecture protocol between clients (heterogeneous modelling) and knowledge between models is shared through a common public dataset available to all clients and a knowledge distillation technique to align the different knowledge distributions. To overcome noisy inputs, locally it utilises the noise-robust Symmetric Cross Entropy loss and between other clients, it ranks the reliability/quality of other clients' data.

Assuming noisy annotated data and heterogeneous Clients, FEDCNI~\citep{Wu2023LearningCI} proposes the usage of a prototypical noise detector to distinguish noisy samples, a pseudo labelling method and a denoise Mixup training strategy. First, on a client, they obtain the model computed features of a data point and through a similarity metric they classify, using a Gaussian Mixture Model, the quality of the provided annotated label. Then they use a pseudo-labeling method to correct the labels. Finally, they use a robust aggregation method, where until a specific timestep they operate a FedAvg aggregation scheme but later they weigh each client update based on their noise rate.

Based on user feedback from mobile application interfaces~\citep{amazon_noisy_feedback}, research has explored modelling feedback noise using a parametric approach. This method assigns user-specific probabilities to correctly providing positive and negative feedback. By adjusting these probabilities, it becomes possible to simulate different user behaviours, including well-behaved and adversarial users.

\subsubsection{Fairness}
\label{sub_sub_sec:fairness}

Any learning process exploits the intrinsic nature of bias, which can have a positive or negative impact. A trustful Machine Learning system must weigh such bias nature upon its design to avoid discrimination/lack of representativity, among other issues. For instance, the FedAvg~\citep{FedAvg} algorithm has been shown to lead to unfair results when clients' local data distributions depart from each other~\citep{non_iid_issues_fl}, due to clients' weight divergence.

Specifically targeting FL, fairness is exploited in the contexts of Client Selection and Fair Optimisation.

\paragraph{\textbf{Fair client selection.}}

In Federated Learning, often only a cohort of the total clients from a federation is selected, either as a consequence of their hardware resources limitations, availability, performance or secure breaches suspicions.

FL Dropout~\citep{fl_dropout}, concerned with communication bandwidth costs and inspired by the popular Deep Learning ``Dropout'' technique developed a framework to locally train sub-models of the global model to reduce the transfer size footprint. In this setting, each client does not know the global model architecture, therefore it trains its local sub-model and sends its update to a server which consequently aligns these sub-model updates back to the global model,

Engineered solutions, which through some key attributes provide preference to selecting some clients over others to overcome: learning contamination, such as Contra~\citep{contra}; faulty convergence
Power-of-Choice~\citep{power_of_choice} can put at risk their fairness nature. Balancing good performance with fairness characteristics is therefore a challenge.

\paragraph{\textbf{Fair optimisation.}}

Federated optimisation algorithms, directly or indirectly introduce unfair biases. Considering the local clients' parameters, upon a round of training and the error measures, the respective adjustments (gradients) to be performed on each client's parameters to lower the errors can be substantially different for each client. 

FedFV~\citep{fl_fair_averaging} explores the paramount behaviour of the difference of adjustments (gradients) for each client parameter, both in magnitude and direction. FedFV does not communicate parameters, but rather the gradients (the changes necessary to improve the error considering each parameter). This work uses the Gradient Projection technique to resolve Cohort and global gradient conflicts better.

\begin{notebox}
We encourage the reader to explore the topic of fairness in Federated Learning in greater depth through the survey~\cite{fairness_survey}.
\end{notebox}

\subsubsection{Informed Learning}
\label{sub_sub_sec:informed_fl}

One of the most prominent use cases of Federated Learning lies in the cooperation of smartphone applications. In such scenarios, even when data annotations are absent, user feedback pseudo-labels~\citep{amazon_noisy_feedback} are often available masked with some uncertain noise, which is useful for supervised learning tasks. In contrast, in Federated Learning the availability of annotated data is scarce. One of the main reasons relates to the nature of the data produced in edge devices following a streaming fashion with a high-velocity rate.

To address the challenge of unlabelled data and enhance the robustness of Machine Learning frameworks, a promising avenue of research is to incorporate explicit knowledge through Informed Machine Learning. Informed Machine Learning can have many flavours: constraining parameter solutions to physical laws or geometric rules, constraining parameter space to pre-defined logical rules about the world. Similarly, the mechanism for injection explicit noise can be performed in many ways: domain knowledge input data, loss function, and specialised neural network architecture blocks.

In FL, \textbf{Physics-Informed Neural Networks} have been used, for instance, to 
estimate the state of vehicle traffic~\citep{PI_FL_traffic_estimation}, integrating traffic flow physical theory in the model losses and using Cell Transmission Models. F-MADRL~\citep{Li2022FederatedMD} is a reinforcement learning solution for Multi-Microgrid energy management, which leverages physically informed rewards to score the actions being learnt. This work stands apart from other reinforcement learning studies by conscientiously addressing the inherent physical constraints of the problem.

Concerning \textbf{Neuro-Symbolic AI}, FedSTL~\citep{formal_logic_fl} explores temporal logic reasoning in an FL setting, where each client learns local concept properties and shares these concepts along with the learnt model.  
LR-XFL~\citep{LR_XFL}, also in the symbolic domain, researches explainable properties of FL frameworks using symbolic mechanisms to capture logic rules from entropy-based linear layers~\citep{EntropybasedLE}.

\subsubsection{Visualisation}
\label{sub_sub_sec:visualisation}

There is an underlying interest in providing visual cues about the learning process of complex systems (e.g. Federated Learning). For example, privacy enhancement techniques like differential privacy are often integrated into FL optimisation, which can further reduce explainability and limit confidence in the results.

HetVis~\citep{HetVisAV} has developed a pipeline visualisation framework, facilitating the introspection of horizontal Federated Learning applications in 3 distinct phases (model learning, predictions comparison, and heterogeneity examination).

More focused on data privacy implications, \citet{Guo2023SeeingIB} developed a framework allowing data owners to visually analyze privacy risks in FL.

\begin{notebox}
    To conclude the FL Trustworthiness section, an essential foundation for real-world applications, we encourage the reader to explore a more in-depth survey on trustworthy Federated Learning~\citep{trustworthy_fl_survey}.
\end{notebox}

\subsection{Final Reflections on the Meta-Framework Perspective}
\label{sub_sec:wrap_up}

\begin{figure}[!htbp]
    \centering
    \includegraphics[clip,height=0.9\textheight]{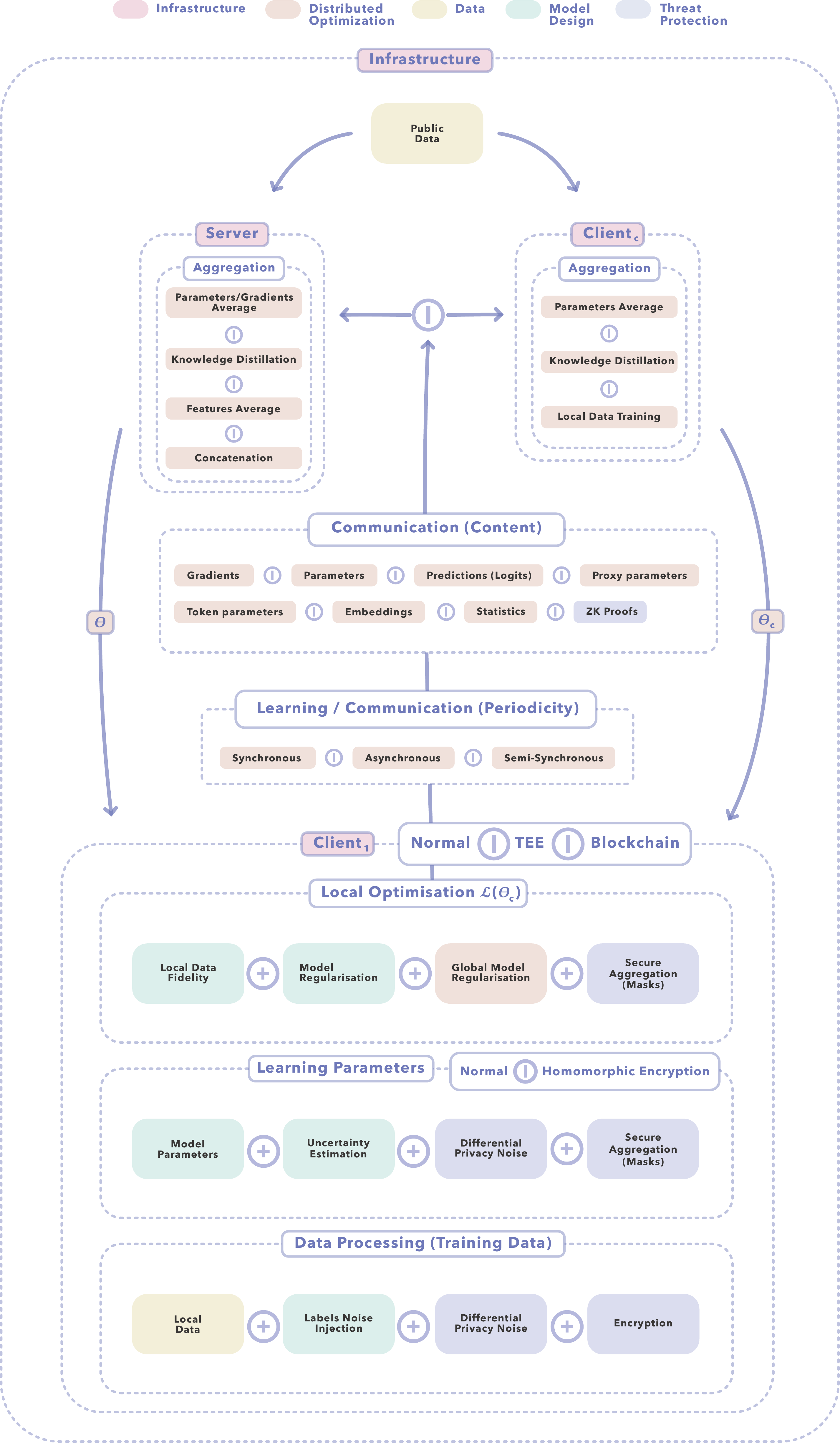}
    \caption{An architectural overview of the modular options for building an FL framework is presented. The \textbf{+} operator indicates the possibility of stacking modules, while the \textbf{|} operator represents a modular design choice. Although we aimed to cover many modular possibilities in this diagram, it is not exhaustive (e.g. Differential Privacy can be applied to parameters, loss functions, or optimisers, among others).}
    \label{fig:toolbox_concrete_overview}
\end{figure}

In this section, we have provided a detailed review of the literature addressing each sub-module within our modular Meta-Framework perspective. We proposed that Federated Learning can be viewed as a modular system composed of interchangeable and complementary components. To illustrate how these modules integrate within the broader FL framework, we refer the reader to Fig.~\ref{fig:toolbox_concrete_overview}. We hope this visual representation facilitates a deeper understanding of FL as a structured and adaptable framework, where different methods can be combined to enhance key properties such as security, privacy, trustworthiness, efficiency, and convergence.

%% file: tables/sync_comparison.tex
\scalebox{0.8}{
\begin{tabular}{lccccc}
\toprule
\textbf{Protocol} & \makecell{Processing\\ Cost} & \makecell{Communication\\ Cost} & \makecell{Energy \\ Cost} & Iddle-free &  Stale-free \\ \midrule
Synchronous   & High & Low & High & \textcolor{red}{$\times$} & \textcolor{darkgreen}{$\checkmark$}            \\
Asynchronous   & Low & High & Medium & \textcolor{darkgreen}{$\checkmark$} & \textcolor{red}{$\times$}            \\
Semi-Synchronous   & Low & Low & Low & \textcolor{darkgreen}{$\checkmark$} & \textcolor{darkgreen}{$\checkmark$}            \\
\bottomrule
\end{tabular}
}

%% file: tables/datasets.tex
\begin{longtable}{@{}m{7cm}m{8cm}@{}}
\caption{Overview of datasets and benchmarks used in Federated Learning research.}
\label{tab:datasets_benchmarks} \\

\toprule
\textbf{Dataset/Benchmark} & \textbf{Key Features/Applications} \\ \midrule
\endfirsthead

\multicolumn{2}{c}{{\tablename\ \thetable{} -- continued from previous page}} \\ \toprule
\textbf{Dataset/Benchmark} & \textbf{Key Features/Applications} \\ \midrule
\endhead

\midrule \multicolumn{2}{r}{{Continued on next page}} \\ \midrule
\endfoot

\bottomrule
\endlastfoot

\multicolumn{2}{l}{\textbf{Image Recognition and Classification}} \\ \midrule
FASHION-MNIST~\citep{FashionMNIST} & Handwritten fashion item images for classification tasks. \\

\addlinespace[2pt]

\cdashline{1-2}[1pt/1pt]
\addlinespace[2pt]
CIFAR-10, CIFAR-100~\citep{CIFAR10} & Object classification datasets, extended in CIFAR-N~\citep{CIFAR_N} to include noisy labels. \\ \cdashline{1-2}[1pt/1pt]
ImageNet~\citep{ImageNet} & Large-scale image classification dataset for deep learning models. \\
\addlinespace[2pt]
\cdashline{1-2}[1pt/1pt]
\addlinespace[2pt]
CINIC-10~\citep{cinic10} & Combines CIFAR-10 and ImageNet to create a more diverse dataset. \\ 
\addlinespace[2pt]
\cdashline{1-2}[1pt/1pt]
\addlinespace[2pt]
CLOTHING-1M~\citep{CLOTHING1M} & Fashion image dataset with noisy labels. \\ \midrule

\multicolumn{2}{l}{\textbf{Natural Language Processing (NLP)}} \\ \midrule
LEAF~\citep{leaf_benchmark} & General benchmark supporting sentiment analysis, language modeling, and meta-learning. \\ 
\addlinespace[2pt]
\cdashline{1-2}[1pt/1pt]
\addlinespace[2pt]
FEDLEGAL~\citep{fedlegal} & Real-world FL benchmark for legal NLP tasks. \\ 
\addlinespace[2pt]
\cdashline{1-2}[1pt/1pt]
\addlinespace[2pt]
FedLLM-Bench~\citep{fedllm} & Designed for evaluating FL methods on large language models. \\ \midrule

\multicolumn{2}{l}{\textbf{Speech and Audio Processing}} \\ \midrule
LibriSpeech~\citep{asr_librisspeech_dataset} & Large-scale automatic speech recognition dataset. \\ 
\addlinespace[2pt]
\cdashline{1-2}[1pt/1pt]
\addlinespace[2pt]
Common Voice v13.0~\citep{asr_commonvoice_dataset} & Diverse speaker dataset for evaluating FL systems in ASR tasks. \\ \midrule

\multicolumn{2}{l}{\textbf{Noisy Learning Datasets/Benchmarks}} \\ \midrule
CIFAR-N~\citep{CIFAR_N} & An extension of CIFAR with corrupted target labels for noisy data research. \\ 
\addlinespace[2pt]
\cdashline{1-2}[1pt/1pt]
\addlinespace[2pt]
FedNoisy~\citep{FedNoisy} & Benchmark designed to evaluate FL techniques on non-iid, noisy data. \\ 
\addlinespace[2pt]
\cdashline{1-2}[1pt/1pt]
\addlinespace[2pt]
CLOTHING-1M~\citep{CLOTHING1M} & Fashion dataset with noisy labels. \\ \midrule

\multicolumn{2}{l}{\textbf{Security and Adversarial Environments}} \\ \midrule
TON\_IoT~\citep{ton_iot_dataset} & IoT dataset for simulating network intrusion and attacks. \\ 
\addlinespace[2pt]
\cdashline{1-2}[1pt/1pt]
\addlinespace[2pt]
Kitsune~\citep{Kitsune} & Network attack dataset for adversarial FL research. \\ 
\addlinespace[2pt]
\cdashline{1-2}[1pt/1pt]
\addlinespace[2pt]
FLHetBench~\citep{flhetbench} & Benchmarks device and state heterogeneity in FL systems. \\ \midrule

\multicolumn{2}{l}{\textbf{Edge Computing}} \\ \midrule
FLEdge~\citep{fledge} & Benchmarks FL workloads in edge computing, focusing on hardware heterogeneity and energy efficiency. \\ \midrule

\multicolumn{2}{l}{\textbf{Wearable and IoT Applications}} \\ \midrule
FitRec~\citep{fitrec_dataset} & Tracks physical activity attributes like heart rate and GPS. \\ 
\addlinespace[2pt]
\cdashline{1-2}[1pt/1pt]
\addlinespace[2pt]
ExtraSensory~\citep{ExtraSensory_dataset} & Smartphone-based activity recognition dataset. \\ \addlinespace[2pt]
\cdashline{1-2}[1pt/1pt]
\addlinespace[2pt]
Anguita et al.~\citep{Anguita2013APD} & Human activity recognition dataset capturing six activity types. \\ \midrule

\multicolumn{2}{l}{\textbf{Multimodal Learning}} \\ \midrule
FedMultimodal~\citep{fedmultimodal} & Benchmark for multimodal FL applications, covering five representative tasks. \\ \midrule

\multicolumn{2}{l}{\textbf{Vertical Federated Learning}} \\ \midrule
PyVertical Framework~\citep{PyVertical_mnist} & Simulates User ID Alignment issues using modified MNIST. \\ 
\addlinespace[2pt]
\cdashline{1-2}[1pt/1pt]
\addlinespace[2pt]
MedPerf~\citep{med_perf} & Medical benchmark testing FL models on edge devices in realistic scenarios. \\ 
\addlinespace[2pt]
\cdashline{1-2}[1pt/1pt]
\addlinespace[2pt]
hERG dataset~\citep{FedGraphNN} & Molecular biophysics dataset for FL systems focused on cardiac toxicity prediction. \\

\end{longtable}

%% file: tables/anomaly.tex
\begin{tabular}{p{7cm}cccc}
\toprule
\textbf{Thresholding Method} & \textbf{\makecell{Anomalies in  \\ Threshold Calculation}} & \textbf{Statistics Used} & \textbf{\makecell{Local Data Distribution \\ Consideration}} \\ \midrule
\citet{anomaly_threshold} & \textcolor{darkgreen}{$\checkmark$} & Mean, Variance, etc. & \textcolor{darkgreen}{$\checkmark$}  \\
Fed-MinMax~\citep{fed_min_max}      & \textcolor{darkgreen}{$\checkmark$} & Min/Max  & \textcolor{red}{$\times$}  \\
Fed-MSE-StD~\citep{fed_mse_std}     & \textcolor{red}{$\times$} & Mean, StD  & \textcolor{red}{$\times$}  \\
Fed-Filtered~\citep{fed_filtered}    & \textcolor{red}{$\times$} & Mean, StD  & \textcolor{red}{$\times$}  \\
\bottomrule
\end{tabular}

%% file: tables/privacy_security_comparison.tex
\begin{table}[h]
    \centering
    \renewcommand{\arraystretch}{1.1} %
    \caption{Comparison of research works on privacy and security guarantees. Adapted from~\cite{trusted_zkp}.}
    \label{tab:fl_comparison}
    \resizebox{\textwidth}{!}{ %
    \begin{tabular}{p{3.5cm}ccccccc} %
        \toprule
        \textbf{Representative Works} & \textbf{Model Privacy} & \textbf{Poisoning Resilience} & \textbf{Malicious Aggregator} & \textbf{Efficient Aggregation} & \textbf{Integrity Check} & \textbf{Inter-client Aggregation} & \textbf{Server Fault Tolerance} \\
        \midrule
        FedAvg~\cite{FedAvg} & \textcolor{red}{$\times$} & \textcolor{red}{$\times$} & \textcolor{red}{$\times$} & \textcolor{darkgreen}{$\checkmark$} & \textcolor{red}{$\times$} & \textcolor{red}{$\times$} & \textcolor{red}{$\times$} \\
        
        Krum~\cite{krum} & \textcolor{red}{$\times$} & \textcolor{darkgreen}{$\checkmark$} & \textcolor{red}{$\times$} & \textcolor{darkgreen}{$\checkmark$} & \textcolor{red}{$\times$} & \textcolor{darkgreen}{$\checkmark$} & \textcolor{red}{$\times$} \\

        Median~\cite{GeoMedian} & \textcolor{red}{$\times$} & \textcolor{darkgreen}{$\checkmark$} & \textcolor{red}{$\times$} & \textcolor{darkgreen}{$\checkmark$} & \textcolor{red}{$\times$} & \textcolor{darkgreen}{$\checkmark$} & \textcolor{red}{$\times$} \\
        
        SecAgg~\cite{PracticalSMPC} & \textcolor{darkgreen}{$\checkmark$} & \textcolor{red}{$\times$} & \textcolor{red}{$\times$} & \textcolor{darkgreen}{$\checkmark$} & \textcolor{red}{$\times$} & \textcolor{red}{$\times$} & \textcolor{red}{$\times$} \\
        
        BREA~\cite{brea} & \textcolor{darkgreen}{$\checkmark$} & \textcolor{darkgreen}{$\checkmark$} & \textcolor{red}{$\times$} & \textcolor{red}{$\times$} & \textcolor{red}{$\times$} & \textcolor{darkgreen}{$\checkmark$} & \textcolor{red}{$\times$} \\
        
        RoFL~\cite{RoFL} & \textcolor{darkgreen}{$\checkmark$} & \textcolor{darkgreen}{$\checkmark$} & \textcolor{red}{$\times$} & \textcolor{red}{$\times$} & \textcolor{red}{$\times$} & \textcolor{red}{$\times$} & \textcolor{red}{$\times$} \\
        
        EIFFeL~\cite{EIFFeL} & \textcolor{darkgreen}{$\checkmark$} & \textcolor{darkgreen}{$\checkmark$} & \textcolor{darkgreen}{$\checkmark$} & \textcolor{darkgreen}{$\checkmark$} & \textcolor{darkgreen}{$\checkmark$} & \textcolor{red}{$\times$} & \textcolor{red}{$\times$} \\
        
        zkFL~\cite{wang2024zkfl} & \textcolor{darkgreen}{$\checkmark$} & \textcolor{darkgreen}{$\checkmark$} & \textcolor{red}{$\times$} & \textcolor{darkgreen}{$\checkmark$} & \textcolor{darkgreen}{$\checkmark$} & \textcolor{red}{$\times$} & \textcolor{red}{$\times$} \\
        
        ELSA~\cite{elsa} & \textcolor{darkgreen}{$\checkmark$} & \textcolor{darkgreen}{$\checkmark$} & \textcolor{darkgreen}{$\checkmark$} & \textcolor{darkgreen}{$\checkmark$} & \textcolor{darkgreen}{$\checkmark$} & \textcolor{red}{$\times$} & \textcolor{darkgreen}{$\checkmark$} \\
        
        ScionFL~\cite{scionfl} & \textcolor{darkgreen}{$\checkmark$} & \textcolor{darkgreen}{$\checkmark$} & \textcolor{darkgreen}{$\checkmark$} & \textcolor{darkgreen}{$\checkmark$} & \textcolor{red}{$\times$} & \textcolor{red}{$\times$} & \textcolor{red}{$\times$} \\
        
        zProbe~\cite{zprobe} & \textcolor{darkgreen}{$\checkmark$} & \textcolor{darkgreen}{$\checkmark$} & \textcolor{red}{$\times$} & \textcolor{darkgreen}{$\checkmark$} & \textcolor{red}{$\times$} & \textcolor{darkgreen}{$\checkmark$} & \textcolor{red}{$\times$} \\
        
        PPSFL~\cite{ppsfl} & \textcolor{darkgreen}{$\checkmark$} & \textcolor{darkgreen}{$\checkmark$} & \textcolor{darkgreen}{$\checkmark$} & \textcolor{darkgreen}{$\checkmark$} & \textcolor{darkgreen}{$\checkmark$} & \textcolor{red}{$\times$} & \textcolor{red}{$\times$} \\
        
        ZKFL~\cite{trusted_zkp} & \textcolor{darkgreen}{$\checkmark$} & \textcolor{darkgreen}{$\checkmark$} & \textcolor{darkgreen}{$\checkmark$} & \textcolor{darkgreen}{$\checkmark$} & \textcolor{darkgreen}{$\checkmark$} & \textcolor{darkgreen}{$\checkmark$} & \textcolor{darkgreen}{$\checkmark$} \\
        \bottomrule
    \end{tabular}
    }
\end{table}

%% file: survey_sections/frameworks.tex
\section{Frameworks \& Libraries}
\label{sec:frameworks}

Unlike classical Machine Learning or Deep Learning projects mainly using a learning framework like PyTorch or Jax, Federated Learning also requires a special learning infrastructure to deal with clients and server processes/communication. Therefore, in an attempt to ease the burden of every FL researcher creating a boilerplate for the FL infrastructure and also in a pursuit to create a distributed learning framework standard, several frameworks have been proposed. We explore the vast alternatives and expand upon a previous analysis that encompasses several FL frameworks~\citep{frameworks_comparison}.

\subsection{Federated Learning Frameworks}

\paragraph{General Federated Learning Frameworks:}
\begin{itemize}
    \item \href{https://github.com/tensorflow/federated}{Tensorflow Federated (TFF)}~\citep{tensorflow_federated}.  TFF is an open-source FL framework originated by the TensorFlow team. The framework interface is divided into two core layers: Federated Learning (FL) API -- high-level layer designed for applying pre-built FL algorithms and evaluation schemes; Federated Core (FC) API -- low-level layer to customise communication and federated algorithms (it is the basis of the first layer).
        \item \href{https://github.com/FederatedAI/FATE}{FATE}~\citep{fate}. Open-source FL framework designed for industry deployments and deeply rooted to secure computation protocols based on Homomorphic Encryption and Multi-Party Computation (MPC)
        \item \href{https://github.com/OpenMined/PySyft}{PySyft}~\citep{PySyft}. Open-source data science framework with some Federated Learning capabilities.
    \item \href{https://github.com/NVIDIA/NVFlare}{NVIDIA Flare (NVFlare)}~\citep{NVIDIA_FLARE}. NVIDIA Federated Learning Application Runtime Environment is an open-source FL Framework designed for private and secure real-world deployments, and it is flexible enough for research applications.
    \item \href{https://github.com/IBM/federated-learning-lib}{IBM FL}~\citep{ibm_fl}. IBM Federated Learning is a Python framework for Federated Learning that includes built-in fusion algorithms and allows for customisation of aggregation methods.
    \item \href{https://github.com/securefederatedai/openfl}{OpenFL}~\citep{openfl_citation}. Originally designed for the healthcare domain, OpenFL is an open-source Federated Learning framework concerned with providing a scalable and secure implementation for both research and production use. All federation collaborators agree on a Federated Plan (inspired by \citep{fl_plan_original}), which is parsed both at the client and server side to process and communicate the learning procedure.
    \item \href{https://github.com/NevronAI/metisfl}{MetisFL}~\citep{MetisFL}. MetisFL proposes a blatant fast open source Federated Learning framework, designed with asynchronous communication by default. MetisFL implemented the server component in C++ to provide low-level optimisations to the server operations, along with other key useful abstractions: model-store databases, and secure aggregators. They introduce this notion of a Federation Driver driven by a user/client configuration file, which is responsible for Secure Key generation, model/system initiation and monitoring.
    \item \href{https://github.com/PaddlePaddle/PaddleFL}{PaddleFL}~\citep{paddle}. An open-source Federated Learning framework based on PaddlePaddle. Supports both Horizontal and Vertical Learning strategies.
    \item \href{https://github.com/SMILELab-FL/FedLab}{FedLab}~\citep{fedLab}. FedLab is an open-source FL framework with several federated optimisation algorithms implemented and data partitioning techniques.
    \item \href{https://github.com/adap/flower}{Flower}~\citep{flower}. An open-source Federated Learning framework, very flexible by design providing easy-to-follow and in-depth documentation, with key characteristics (e.g. asynchronous communication protocols).
    \item \href{https://docs.substra.org/en/stable/index.html}{SubstraFL}~\citep{substra}. Exploring the deficiencies of other FL frameworks regarding robustness to real-world deployments, while guaranteeing easy and fast R\&D Research Substra and SubstraFL were created. Substra is an SDK library running on clients, whereas SubstraFL is a Python high-level framework based on Substra to run Federated Learning experiments at scale.
    \item \href{https://github.com/FedML-AI/FedML}{FedML Federate}~\citep{fedML}. FedML is a commercial ML framework to train, to collect data and to serve models. One of their solutions is the FedML Federate to train and serve federated models.
    \item \href{https://github.com/scaleoutsystems/fedn}{FEDn}~\citep{fedN}. FedN is an efficient and lightweight open-source Federated Learning framework providing a Hierarchical FL solution. It is deeply inspired by the MapReduce programming model since it hierarchically combines and reduces distributed computation.
    
\end{itemize}

\paragraph{Tree-based Federated Learning Frameworks:}
\begin{itemize}
    \item \href{https://github.com/dmlc/xgboost/blob/master/python-package/xgboost/federated.py}{Federated XGBoost}~\citep{fed_xgboost}. This framework is a Federated Learning extension to the XGBoost-optimised distributed gradient boosting library.
    \item \href{https://github.com/dmlc/xgboost/blob/master/python-package/xgboost/federated.py}{FedTree}~\citep{fedtree}. It is a framework to train gradient boosting decision tree models, having in consideration privacy/security and data heterogeneity concerns.
\end{itemize}

\paragraph{Domain focused frameworks:}
\begin{itemize}
    \item \href{https://fedbiomed.org/latest/getting-started/fedbiomed-workflow/}{Fed-BioMed}~\citep{fed_bio_med}. A framework designed to be used for real-world medical applications including Federated Learning/analytics.
\end{itemize}

Apart from the Tensorflow Federated project, Google has also proposed FedJax~\citep{FedJAXFL}, a Federated Learning simulations framework specially designed for JAX. More recently, Google proposed a federated auto differentiation framework~\citep{FederatedAutoDiff} and published a library (FAX:~\citet{fax}) based on the JAX deep learning framework to utilise it.

Apple has developed pfl~\citep{pfl2024}, a Python framework for Private Federated Learning simulations focusing on privacy features, flexible ML model support and straightforward benchmark testing.

P2PFL~\citep{p2pfl} is an open-source Python library designed to develop applications in real and simulated environments. It focuses on peer-to-peer networks, utilizing decentralised and gossip protocols.

ChainFL~\citep{chainFL} is a general simulation framework written in Python for training Federated Learning models on distributed edge environments. It provides security guarantees similar to blockchain algorithms, ensuring updated parameters' integrity by preventing modification.

\subsection{Frameworks in-depth comparison}

Inspired by the framework comparison approach introduced by MetisFL~\citep{MetisFL}, we present systematic tables that summarise the key differences among general-purpose Federated Learning frameworks.

\begin{itemize}
    \item \textbf{Table~\ref{tab:frameworks_comparison_data_privacy}} provides an overview of the privacy and security protocols implemented in each framework.

    \begin{table}[htbp]
        \centering
        \caption{Comparison between FL frameworks in regards to their respective data, privacy \& security implementations.}\bigskip
        \input{tables/frameworks_features_1}
        \label{tab:frameworks_comparison_data_privacy}
    \end{table}

    \item \textbf{Table~\ref{tab:frameworks_comparison_communication}} highlights the communication strategies and implementation choices.

    \begin{table}[htbp]
        \centering
        \caption{Comparison between FL frameworks in regards to their corresponding federated communication utilities.}\bigskip
        \input{tables/frameworks_features_2}
        \label{tab:frameworks_comparison_communication}
    \end{table}
    
    \item \textbf{Table~\ref{tab:frameworks_comparison}} offers a broader comparative analysis, capturing the general distinctions between the frameworks.

    \begin{table}[htbp]
        \centering
        \caption{Meta-information comparison between FL frameworks. (*) indicates low quality, while (***) represents high quality.}\bigskip
        \input{tables/frameworks_meta_info}
        \label{tab:frameworks_comparison}
    \end{table}

\end{itemize}

These structured comparisons aim to provide a clear and comprehensive understanding of the current state of Federated Learning frameworks.

\subsection{On-device Learning Frameworks}
\label{sub_sec:on_device_learning}

Concerned with the hardware limitations of edge devices, Tiny Training Engine~\citep{lin2020mcunet}, TinyTL~\citep{TinyTL} and TinyTrain~\citep{TinyTrainDN} frameworks have been proposed to depart from the continuous communication dependency between devices in a Federated Learning Schema. These frameworks focus on leveraging techniques to make the on-device learning process more efficient while using fewer resources (e.g. Transfer Learning, Layers Pruning, Meta-Learning, and Sparse Learning techniques). Moreover, \href{https://github.com/tinygrad/tinygrad}{Tinygrad}~\citep{tinygrad} is a promising low-resource and open-source deep learning framework with a strong focus on being compatible with on-device chip accelerators. Since these frameworks do not consider FL algorithms design, we do not explore them in detail. However, we believe that the combination of on-device learning methodologies and FL is a step forward to more robust real-world ML Edge applications.

%% file: tables/frameworks_features_1.tex
\scalebox{0.85}{
\begin{tabular}{lcccccc}
\toprule
\textbf{Framework} & \textbf{Purpose} & \textbf{\makecell{ML\\Backend}} & \textbf{\makecell{Horizontal\\FL}} & \textbf{\makecell{Vertical\\FL}}  & \textbf{\makecell{Privacy \& Secure\\Mechanisms}} & \textbf{\makecell{Crypto\\Library}} \\ \midrule
NVFlare   &  FL &   Torch | TF | MONAI &  \textcolor{darkgreen}{$\checkmark$} &  \textcolor{red}{$\times$} &  FHE & TenSeal           \\
FATE   &  FL &   Torch &  \textcolor{darkgreen}{$\checkmark$} &  \textcolor{darkgreen}{$\checkmark$} &  MPC | PSI | FHE | SE &  \makecell{curve25519\_dalek\\| x25519\_dalek \\| native}           \\
PySyft   &  Data Science | FL &   Torch &  \textcolor{darkgreen}{$\checkmark$} &  \textcolor{red}{$\times$} &  FHE & TenSeal           \\
IBM FL  &  FL &   Torch | TF &  \textcolor{darkgreen}{$\checkmark$} &  \textcolor{red}{$\times$} &  FHE & HElayers \\
OpenFL  &  FL &   Torch | TF &  \textcolor{darkgreen}{$\checkmark$} &  \textcolor{red}{$\times$} &  TEE & Graphene\\
TFF  &  FL | Analytics &   TF &  \textcolor{darkgreen}{$\checkmark$} &  \textcolor{red}{$\times$} &  DP | SE | etc & native (TFP) \\
MetisFL  &  FL &   Torch | TF &  \textcolor{darkgreen}{$\checkmark$} &  \textcolor{red}{$\times$} &  FHE & PALISADE \\
FedLab     &  FL &   Torch &  \textcolor{darkgreen}{$\checkmark$} &  \textcolor{red}{$\times$} &  - & - \\
PaddleFL    &  FL &   Torch  &  \textcolor{darkgreen}{$\checkmark$} &  \textcolor{darkgreen}{$\checkmark$} &  FHE & TenSeal \\
Flower    &  FL | Analytics &   agnostic &  \textcolor{darkgreen}{$\checkmark$} &  \textcolor{red}{$\times$} & DP | Masking | FHE & Native \\
SubstraFL    &  FL | Analytics &   agnostic &  \textcolor{darkgreen}{$\checkmark$} &  \textcolor{red}{$\times$} &  - & - \\
FedML Federate  &  FL &   Torch | TF | MX | JAX &  \textcolor{darkgreen}{$\checkmark$} &  \textcolor{red}{$\times$} &  Masking | FHE & Native \\
FEDn  &  FL &   agnostic &  \textcolor{darkgreen}{$\checkmark$} &  \textcolor{red}{$\times$} &  - & - \\
\bottomrule
\end{tabular}
}

%% file: tables/frameworks_features_2.tex
\scalebox{0.95}{
\begin{tabular}{lcccccc}
\toprule
\textbf{Framework} & \textbf{Centralised} & \textbf{Decentralised}  & \textbf{Hierarchical} & \textbf{Protocol} & \textbf{Synchronous} &\textbf{Asynchronous} \\ \midrule
NVFlare   &  \textcolor{darkgreen}{$\checkmark$} &  \textcolor{red}{$\times$} &  \textcolor{red}{$\times$} & gRPC &  \textcolor{darkgreen}{$\checkmark$} &  \textcolor{red}{$\times$}            \\
FATE   & \textcolor{darkgreen}{$\checkmark$} & \textcolor{red}{$\times$} & \textcolor{red}{$\times$} & gRPC/MQ & \textcolor{darkgreen}{$\checkmark$} & \textcolor{red}{$\times$}            \\
PySyft   & \textcolor{darkgreen}{$\checkmark$} & \textcolor{red}{$\times$} & \textcolor{red}{$\times$} & REST & \textcolor{darkgreen}{$\checkmark$} & \textcolor{red}{$\times$}           \\
IBM FL  &  \textcolor{darkgreen}{$\checkmark$} &  \textcolor{red}{$\times$} &  \textcolor{red}{$\times$} & AMQP &  \textcolor{darkgreen}{$\checkmark$}  &  \textcolor{red}{$\times$}         \\
OpenFL  &  \textcolor{darkgreen}{$\checkmark$} &  \textcolor{red}{$\times$} &  \textcolor{red}{$\times$} & gRPC &  \textcolor{darkgreen}{$\checkmark$} &  \textcolor{red}{$\times$}           \\
TFF  & \textcolor{darkgreen}{$\checkmark$} & \textcolor{red}{$\times$} & \textcolor{red}{$\times$} & gRPC & \textcolor{darkgreen}{$\checkmark$} & \textcolor{red}{$\times$}           \\
MetisFL     &  \textcolor{darkgreen}{$\checkmark$} &  \textcolor{red}{$\times$} &  \textcolor{red}{$\times$} & gRPC &  \textcolor{darkgreen}{$\checkmark$} &  \textcolor{darkgreen}{$\checkmark$}            \\
FedLab     & \textcolor{darkgreen}{$\checkmark$} & \textcolor{red}{$\times$} & \textcolor{darkgreen}{$\checkmark$} & Gloo 
 & \textcolor{darkgreen}{$\checkmark$} & \textcolor{darkgreen}{$\checkmark$}        \\
PaddleFL     & \textcolor{darkgreen}{$\checkmark$} & \textcolor{darkgreen}{$\checkmark$} & \textcolor{darkgreen}{$\checkmark$} & gRPC & \textcolor{darkgreen}{$\checkmark$} & \textcolor{red}{$\times$}            \\
Flower     & \textcolor{darkgreen}{$\checkmark$} & \textcolor{red}{$\times$} & \textcolor{red}{$\times$} & gRPC & \textcolor{darkgreen}{$\checkmark$} & \textcolor{red}{$\times$}          \\
SubstraFL     & \textcolor{darkgreen}{$\checkmark$} & \textcolor{red}{$\times$} & \textcolor{red}{$\times$} & REST & \textcolor{darkgreen}{$\checkmark$} & \textcolor{darkgreen}{$\checkmark$}          \\
FedML Federate     &  \textcolor{darkgreen}{$\checkmark$} &  \textcolor{darkgreen}{$\checkmark$} &  \textcolor{red}{$\times$} & MPI &  \textcolor{darkgreen}{$\checkmark$} &  \textcolor{red}{$\times$}            \\
FEDn     &  \textcolor{darkgreen}{$\checkmark$} &  \textcolor{red}{$\times$} &  \textcolor{darkgreen}{$\checkmark$} & gRPC \& REST &  \textcolor{darkgreen}{$\checkmark$} &  \textcolor{darkgreen}{$\checkmark$}           \\
\bottomrule
\end{tabular}
}

%% file: tables/frameworks_meta_info.tex
\scalebox{0.9}{
\begin{tabular}{lcccc}
\toprule
\textbf{Framework} & \textbf{\makecell{Deprecated / \\ Unmaintained}}  & \textbf{Host} & \textbf{\makecell{Documentation \\ Quality}} & \textbf{\makecell{User \\ Friendly}} \\ \midrule
NVFlare   & \textcolor{red}{$\times$} & NVidia & *** & **            \\
FATE    & \textcolor{red}{$\times$} & Linux Foundation & ** & **            \\
PySyft   & \textcolor{red}{$\times$} & OpenMined & ** & **           \\
IBM FL   & \textcolor{darkgreen}{$\checkmark$} & IBM & ** & *          \\
OpenFL   & \textcolor{red}{$\times$} & Linux Foundation & ** & *           \\
TFF   & \textcolor{red}{$\times$} & Tensorflow & *** & ***           \\
MetisFL   & \textcolor{red}{$\times$} & MetisFL & ** & ***            \\
FedLab    & \textcolor{red}{$\times$} & SMILELab-FL
 & ** & **        \\
PaddleFL   & \textcolor{darkgreen}{$\checkmark$} & PaddlePaddle & * & *            \\
Flower     & \textcolor{red}{$\times$} & Flower & *** & ***           \\
SubstraFL   & \textcolor{red}{$\times$} & Linux Foundation & *** & ***          \\
FedML Federate   & \textcolor{red}{$\times$} & FedML & ** & **            \\
FEDn      & \textcolor{red}{$\times$} & Scaleout Systems & ** & **            \\
\bottomrule
\end{tabular}
}

%% file: survey_sections/alternative_paths.tex
\section{Alternative Paths in Distributed Learning}
\label{sec:alternatives}

\subsection{Distributed Deep Neural Networks}
\label{sub_sec:ddnn}

At present, large models with billions are extensively being trained, either in the Natural Language Processing or in the image/video synthesis domain. The training pipeline duration of those large architectures can last a considerable number of hours/days. Thus, the necessity to accelerate the training procedure is real. One way to accelerate the training process is to dismount the complete model architecture into sub-parts within several distributed nodes. Such a distributed learning mechanism is called \textbf{Model Parallelism} of Distributed Deep Neural Networks (DDNN:~\citet{megatron}). This methodology proves to be complex since it requires a decision on how and where to partition the model into several network nodes~\citep{p3_graph_nn}.

In an orthogonal direction, \textbf{Data Parallelism}, consists of learning the same model through multiple distributed data silos. From a holistic perspective, homogeneous model training within Federated Learning can be considered a take on \textbf{Data Parallelism} within a distributed learning setting.

There is also the \textbf{Pipelining parallelism} technique, which 
divides training tasks for DNN models into sequential processing stages, through the assumption that models are sequentially composed and can be relaxed (e.g. using skip connections). Consider Torchgpipe work~\citep{torchgpipeOP}, a torch implementation of GPipe~\citep{gpipe}, a framework for computing a model training distributively by pipelining model layers computations in a parallel arrangement.

Finally, some works utilise \textbf{Hybrid Parallelisation} techniques~\citep{pipe_dream, dapple}, which combine Data and Model Parallelisation methodologies to mitigate communication overhead/increase training speed convergence. 

Orthogonal to these methodologies, there are also Distributed Tensor Computation techniques, which partition tensor operations into distributed machines to speed up computation or permit bigger model size training~\citep{distal}.

\begin{notebox}
We invite the reader to further explore this theme through the survey~\citep{distributed_nn_survey}.
\end{notebox}

\subsection{Modular Deep Learning}
\label{sub_sec:model_fusion}

Modular Deep Learning~\citep{ModularDL} is gaining popularity, as a solution to build an autonomous multi-task model combining many expert models on specific tasks, whilst mitigating the negative interference of the models merging. In this line of Modular Deep Learning, DiPaCo~\citep{DiPaCoDP} has proposed a sparsely-activated modular learning network where data and computation are distributed between clients. This results in a distributed learning network which on inference selects one of many expert paths of nodes to make a prediction, thus, using only a sparse set of clients.

\subsubsection{Model Merging}
\label{sub_sub_sec:model_fusion}

In the Large Language Model (LLM) literature, the notion of model fusion where different task-based models are merged to procreate a more general model with better out-of-distribution performance is gaining popularity. It entails an orthogonal direction to Federated Learning since no distributed learning is performed, however, the action of model fusion is intrinsically customary to the FL framework. Consequently, this raises the question of whether there are some methodologies from this new model fusion trend which could inspire some better model alignment in FL. Since there is an analogous relationship between Federated Learning model alignment and the new approaches to LLM model merging, we briefly discuss the main model fusion techniques: Model Soup~\citep{model_soups}, Task Arithmetic~\citep{task_arithmetic}, TIES-Merging~\citep{ties}, DARE~\citep{dare}.

\paragraph{\textbf{Model Soup.}} This technique uses a weighted average of trained models, without retraining them. In spirit, this merging technique is similar to a weighted FedAvg, not only due to the weighted model aggregation but also due to the data used for training each model not being the same. In Model Soup, model training further distinguishes itself by the usage of different training hyper-parameters.

\paragraph{\textbf{Task Arithmetic.}} Focused on the notion of task vectors (delta between the weights of a task fine-tuned model and its corresponding pre-trained model), Task Arithmetic is a method, which applies vector operations over a set of task-specific fine-tuned models (adding/negating/relating vectors) to have an end-model more general.

\paragraph{\textbf{Ties-Merging.}} Confronted with the redundancy of parameters and conflicting element-wise sign values in model merging techniques, Ties-Merging proposes an extension to Task Arithmetic work. Ties-Merging trims useless parameter values, resolves conflicting signs and aligns parameters which are in agreement.

\paragraph{\textbf{DARE.}} Delta parameters (i.e. the difference between fine-tuned and pre-trained parameters) do not all have the same beneficial impact on the fine-tuned task. DARE took this idea and systematically dropped delta parameters which do not influence performance when merging models, whilst increasing the capabilities of the final model.

Considering the panoramic view of these works, one could use these methodologies (e.g. Task Arithmetic based), in a peer-to-peer network with homogeneous models to better guide the learning from each local client model.

\subsection{Federated Statistics}
\label{sub_sec:f_statistics}

We have showcased that distributed systems can be designed to support learning mechanisms, as discussed throughout this survey. However, general data science tasks can also be performed in a federated setting (e.g., statistical analysis such as binning data into intervals or building histograms) without compromising user data. This adjacent field to Federated Learning is called Federated Statistics. Differential Privacy and Secure Aggregation mechanisms we explored in Section.~\ref{sec:privacy} and Section.~\ref{sec:security} represent the foundational mechanism for computing statistics of the federation population, whilst preserving some level of anonymity. \citet{federated_stats_privacy} extend this approach by incorporating an on-device Differential Privacy mechanism, allowing the end-user/client to control the privacy level.

\subsection{Distributed Consensus}
\label{sub_sub_sec:consensus}

In Federated Learning, the most common objective is to obtain a global consensus on a global model which best fits all the clients. This implies that some sort of network topology is in place ensuring that they can communicate and that it is robust to network perturbations. In the distributed systems field, this inherent consensus goal of FL dates back to the 1980s through the work of \citet{Lamport1998ThePP}. Lamport presented the Paxos consensus algorithm, focusing mainly on single-decree Paxos.

Raft~\citep{raft} was presented in 2014, attempting to create a simpler consensus protocol and combat deficiencies of the Paxos method. Raft defines the existence of a leader in a network as one who is elected by other nodes in a network. This leader changes occasionally, and the network is designed to be fault-tolerant, and a new leader is elected when needed. The objective of the leader is to ensure a consensus is guaranteed in good terms. The notary properties of Raft (e.g. a fault-tolerant algorithm) make it useful for FL decentralised network topologies, as explored by \citet{raft_fl}.

%% file: survey_sections/discussion.tex
\section{Federated Learning in Practice and Deployment}
\label{sec:discussion}

\subsection{Contemporary use-cases}
\label{sub_sec:usecases}

Conventional Machine Learning applications on business projects stand heavy on exploring the data and conducting analysis over it. Since the data is localised and not accessible, monitoring and debugging stand as a challenging aspect of building federated solutions. By maintaining data processing, model training, and model inference locally on devices, sensitive user data is not transmitted across borders or regions, thereby avoiding major compliance challenges for organizations.
Nevertheless, the useful properties of FL have led to developments and research in several fields:

\subsubsection{Healthcare}
\label{sub_sub_sec:helthcare}

In the health domain, patient privacy is taken very seriously, which hinders the availability and access to data. The data itself is also very expensive to acquire/annotate: BrainTorrent~\citep{descentralized_BrainTorrent} notes that a 3D brain MRI scan manual annotation can take up to a week by a trained professional. In this work, they take a peer-to-peer approach to Federated Learning. In contrast, \citet{health_fl} adheres to a distributed star-schema topology, exemplifying a collaborative effort among multiple hospitals aimed at enhancing the accuracy and efficiency of predicting patients' histological response to Neoadjuvant Chemotherapy (NACT).

On Drug Discovery research, the \href{https://www.melloddy.eu}{MELLODY} project is an attempt to enhance predictive Machine Learning models on decentralised data over many partner companies, without exposing proprietary information.

The \href{https://www.fchampalimaud.org/news/champalimaud-foundation-scientists-create-health-data-cloud-infrastructure-research-first-time}{MetaBreast} project in an orthogonal direction has paved the way for creating a public cloud data health centre with anonymous data.

FL-based smart healthcare applications can also be found under the domain of the Medical Internet of Things (MIoT), which normally follows a more complex network infrastructure~\citep{miot_fl}. Typical solutions, normally invoke the combination of IoT devices (e.g. smart inhaler, smart-watch) with more resilient edge devices (e.g. smartphone). Contrasting with standard FL applications, the data in a local network can move from an IoT device to a more powerful edge device, to perform training/inference with better reliability.

\paragraph{\textbf{Accessibility.}}
\label{par:accessability}

IoT/Embedded systems advancements allow for potential breakthroughs in designing systems to help people with accessibility issues. Most importantly, FL helps to bring personalised solutions to these people. Machine Learning Systems with TinyML book~\citep{ml_systems_tinyml} suggests some interesting applications: \textbf{a.} Responsive prosthetic limbs, which through the analysis of nerve signals, and muscle tension can move and adjust the grip of the prosthetics and exoskeletons dynamically;
\textbf{b.} Voice-enabled devices can be designed to translate non-verbal inputs to personalised vocal outputs;
\textbf{c.} The creation of interfaces to convert gestures into natural speech tailored for individual users and the opposite.
\textbf{d.} Development of smart glasses to convert visual information into audio input.
\textbf{e.} Hearing aids to help intelligently amplify different sources of audio.

\subsubsection{Factories (Industry 4.0)}

Industry 4.0 represents a significant advancement in the manufacturing industry, where the seamless interconnectivity of machines serves as a cornerstone for enhanced monitoring, performance optimization, and accountability. In this envisioned landscape, wireless networks emerge as indispensable due to their unparalleled flexibility and portability, offering distinct advantages over traditional cable or optical fibre solutions.

Digital Twins are the basis of this new revolution in which physical machines have a corresponding digital representation, through the usage of sensors and other IoT devices deeply connected to the original devices. The presence of such technology
in factories, for instance, permits machine equipment to be automatically diagnosed, either for faulty clues or for production quality inspection with the help of Federated Learning methodologies.

FedGS~\citep{industry_fl} proposes a clustered Federated Learning solution for factory environments where OCR is used to identify factory artefacts (e.g., packing boxes) through the reading of their badges. Faced with the fact data distributions vary among factories, they propose a constrained gradient-based optimiser (Gradient-Based Binary Permutation Client Selection: GBP-CS) with group/cluster synchronisation.

\subsubsection{Energy \& Agriculture}

Smart Homes are gaining popularity through the increasing integration of small IoT devices inside peoples' houses. Their utility explains their search in helping to register energy consumption, optimise energy spending, and record raw data (e.g. CCTV Cameras):~\citet{FederatedRL}. Water supply can also be optimised through personalised analysis of individual houses, resulting in smaller bills and wasted resources~\citep{FederatedLearning_water}.
The distributed nature of energy and water resources enhances the collaboration dynamism and reliability of estimations when using Federated Learning techniques.

Drones equipped with sensors and cameras are helping to gather data previously very hard and expensive to obtain. Further, a coordinated mesh of drones can capture with much more fidelity ground terrains, leading to better quality data which can increase models' predictive performance~\citep{lidar_dataset}.

Digital twins can also be used in renewable energy-producing devices along with Federated Learning to better maintain them and optimise their efficiency.

To better preserve limited resources (e.g. water), there is interest in turning agriculture more efficient, sustainable and more robust to natural disasters.
Smart Agriculture is a paradigm that utilises low-energy devices (e.g. sensors) to help better monitor agriculture production. 

On this domain, PEFL~\citep{smart_agriculture}, concerned with malicious attacks, proposes a combination of a two-level privacy module of perturbation-based encoding with an intrusion detection trained using the FL framework and a Gated Recurrent Unit (GRU) model to flag out corrupted devices.

For diseases/pests control of crops, \citet{pests_fl} have proposed a Federated Learning framework using R-CNN architectures.

\subsubsection{Space}
\label{sub_sub_sec:space}

The space domain faces many challenges, from logistical cooperation between organisations to practical operations such as rocket launching, and lunar vehicle mobilisation. FRIENDS~\citep{friends} proposes an inter-connected network of Lunar nodes for a more robust lunar exploration. A federated deployment of MoonNet was used and it resorted to pseudo-labelled training updates.

Furthermore, there is interest in the collaboration of different public and private entities, provided that space exploration is a global effort. Initiatives, such as SpaceDAO~\citep{SPACEDAO}, are developing a Blockchain-Based Consensus Mechanism to support, among other use cases, a Federated Learning ecosystem to ensure safer collaborative space traffic.

\subsubsection{Others}
\label{sub_sub_sec:others}

Google Keyboard (Gboard) is historically an early adopter of Federated Learning~\citep{keyboard_prediction,gboard}, using it for three main tasks: \textbf{1.} Next Word Prediction (NWP) to predict the next word to be typed; \textbf{2.} Smart Compose (SC) to suggest probable inline characters to be typed; \textbf{3.} On-The-Fly Re-scoring (OTF) to re-rank next-word candidates.

Recently, ``Generative AI'' gained considerable popularity by introducing high-quality Large Language Models in mundane applications, along with state-of-the-art Diffusion Models in the image generation realm. Adapting these models to federated training/fine-tuning is a promising future direction. FedTabDiff~\citep{fedTabDiff}, in the domain of table data synthetic generation, proposes a federation learning approach to Denoising Diffusion Models.

Mixing genetic evolutionary algorithms with clustering techniques has also been proposed to improve training hyperparameter tuning~\citep{geneticCFL}.

In a different line of work, there is research~\citep{fl_maritinme_airComp} fusing Federated Learning with over-the-air computation in the maritime domain to reduce the communication overhead, by applying computation over communication frequencies.

\subsection{Federated Learning as a Service (FLaaS)}
\label{sub_sec:marketplace}

Machine Learning as a Service (MLaaS) has exploded in popularity in recent years since it provides easy access to public data, novel models and infrastructure provisioning. To expand the cloud provisioning of Machine Learning solutions we argue that there is a considerable market for Federated Learning as a Service (FLaaS). 

FLaaS may provide collaborative heterogeneous learning solutions, where entities interested in a similar learning task may interchange a proxy model learnt on local data to train a global model accessible to the participating parties, without revealing either raw local data or model architectures.

In the literature, CrowdFL~\citep{fl_marketplace} implements a simple collaborative learning infrastructure, focusing on mobile devices, but without exploring the essence of the heterogeneous virtue of the data and even models underlying federation learning applicability. FLaaS~\citep{FLaaSFL} explores a robust take on Federated Learning as a Service, whilst building an infrastructure with specific APIs for data, model, protection and training resources. 

We envision a key role of homogeneous and heterogeneous model learning and, therefore provide two clear paths for an FLaaS Marketplace solution: 
\begin{itemize}
    \item  Within a network each independent entity agrees on a shared model architecture protocol to solve a specific task (the local and shared models are the same);
    \item Similar to Proxy Model Sharing work~\citep{ProxyDecentralizedFL}, following a more relaxed cooperation mechanism, each entity in the federation solely agrees on a task. This vision assumes heterogeneous modelling between the entities participating in the federation. Thus, the communicated/shared model acts as a communication protocol, where the data transmitted is a local model representing the heterogeneous compressed knowledge.
\end{itemize}

\subsection{Federated Learning Operations (FLOps)}
\label{sub_sec:flops}

\begin{figure}[!htbp]
    \centering
    \includegraphics[width=\linewidth]{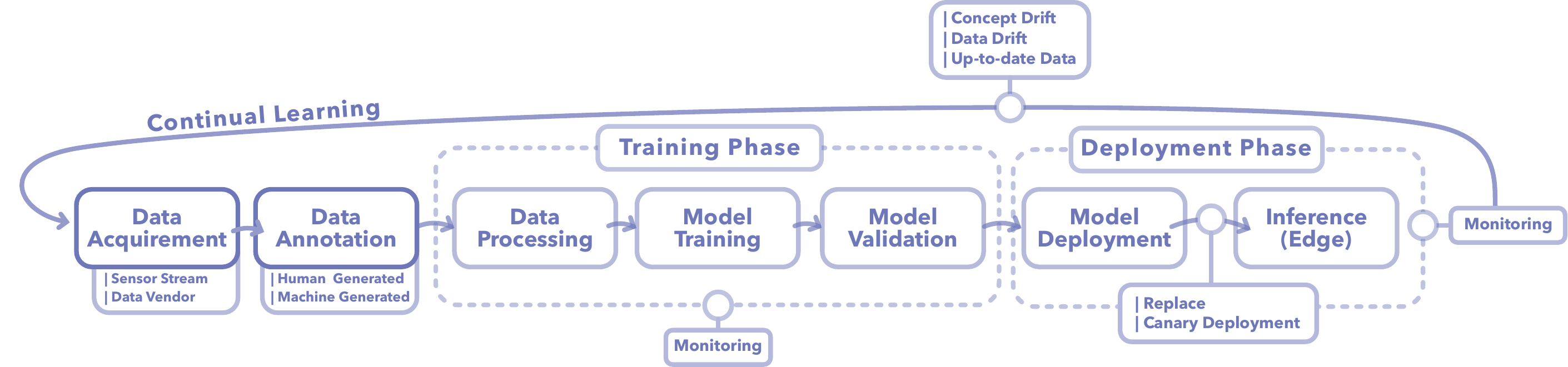}
    \caption{Standard Machine Learning Application life-cycle. Deployment Machine Learning pipelines, normally, have 3 main phases: \textbf{1.} Data Management; \textbf{2.} Model Training; \textbf{3.} Model Deployment. To make the pipeline robust, the introduction of a cycle comes naturally with the help of monitoring mechanisms. Whereas during training, monitoring helps debug processes and reproduce experiments, in the deployment phase it helps to audit shifts in performance. This notion of a cycle represents the system's ongoing evolution in its quest for enhanced performance within an ever-changing environment.}
    \label{fig:mlops}
\end{figure}

Designing a Machine Learning pipeline involves a thoughtful life-cycle development. This practice of orchestrating well-grounded workflows and robust Machine Learning systems is called Machine Learning Operations (MLOps). Normally, it consists of a hierarchical directed graph with three main phases: \textbf{1.} Data Handling Phase; \textbf{2.} Training Phase; \textbf{3.} Deployment Phase (See Fig.~\ref{fig:mlops} for a holistic view of the standard approach to a Machine Learning pipeline).

Federated Learning Operations can be regarded as an extension of the well-studied domain of Machine Learning Operations, where the mechanics of a Machine Learning life-cycle are put into practice in a distributed setting.

A combination of Federated Learning Frameworks (e.g. Flower) focusing on easing the distributed learning infrastructure; Machine Learning Monitoring frameworks (e.g. WandB:~\citet{wandb}); And On-Device Learning frameworks (e.g.~\href{https://edgeimpulse.com/}{Edge Impulse}:~\citet{edge_impulse}) designed for monitoring/deploying inference real-time products provide a solid FLOps architecture.

\subsection{Environment Impact}
\begin{figure}[!htbp]
    \centering
    \includegraphics[clip,width=0.7\linewidth]{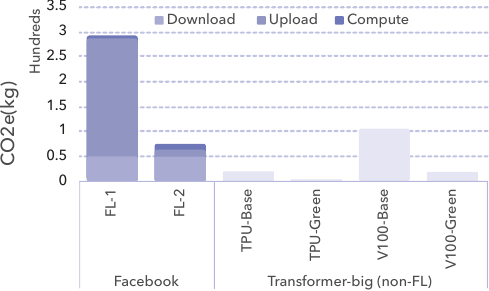}
    \caption{According to \citet{SustainableAI}, Federated Learning can have considerable CO2 emissions (carbon footprint), when compared with standard centralised Large Models training.}
    \label{fig:co2_fl}
\end{figure}

Federated Learning reduces the bandwidth of sending raw data across devices and the cloud, however the computation and communication overhead of aligning the several intertwined devices' models computed by heterogeneous devices is non-negligible. As stated by \citet{SustainableAI}, FL pipelines can be malign to the environment because most devices are on the edge computing on low energy efficient devices, consuming fossil-based energy. Cloud premises, differently, have further control since green energies can power them and use special hardware for more optimal resource spending (See Fig.~\ref{fig:co2_fl}, an adaptation figure from ~\citet{SustainableAI}).

Due to the limited resources found in Edge Devices, each training round is expected to take considerably more time than training it on a cloud server, considering an identical task.

Green Federated Learning~\citep{green_fl}, aware of the inefficiency attribute of Federated Learning has suggested the consideration of Carbon Emissions as an optimisation variable. They establish a strong correlation between the training time and the cohort size of a federation as strong factors for large-scale carbon emissions. Thus, carbon footprint optimisation relies on the client side (training) and communication side (upload/download weights/statistics). From their analysis, with the right parameterisation, AsyncFL converges faster reducing training time, however, it pollutes considerably more than the respective synchronous version. 

Model compression and quantisation can also help reduce the environmental impact while increasing the efficiency of the training/communication. For instance, implementing the widely-used int8 quantization technique in FL, as developed by \citet{int8_fl}, and estimated by \citet{green_fl}, can potentially decrease the carbon footprint by a factor of 1.8.

The rise of interest in measuring ML carbon footprint has given birth to the CodeCarbon~\citep{codecarbon} framework, which provides extensive utilities to measure an estimation of CPU, GPU and RAM computations carbon footprint.

%% file: survey_sections/next_steps.tex
\section{Future Directions}
\label{sec:next_steps}

Federated Learning (FL) continues to evolve, driven by its ability to facilitate distributed learning while addressing privacy, security, and data sovereignty concerns. However, despite its growing adoption, several challenges persist, particularly in terms of privacy guarantees, communication efficiency, resource heterogeneity, and the lack of standardised FL frameworks. To advance the field, we outline key research directions that will be crucial for future developments.

\subsection{Enhancing Trustworthiness}  
Ensuring trust in FL is critical for its adoption in high-stakes domains such as healthcare, finance, and autonomous systems. A major research priority is improving \textbf{uncertainty estimation} techniques, which allow models to quantify confidence in their predictions and detect unreliable outputs. Additionally, more robust learning mechanisms must be developed to handle adversarial threats, including data poisoning and model inversion attacks. Privacy-preserving \textbf{model auditing} frameworks are also essential to ensure that FL systems comply with ethical AI principles and regulatory requirements, fostering broader adoption in privacy-sensitive applications. As we have explored, \textbf{informed learning} in Federated Learning plays a crucial role in enabling more controlled and robust distributed models. This approach allows for the integration of domain-specific rules, expert knowledge, mathematical constraints, and physical dynamics, enhancing both the reliability and interpretability of FL systems.

\subsection{Optimising Communication and Resource Management}  
FL operates in highly heterogeneous environments where participating devices differ in computational capacity, energy constraints, and network connectivity. Future work should focus on \textbf{adaptive parameter optimisation} methods that dynamically balance local computation efficiency with global convergence speed. Additionally, \textbf{communication-aware aggregation strategies} must be developed to reduce bandwidth consumption without degrading model performance. Research into \textbf{efficient offloading techniques} that intelligently distribute computation across heterogeneous hardware, including edge devices, microcontrollers, and AI accelerators, will be key to improving FL’s scalability.

\subsection{Standardisation and Open Federated Learning Platforms}  
The absence of standardised frameworks for FL hinders its widespread adoption. A promising direction is the development of \textbf{federated open platforms} where multiple organisations can collaborate securely on decentralised learning tasks. This requires advances in \textbf{secure multi-party computation}, ensuring that entities with different trust levels can jointly train models without compromising data privacy. Establishing \textbf{interoperability standards} for FL systems, along with curated benchmark datasets and evaluation metrics, would further facilitate FL research and deployment across different domains.

\subsection{Resource-Constrained Learning}  
Deploying FL on low-power devices, such as IoT sensors and embedded systems, remains challenging due to their limited computational resources. Future research should explore \textbf{energy-efficient training algorithms} that optimise power consumption while maintaining learning effectiveness. Hardware-aware FL optimisations, leveraging AI chip accelerators, will also be necessary to enhance model inference efficiency. Moreover, advances in \textbf{gradient compression techniques} could significantly reduce communication overhead, making FL more viable in bandwidth-constrained environments.

\subsection{Model Compression and Quantisation Strategies}  
Communication costs continue to be a major bottleneck in FL. Future research should investigate \textbf{model compression techniques} such as pruning, distillation, and tensor decomposition to reduce the complexity of FL models while preserving performance. \textbf{Low-precision quantisation} methods can further decrease transmission costs by encoding model updates more efficiently. Additionally, \textbf{sparse learning approaches} and federated dropout strategies could improve communication efficiency while maintaining model robustness.

\subsection{Interdisciplinary Synergies and Emerging Applications}  
FL can benefit from advances in complementary research fields. Integrating \textbf{neurosymbolic AI} could improve model interpretability by combining FL with structured reasoning. Similarly, \textbf{self-supervised and continual learning} could enhance FL’s ability to adapt to dynamic, non-stationary data distributions. Another promising direction is \textbf{federated reinforcement learning}, which has the potential to optimise decentralised decision-making in applications such as smart grids, autonomous vehicles, and industrial automation.

%% file: survey_sections/conclusion.tex
\section{Conclusion}
\label{sec:conclusion}
Federated Learning (FL) is transforming machine learning in critical domains such as healthcare, finance, and edge computing by enabling distributed learning while offering the potential for enhanced privacy, security, and trustworthiness. However, as FL adoption grows, it is crucial to develop systematic methodologies that enhance its design, adaptability, and real-world applicability.

In this survey, we have proposed a Meta-Framework perspective that organises FL as a modular and composable system, structuring its key dimensions into interoperable components. This approach provides a unifying blueprint for systematically designing and implementing FL frameworks, offering both a theoretical foundation and a practical guide for researchers and practitioners.

Additionally, we introduce a novel taxonomy for Alignment, positioning it as a fundamental counterpart to Aggregation in FL. By recognising that aggregation not only combines parameters but also constrains them to ensure knowledge alignment, this perspective deepens our understanding of FL’s learning dynamics and its effectiveness in heterogeneous environments.

To further contextualise FL, we have traced its historical evolution, linking it to distributed optimisation and demonstrating how its foundations have shaped modern advancements. Recognising the need for accessibility, we have also provided a structured introduction to FL, lowering the barrier to entry for new researchers. Additionally, we offer a systematic analysis of Python-based FL frameworks, identifying key tools for hands-on experimentation and deployment. Finally, we have categorised pressing challenges across FL sub-fields, outlining key research directions for the community.

Despite significant progress, FL still faces major hurdles. Future research must focus on parameter-efficient models, communication-aware optimisation, knowledge alignment techniques, and the development of Green FL methodologies to enhance sustainability. Equally critical is the advancement of trustworthy FL, ensuring fairness, robustness, and interpretability, particularly in critical applications.

Ultimately, FL represents a paradigm shift in machine learning, offering a scalable alternative to traditional centralised approaches. It provides potential privacy benefits, addresses data governance challenges, and offers solutions for problems that are inherently distributed. As research progresses, refining FL methodologies and addressing existing challenges will be crucial in unlocking its full potential. By viewing FL through a Meta-Framework perspective, we provide a roadmap for more structured, adaptable, and impactful advancements in this evolving field.